\documentclass{article}
\usepackage[utf8]{inputenc}
\usepackage{amsfonts}
\usepackage{amsmath}
\usepackage[percent]{overpic}
\usepackage{verbatim}
\usepackage{xcolor}
\usepackage{mdframed}
\usepackage{xstring}
\usepackage{multirow}
\usepackage{anyfontsize}
\usepackage{tabularx}
\usepackage{wrapfig}
\usepackage{datatool}
\DTLsetseparator{:}
\usepackage{xifthen}
\usepackage{listofitems}
\usepackage{floatrow}
\usepackage[affil-it]{authblk}
\usepackage{ifthen}
\usepackage{comment}
\usepackage{bbm}
\usepackage[many]{tcolorbox}
\usepackage[euler]{textgreek}

\usepackage{transparent}
\usepackage[margin=1in]{geometry}

\title{Skillful Twelve Hour Precipitation Forecasts \\ using Large Context Neural Networks}

\author[*1]{Lasse Espeholt}
\author[2]{Shreya Agrawal}
\author[1]{Casper Sønderby}
\author[1]{Manoj Kumar}
\author[1]{Jonathan Heek}
\author[2]{Carla Bromberg}
\author[2]{Cenk Gazen}
\author[2]{Jason Hickey}
\author[3]{Aaron Bell}
\author[*1]{Nal Kalchbrenner}

\affil[1]{Google Research, Brain team}
\affil[2]{Google Research, AI for Weather team}
\affil[3]{Google Research, Kernel team}
\affil[*]{equal contribution}

\date{August 2021}

\usepackage[square,sort,comma,numbers]{natbib}
\usepackage{graphicx}
\usepackage{tikz}
 
\usepackage{todonotes}
\usepackage{subcaption}
\usepackage{caption}
\usepackage{booktabs}
\usepackage{float}
\usepackage{diagbox}

\usepackage{makecell}

\newcommand{\smallb}[1]{\scalebox{0.6}{{\normalsize #1}}}

\newcommand{\mt}{MetNet-2}

\begin{document}

\maketitle

\begin{abstract}
    The problem of forecasting weather has been scientifically studied for centuries due to its high impact on human lives, transportation, food production and energy management, among others.  Current operational forecasting models are based on physics and use supercomputers to simulate the atmosphere to make forecasts hours and days in advance. Better physics-based forecasts require improvements in the models themselves, which can be a substantial scientific challenge, as well as improvements in  the underlying resolution, which can be computationally prohibitive. 
    An emerging class of weather models based on neural networks represents a paradigm shift in weather forecasting: the models learn the required transformations from data instead of relying on hand-coded physics and are computationally efficient.  
    For neural models, however, each additional hour of  lead time poses a substantial challenge as it requires capturing ever larger spatial contexts and increases the uncertainty of the prediction~\citep{chantry2021opportunities}.
    In this work, we present a neural network that is capable of large-scale precipitation forecasting up to twelve hours ahead and, starting from the same atmospheric state, the model achieves greater skill than the state-of-the-art physics-based models HRRR and HREF that currently operate in the Continental United States. 
    Interpretability analyses  
    reinforce the observation that the model learns to emulate advanced physics principles. These results represent a substantial step towards establishing a new paradigm of efficient forecasting with neural networks.

\end{abstract}

\section{Introduction}

Probabilistic forecasts predict the likelihood of weather conditions at a given time and location. Weather conditions of interest can range from core atmospheric variables such as rate of rain and snow, wind velocity and direction, temperature, pressure levels and solar coverage to weather patterns such as hurricanes, wildfires and floods \citep{alleon2020plumenet, 9308323}. For the case of precipitation, a probabilistic forecast answers the question, ``What is the current probability of a given amount of precipitation occurring at a location and time in the future?''  

Short-term forecasting up to twelve hours in advance allows for predicting weather conditions with higher spatial and temporal precision than longer time ranges. This makes it possible for these forecasts to have substantial impact on society by helping with daily planning, energy management, transportation and the mitigation of extreme weather events, among others~\citep{FACETsAProposedNextGenerationParadigmforHighImpactWeatherForecasting}. Short-term forecasting is also a longstanding scientific challenge that combines our best understanding of the physics of the atmosphere with our most advanced computational capabilities.
Current operational models for short-term forecasting are Numerical Weather Prediction (NWP) models that rely on physics-based simulations. The atmospheric simulations make use of supercomputers with heterogeneous hardware that run virtually continuously in data centers around the globe and update the forecasts based on the latest observations.
The weather conditions that the models predict include hundreds of atmospheric and oceanic features. The forecasts usually have a frequency of one or more hours and a grid resolution of 3~km to 12~km. The accuracy of a physics-based forecast is tied to the grid resolution as more precise physics simulations require a finer representation of the  state of the atmosphere. 
This relationship creates a computational bottleneck inherent to physics-based models that has proven challenging to overcome~\citep{bauer2015quiet}. Besides resolution, the accuracy of the forecasts also depends on how well the physical models used in NWP describe the atmosphere at the various relevant scales; improving these models is a substantial scientific challenge by itself \cite{ScientificChallengesofConvectiveScaleNumericalWeatherPrediction}.

Due to the computational bottleneck, the large computational resources and the time lag that physics-based models incur when making a forecast, efficient models based on deep neural networks represent a promising alternative framework for weather modelling \citep{sonderby2020metnet,agrawal2019machine}. Instead of explicitly simulating the physics of the atmosphere,  

neural models learn the relationships between input observations and output variables directly from data. Neural networks can run in a matter of seconds on parallel hardware and can thus generate forecasts more frequently and with higher spatial resolution. The networks are also notably simple and can be specified with generic modules in a few dozens of lines of code without hand tuned routines for a specific task. These properties can not only offer improved forecasts, but also frequent and personalized forecasts~\citep{2018BAMS...99.2025R} and open avenues of new applications that rely on the models' efficiency and flexibility. However, showing that the neural networks are able to learn to emulate the physics of the atmosphere sufficiently well to make skillful high-resolution forecasts for up to twelve hours ahead--a period that requires an advanced understanding of atmospheric physics and is well beyond the skill of extrapolation and short-term nowcasting methods \cite{prudden2020review,xingjian2015convolutional,agrawal2019machine,DBLP:journals/corr/abs-2104-00954}--is a substantial open challenge at the core of our new modelling paradigm.

In this work, we present the Meteorological Neural Network 2 (\mt{}), a probabilistic weather model based on deep neural networks that is a successor to MetNet \cite{sonderby2020metnet}. \mt{} features a forecasting range of up to 12 hours of lead time at a frequency of 2 minutes. In order to capture sufficient input context, \mt{} uses input observations from a 2048~km~$\times$2048~km region and adopts novel neural network architectural elements in order to effectively process the large context. Such elements are (a) a  context-aggregating module that enables the receptive field of the network to double after every layer, (b) a strong lead time conditioning scheme and (c) a model parallel training setup utilizing multiple chips for increased memory and parallel computation.

\begin{figure}
\begin{center}
   \centering
  \includegraphics[width=1\linewidth]{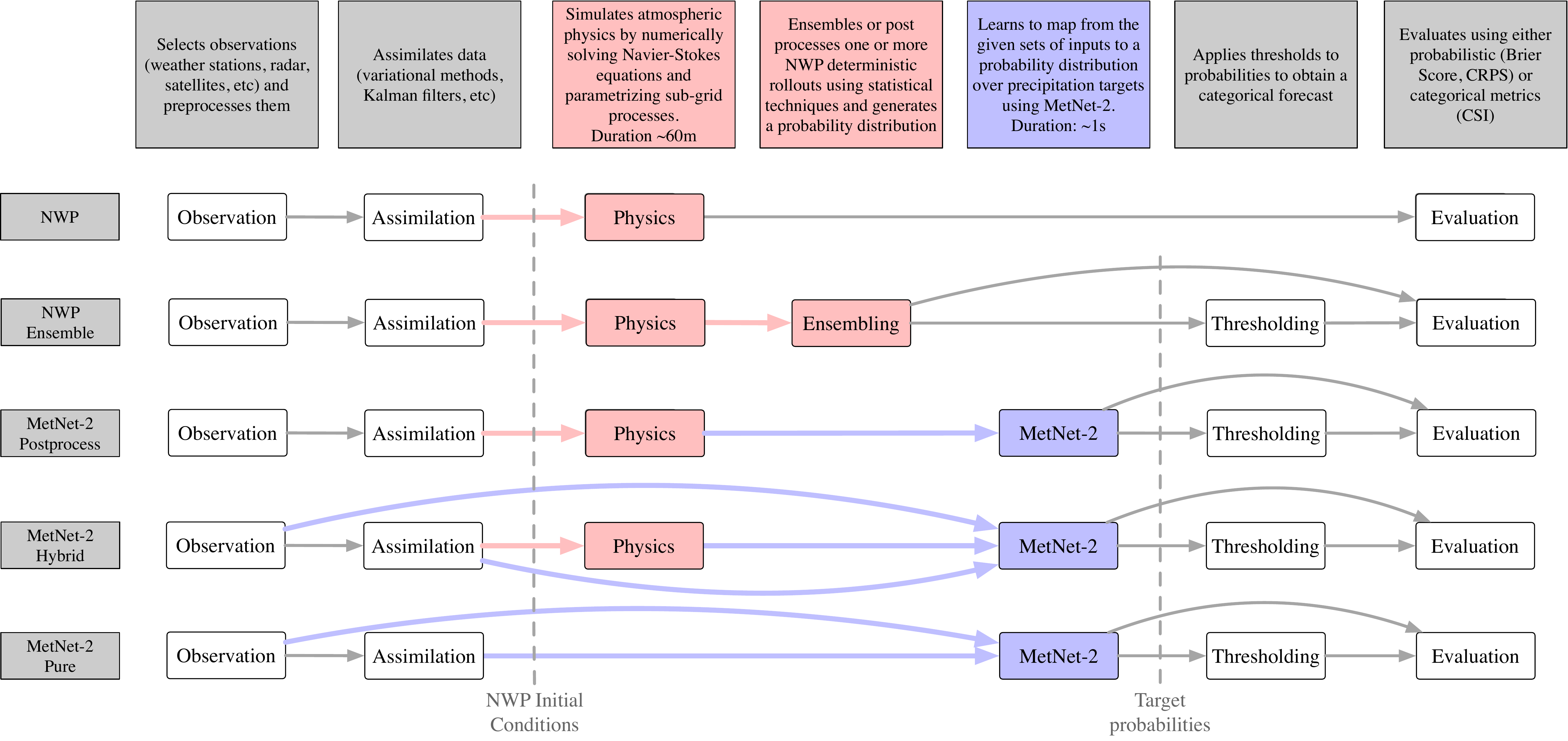}
\caption{Data and computation flow for NWP (HRRR), Ensemble NWP (HREF), MetNet-2 and the two MetNet-2 variants. MetNet-2 does not rely on the atmospheric simulation, whereas the other models do.  The observation and assimilation phases prepare a representation of the state of the atmosphere. Evaluation proceeds via categorical metrics that require calculating decision thresholds or via probabilistic metrics.}
\label{fig:models_flow}
\end{center}
\end{figure}

We train the MetNet-2 to forecast precipitation, a fast changing weather variable,  over a 7000 km $\times$ 2500 km region of the Continental United States (CONUS). 

MetNet-2's forecasts have a spatial resolution of 1~km~$\times$~1~km. 
The state-of-the-art NWP model High-Resolution Rapid Refresh (HRRR)~\citep{ANorthAmericanHourlyAssimilationandModelForecastCycleTheRapidRefresh} that is operational for the CONUS region makes hourly forecasts at a 3~km~$\times$~3~km resolution. According to a diverse set of standard evaluation metrics,  MetNet-2 outperforms the NWP model at predicting precipitation up to 12 hours of lead time without physics-based atmospheric forecasts. MetNet-2 is even more naturally compared with Ensemble NWP models due to the latter producing probabilistic forecasts similar to MetNet-2's own forecasts. A state-of-the-art physics-based ensemble model is the High Resolution Ensemble Forecast (HREF) \cite{href} that also operates in the CONUS region at a 3~km~$\times$~3~km resolution. Here too, MetNet-2 is able to outperform HREF up to 12 hours of lead time at predicting a cumulative rate of precipitation. Further results also indicate that, coupled with the forecast of the HRRR model, MetNet-2 is able to substantially improve HRRR's forecast with post-processing and hybrid modes. These results hold across a broad range of precipitation rates, from low to high.

MetNet-2's performance represents a step forward towards skillful forecasts with neural networks and suggests that MetNet-2 may be learning to emulate aspects of atmospheric physics. We perform an analysis into what MetNet-2 has learnt using state-of-the-art interpretation methods. The analysis reveals that MetNet-2 appears to make use of advanced physics principles when making its forecasts, which the model infers from the data.

\section{Framework}

MetNet-2 and NWP models gather empirical observations in order to obtain an initial state of the atmosphere as a basis for their forecasts. Observations come from a variety of sensors that are located on the ground in weather stations, on satellites, on airplanes and balloons, and on ocean buoys, among others. An important source of observations in our framework are those coming from ground radars that densely populate the Continental United States. The reflectivity, measured by these radars, estimates the amount of precipitation at a given time and location. The estimates are made every few minutes and have a relatively high spatial resolution of 1 km $\times$ 1 km. In our framework, we use two types of precipitation measures: instantaneous precipitation that comes from the radar reflectivity at a temporal frequency of two minutes; and hourly cumulative precipitation that represents the amount of precipitation over the preceding hour. In the latter, rain gauges at weather stations are used to further corroborate the radar measurements improving the data reliability. Both measures are provided by the Multi-Radar Multi-System (MRMS)~\cite{mrms}.  
While radar measurements provide information about the measures of precipitation, they do not describe the many other variables of the atmosphere, such as pressure, temperature and wind velocity and direction. Since the latter are not readily available, in order to incorporate them in our framework, we use the available set of atmospheric observations that result from the data assimilation process in the NWP model HRRR. This process uses various statistical and physics-based techniques to incorporate observations from the atmospheric sensors including those coming from the radars themselves. The resulting state is the starting point for NWP's simulation and we also adopt that state as an input for MetNet-2 to provide the model with more thorough information about the initial state of the atmosphere.
In addition to radar and assimilation features,  MetNet-2 also receives space-time coordinates for longitude, latitude, elevation and forecast time \cite{sonderby2020metnet} as well as optical satellite imagery; see Supplement~\ref{sup:dataset} for a full description of data inputs.

\label{sec:inputs}

\subsection{Ground Truth}

The radar precipitation measures are especially important for our task as they also serve as the ground truth training targets for the MetNet-2. The same MetNet-2 learns to predict both the instantaneous measure and the hourly cumulated measure at 2 minute and 60 minute intervals respectively, reflecting the temporal frequency of the estimates. The precipitation measures range from a rate of 0 mm/hr to 102.4 mm/hr, with the higher and more extreme rates of precipitation becoming increasingly rare in the data; see Supplement~\ref{sup:precip_events} summarizing how often various rates occur in the data.

\subsection{MetNet-2 Postprocess and Hybrid}

To comprehensively showcase MetNet-2's performance, we consider other training modes for MetNet-2 that, contrary to the default MetNet-2, make use of the outcome of NWP's atmospheric simulation. MetNet-2 Postprocess takes as an input NWP's forecast for a given lead time along with static location, altitude and time features and learns to map NWP's forecast as closely as possible to the ground truth. It also learns to correct for any systematic biases in NWP's forecast and makes the forecast probabilistic. MetNet-2 Hybrid learns to extract information from all the available inputs, including the twelve hourly forecasts from the NWP as well as the radar and assimilation inputs used in the default MetNet-2. Whereas an NWP is at its core a deterministic model, MetNet-2 and the variants are probabilistic at their core. An NWP can be made probabilistic through ensembling or through statistical post-processing~\cite{bauer2015quiet} such as, for example, the HREF model that produces an ensemble from 10 deterministic NWP members. Our own MetNet-2 Postprocess variant can be thought of as a probabilistic post-processing model for the deterministic NWP.
Figure~\ref{fig:models_flow} summarizes the framework.

\section{Model and Architecture}

A probabilistic forecast captures the combined uncertainty of both the measurements and the model:
\begin{equation}
    P(\mathbf{r}_{x,y,t}|t_0) = f(\mathbf{o},L,t_0)
\end{equation}
where $\mathbf{r}$ are rates of precipitation, $x,y,t$ are the location and target time of the forecast, $t_0$ is the time at which the forecast is made and $L = t-t_0$ is the lead time of the forecast. \mt{} bins the precipitation rates into 512 categories that allow the model to forecast arbitrary discrete probability distributions~\citep{sonderby2020metnet}.

The size of the input context plays a key role in the design of \mt{}'s architecture. Due to fast changing nature of the atmosphere, the longer the lead time of the forecast for a location $x,y$ the more context the model needs around $x,y$ in order to have sufficient  information for a skillful forecast. The context grows spatially in both dimensions and hence the total number of locations to attend grows quadratically in the length of the lead time. 
For a target patch of size 512~km~$\times$~512~km and forecast lead times of up to 12 hours, \mt{} uses an input context size of 2048~km~$\times$~2048~km. This amounts to between 64 and 85 km of context per hour of lead time in each spatial dimension.

Besides making a large context available to the network, the network must be able to process and attend to the key parts of the context with its architecture. It is a special feature of the weather forecasting task that these key parts vary as a function of lead time: for the same input patch of data as lead time increases, the network must attend to key parts of an ever growing potential region. These variable range dependencies  present a challenge for the design of the neural architecture.

\begin{figure}
    \centering

    \begin{subfigure}[t]{.32\textwidth}
        \includegraphics[width=1\textwidth]{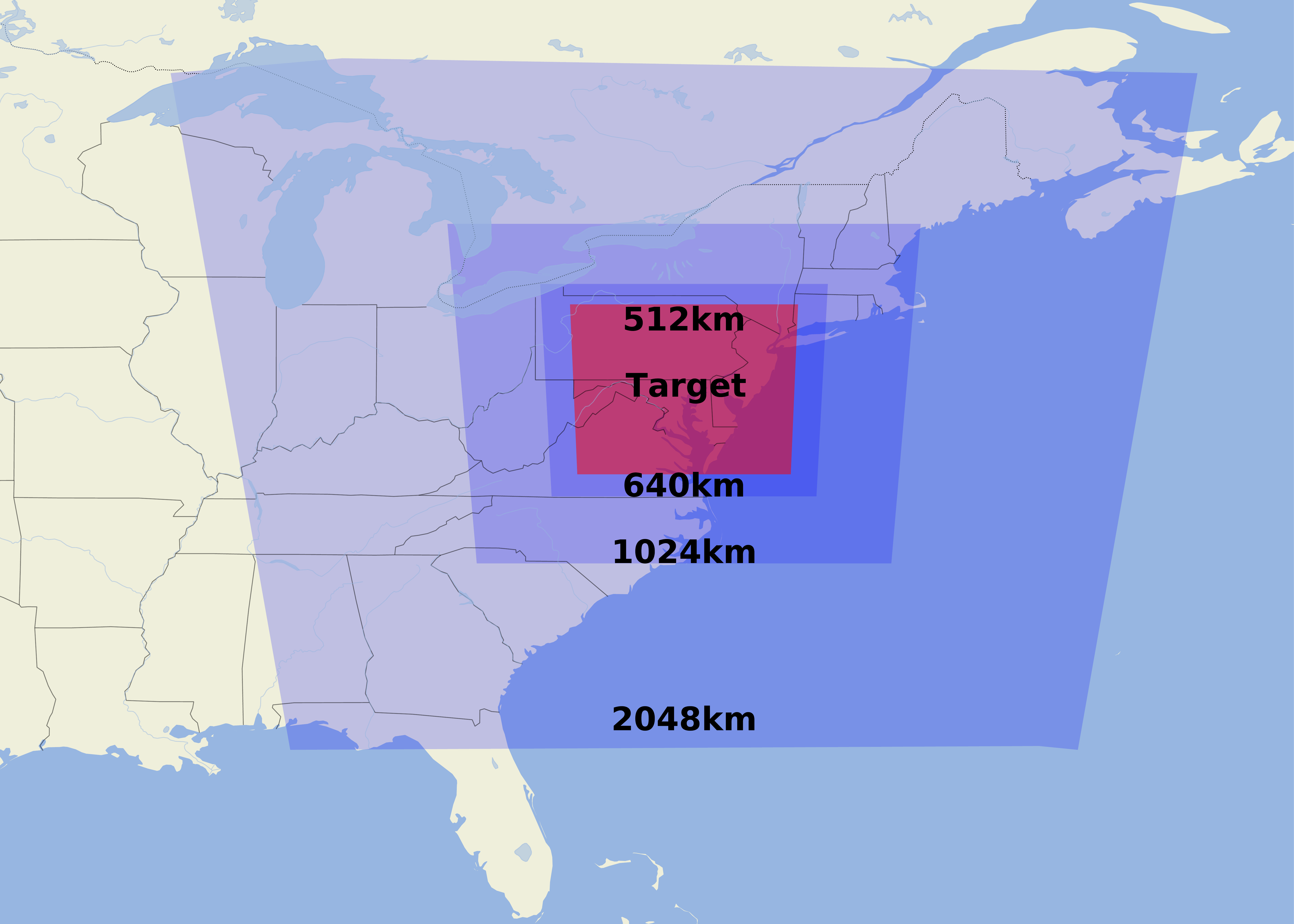}
        \caption{MetNet-2 captures increasing amounts of context around the
        target patch. The figure shows the effects of the orthographic projection for the context and target, onto Earth with an equirectangular projection.}
        \label{fig:context}
    \end{subfigure}\hfill%
    \begin{subfigure}[t]{.32\textwidth}
        \includegraphics[width=1\textwidth]{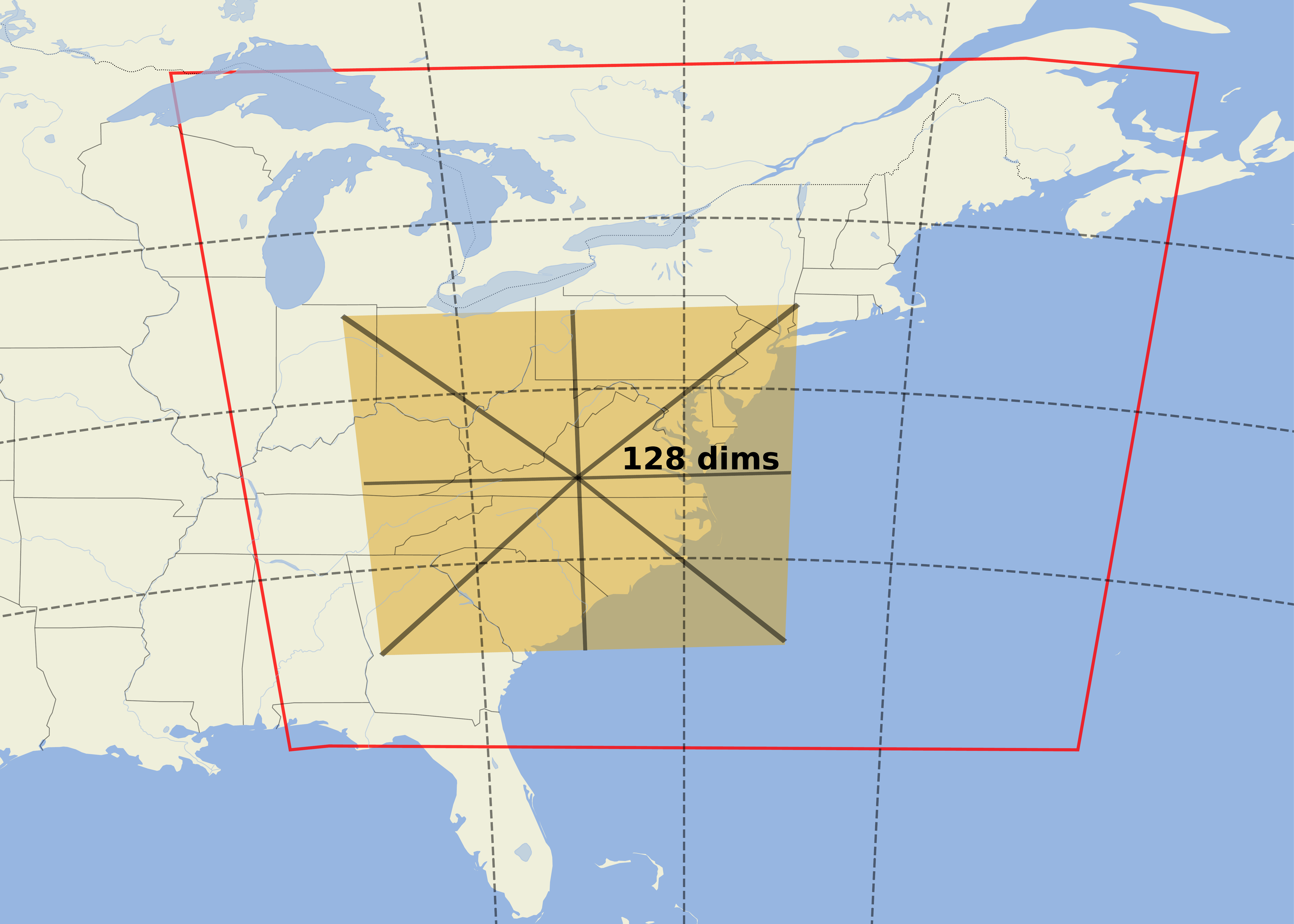}
        \caption{MetNet-2 layers are spread over $4\times 4$ TPU cores. The figure represents a convolution with dilation factor 128 reaching out to neighbouring TPUs to retrieve the relevant parts of the layer.}
        \label{fig:parallel}
    \end{subfigure}\hfill%
    \begin{subfigure}[t]{.32\textwidth}
        \centering
        \includegraphics[width=.8\textwidth, trim=0 1.5cm 0 0, clip]{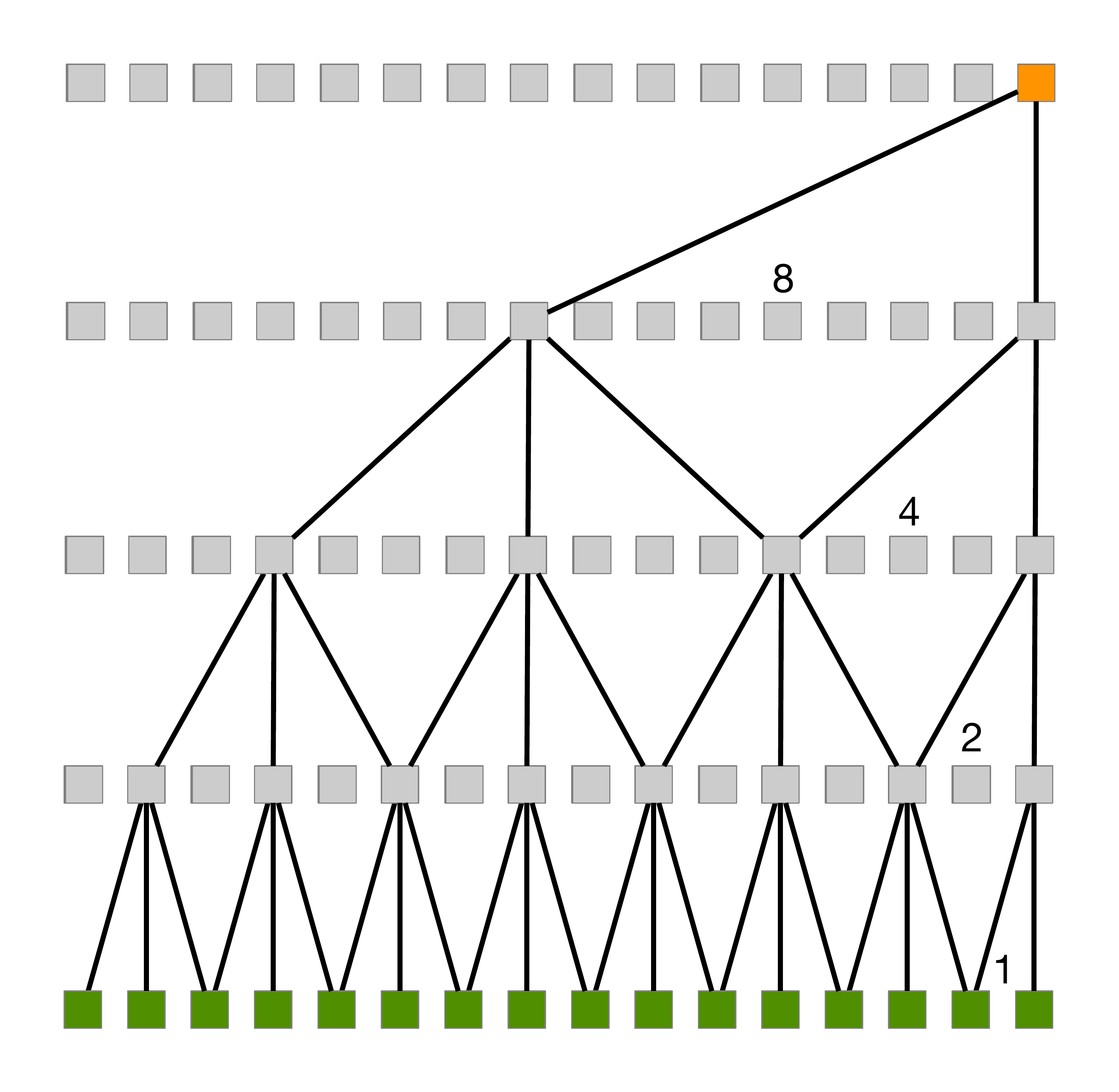}
        \caption{Pattern of connectivity when using convolutions with dilation factors that double at each layer. The total receptive field grows exponentially in the number of layers.}
        \label{fig:dilation}
    \end{subfigure}\\\vspace{.5cm}
    \begin{subfigure}{1\textwidth}
        \centering
        \includegraphics[width=\linewidth]{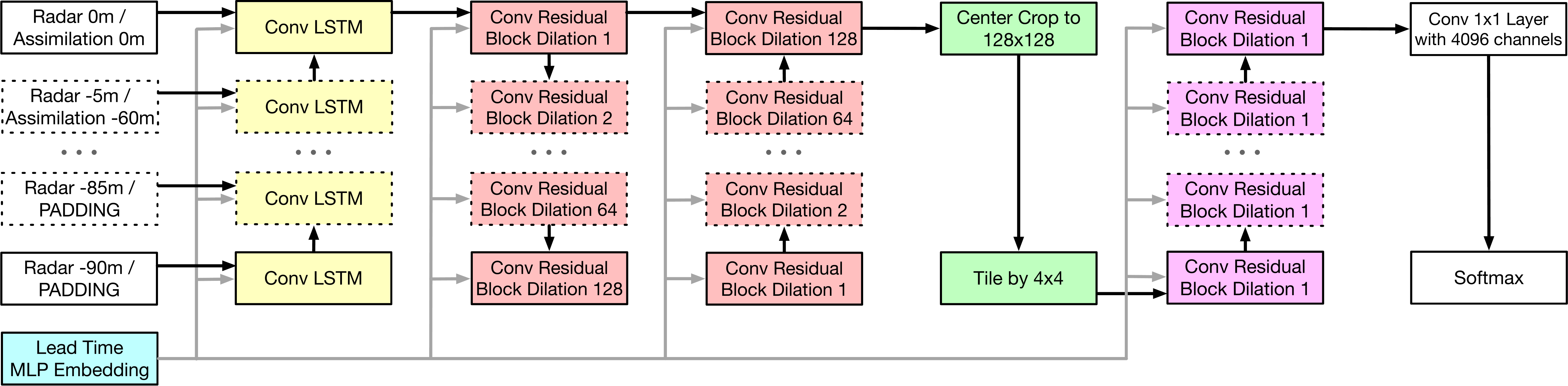}
        \caption{MetNet-2 architecture. Radar, Assimilation, and Geo-spatial features enter the network along a time dimension. A convolutional recurrent network \citep{xingjian2015convolutional} embeds the input step by step. A stack of convolutional blocks with increasing dilation captures the large context of the embedded input. After a center crop corresponding to the target patch area and a tiling operation that restores the 1 km $\times$ 1 km resolution, a final stack of convolutional blocks produces a distribution over precipitation levels for each target patch position. A rich embedding of the lead time index conditions each convolutional layer of the network.}
        \label{fig:architecture}
    \end{subfigure}
  \caption{}
  \label{fig:parmet2}
\end{figure}

\subsection{Input Encoder}

The input to \mt{} captures 2048~km~$\times$~2048~km of weather context for each input feature, but it is downsampled by a factor of 4 in each spatial dimension, resulting in an input patch of 512 $\times$ 512 positions.

In addition to the input patches having spatial dimensions, they also have a time dimension in the form of multiple time slices (see Supplement \ref{sup:data_types} Table~\ref{tab:obs_slices} for details.) This is to ensure that the network has access to the temporal dynamics in the input features.
After padding and concatenation together along the depth axis, the input sets are embedded using a convolutional recurrent network~\citep{xingjian2015convolutional} in the time dimension~\citep{sonderby2020metnet}. 

\subsection{Exponentially Dilated Convolutions}

The next part of MetNet-2's architecture aims at connecting each position in the layer representing the encoded context with every other position in order to capture the full context. \mt{} uses two-dimensional convolutional residual blocks with a sequence of exponentially increasing dilation factors of size 1,~2,~4,~...,~128~\cite{kalchbrenner2017neural,oord2016wavenet}. Dilation factors increase the receptive field of the convolution by skipping positions without increasing the number of parameters (see Figure~\ref{fig:parmet2}). 
Each position connects in this manner to all of the other 512~$\times$~512 positions of the encoded tensor. Figure~\ref{fig:residual_block_appendix} in Supplement D illustrates the exact residual block with the dilated convolutions. Three stacks of 8 residual blocks form this context aggregating part of MetNet-2's architecture.
The target patch of precipitation that \mt{} predicts corresponds to 512~km~$\times$~512~km and is centered in the middle of the 2048~km~$\times$~2048~km of the input patch. Because of that, the 512~$\times$~512 positions from the context aggregation in the input encoder, are cropped to to 128~$\times$~128 positions. To obtain a prediction for the full size target patch, we upsample four times in each dimension, effectively creating another layer of 512~$\times$~512 positions. This  is processed with another shallow network and ends with a categorical prediction over 512 precipitation levels for each target position. See Figure~\ref{fig:parmet2} for a full depiction of the architecture and Supplement~\ref{sup:arch_train} for additional architectural details.

\subsection{Conditioning with Lead Time}
\mt{} encodes the lead time as a one-hot embedding with indices from 0 to 359 representing the range between 2 and 720 minutes~\cite{sonderby2020metnet} and mapped into a continuous representation. Instead of feeding the lead time embedding only at the input of \mt{}, the embedding is applied both an additive and multiplicative factor to each of the two convolutional layers in the residual blocks of \mt{}~\cite{perez2018film}. 
This ensures that the output of each convolutional layer now depends directly on lead time.

\subsection{Neural Network Parallelism}

Due to the large input context, the $512 \times 512 \times d$ input/internal representations and the $512 \times 512$ target patch, the network does not fit on a single TPU core. Instead of reducing the dimensions of the target patch, which will cause redundant computation since each smaller target patch will have overlapping input context, or reducing the dimensions of the internal representations, we use model parallelism. The input and the target is split into a four by four grid and processed by 16 interconnected TPU cores, with each TPU core responsible for $128 \times 128$ of the target, as shown in Figure~\ref{fig:parallel}. The necessary communication at each layer is handled automatically and efficiently~\cite{DBLP:journals/corr/abs-2105-04663,jax2018github}.
This scheme that can be scaled further if needed makes it efficient to compute very large contexts for each target position.

\section{Forecasting Results and Interpretation}

\begin{figure}
\centering

\begin{subfigure}{\textwidth}
\centering
\begin{tabular}{ l  c  c c c c c c c c c c c } 
\toprule
Model / Hours & 1 & 2 & 3 & 4 & 5 & 6 & 7 & 8 & 9 & 10 & 11 & 12 \\ \midrule
NWP (HRRR) &  .17 & .16 & .16 & .15 & .14 & .14 & .15 & .15 & .15 & .14 & .14 & .15  \\ 
    {\mt} & \textbf{.44} & \textbf{.36} & \textbf{.32} & \textbf{.29} & \textbf{.26} & \textbf{.25} & \textbf{.24} & \textbf{.23} & \textbf{.23} & \textbf{.22} & \textbf{.22} & \textbf{.21}  \\ \midrule
    {\mt} Postprocess & .30 & .26 & .25 & .24 & .23 & .23 & {.24} & {.24} & {.24} & {.24} & {.24} & {.24} \\ 
    {\mt} Hybrid & \textbf{.44} & \textbf{.36} & \textbf{.32} & \textbf{.29} & \textbf{.28} & \textbf{.27} & \textbf{.26} & \textbf{.27} & \textbf{.26} & \textbf{.26} & \textbf{.25} & \textbf{.25}  \\
    \bottomrule
\end{tabular}
\caption{Critical Success Index scores of NWP (HRRR) and the \mt{} variants for instantaneous precipitation of $\geq$2 mm/hr. Scores are given for each of the 12 hours of lead time. }
\label{tbl:main_rate_2mm}
\end{subfigure}

\begin{subfigure}[b]{1\textwidth}
    \centering
    \vspace{.5cm}
    \begin{minipage}{.33\textwidth}
    \centering
    \hspace{.5cm}\scalebox{0.8}{2 mm}
    \\
    \includegraphics[width=1\textwidth]{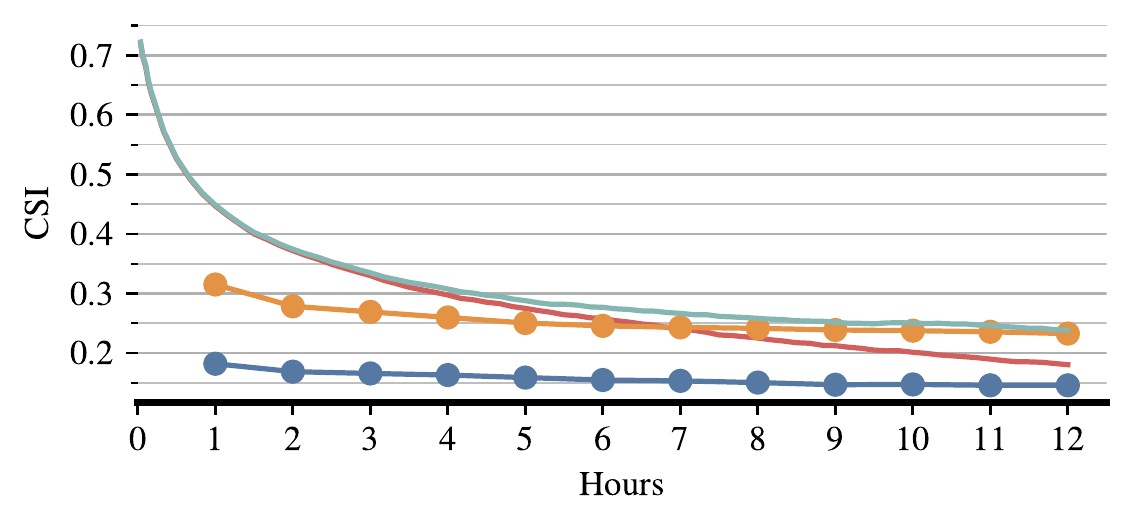}
    \end{minipage}%
    \begin{minipage}{.33\textwidth}
    \centering
    \hspace{.5cm}\scalebox{0.8}{8 mm}
    \\
    \includegraphics[width=1\textwidth]{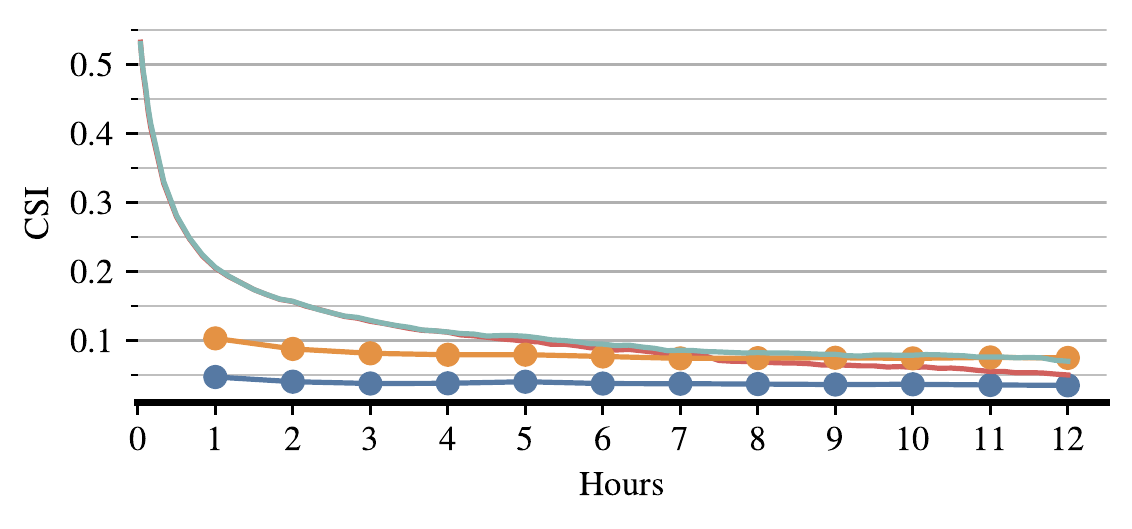}
    \end{minipage}%
    \begin{minipage}{.33\textwidth}
    \centering
    \hspace{.5cm}\scalebox{0.8}{20 mm}
    \\
    \includegraphics[width=1\textwidth]{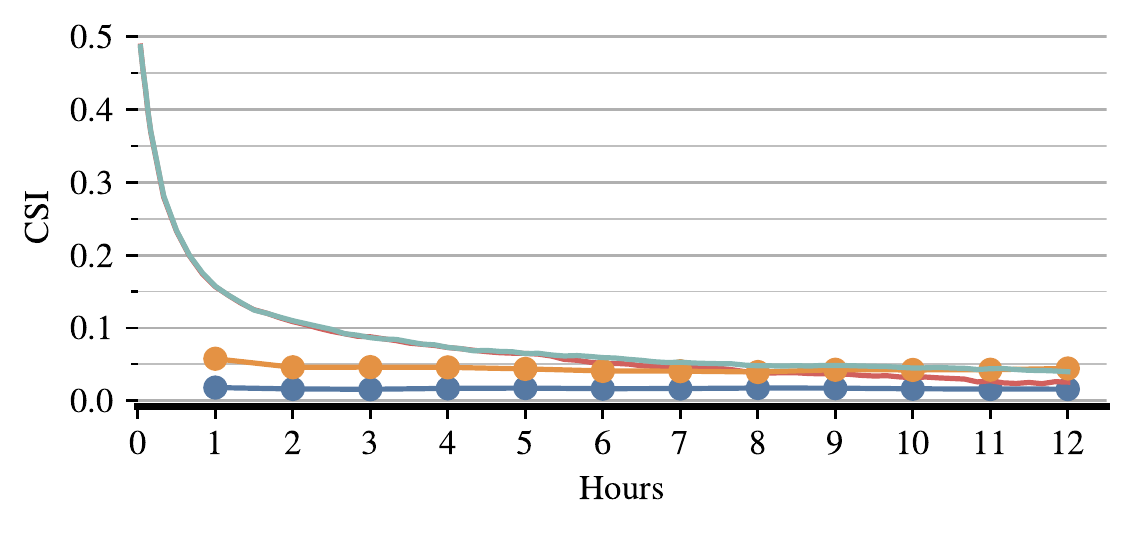}
    \end{minipage}\\
    \includegraphics[scale=.6]{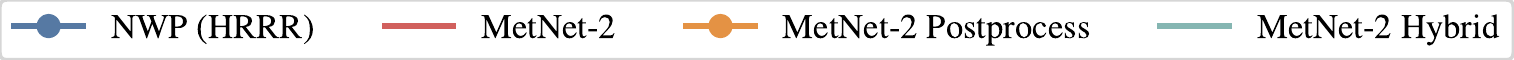}
    \caption{Critical Success Index metric for instantaneous rates of 2 mm/hr, 8 mm/hr and 20 mm/hr.}
    \label{fig:main_csi_rate}
\end{subfigure}

\begin{subfigure}[b]{1\textwidth}
    \centering
    \begin{minipage}{.6\textwidth}
    \centering
    \includegraphics[width=1\textwidth]{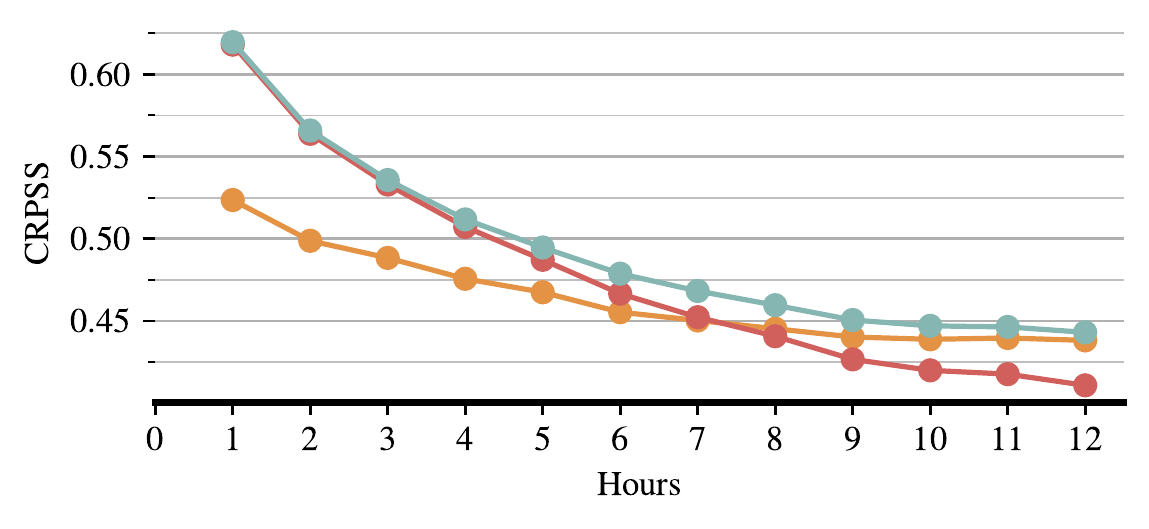}\\
    \includegraphics[scale=.6]{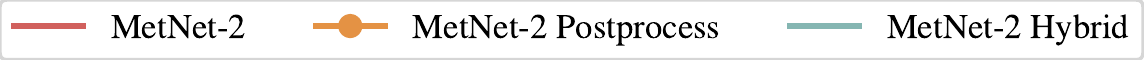}
    \end{minipage}%
    \begin{minipage}[b]{.38\textwidth}
    \caption{Continuous Ranked Probability Score Skill for instantaneous precipitation. The score tracks the relative improvement of MetNet-2, MetNet-2 Postprocess and MetNet-2 Hybrid over NWP (HRRR).}
    \label{fig:main_crpss_a}
    \end{minipage}
\end{subfigure}

\begin{subfigure}[b]{1\textwidth}
    \centering
    \begin{minipage}{.6\textwidth}
    \centering

    \includegraphics[width=1\textwidth]{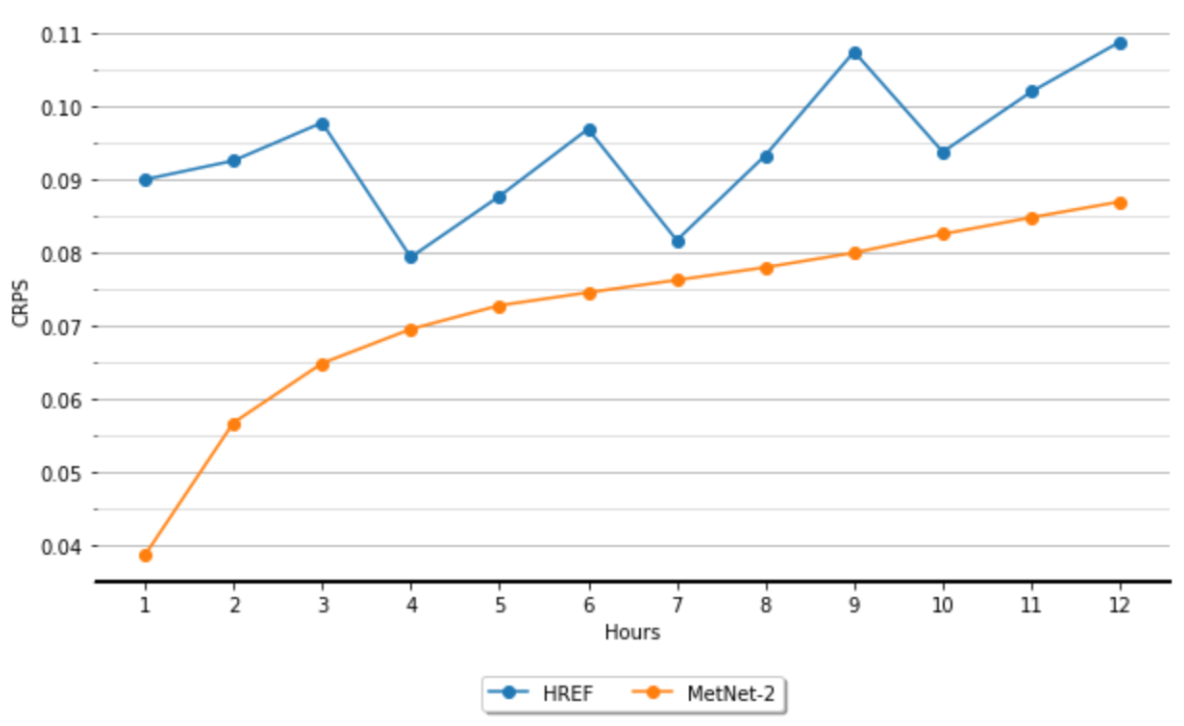}\\
    \end{minipage}%
    \begin{minipage}[b]{.38\textwidth}
    \caption{Continuous Ranked Probability Score for cumulative precipitation for MetNet-2 and the physics-based ensemble HREF. }
    \label{fig:main_crpss_b}
    \end{minipage}
\end{subfigure}

\caption{Tables of results based on CSI and CRPS and CRPSS.}

\label{fig:main_results}
\end{figure}

MetNet-2 does not rely on the atmospheric simulation that is the most computationally intensive part in the NWP models. In the ensemble HREF model, the results of 10 such simulations make up the final forecast. It is then all the more remarkable that MetNet-2 is able to outperform the NWP models across the twelve hour forecasting range, from both low (0.2 mm/hr) to high precipitation levels (20 mm/hr) and for both types of precipitation measures, instantaneous in the case of HRRR and hourly cumulative in the case of both HRRR and HREF. Figure~\ref{fig:main_results} illustrates these results. MetNet-2 scores better on both categorical metrics such as the Critical Success Index (CSI), that require MetNet-2 to threshold the probabilistic output, and on probability-centered metrics like the Brier Score and the Continuous Ranked Probability Score Skill (CRPSS) that measure the probabilistic error in the forecast. The skill gap between MetNet-2 and HRRR is greatest in relative terms at the earliest hours and decreases gradually over the twelve hour range.  Figure~\ref{fig:main_case_a}~and~\ref{fig:main_case_b} represent two case studies of HRRR's and MetNet-2's forecasts. For a video comparison of MetNet-2's and HREF's forecasts, see the accompanying Google AI blog post. One major difference in the forecasts between HRRR, HREF and MetNet-2 is their handling of uncertainty. The NWP model HRRR makes a hard choice of one location over another when forecasting precipitation, whereas HREF and MetNet-2 spread the uncertainty over multiple likely locations with their respective probabilities. The uncertainty grows as lead time increases and tends to be higher for rarer events like higher rates of precipitation. As a result longer lead times tend to feature a greater spread in the predicted distributions with lower probability values assigned at any specific location (see for example the probabilities in Figure~\ref{fig:main_case_a} and Figure~\ref{fig:main_case_b}).  
The high CRPSS for instantaneous rate that MetNet-2 obtains relative to HRRR, from 0.65 at one hour of lead time to 0.35 at 12 hours of lead time, indicates the ability of MetNet-2 to estimate forecast probabilities well when evaluated against the ground truth. A similar pattern also holds for hourly cumulative precipitation (Supplement~\ref{sup:results}). The case studies in Figures~\ref{fig:main_case_a} and \ref{fig:main_case_b} also represent the probabilistic error based on the Brier score attained by the models. Regional evaluation sheds further light on MetNet-2's performance; MetNet-2 performs well across diverse regions that see varying levels of annual precipitation (Supplement~\ref{sup:regional_eval}).

\newtcolorbox{scorebox}[1][]{
    width=51.5pt,
    height=9.5pt,
    arc=0pt,
    boxsep=0cm,
    toprule=0pt,
    leftrule=0pt,
    bottomrule=0pt,
    rightrule=0pt,
    colframe=gray,
    breakable,
    nobeforeafter,
    left=2pt,
    right=0pt,
    top=1pt,
    bottom=0pt,
    enhanced jigsaw,
    opacityframe=0.5,
    opacityback=0.8
}
\newcommand{\predfull}[3] {%
  \raisebox{-0.5\height}{%
    \includegraphics[scale=.1]{case_studies/#1_#2_#3.png}}}
\newcommand{\predinlinefull}[6] {
\raisebox{-0.5\height}{%
\begin{overpic}[scale=.1]{case_studies/#1_#2_#3.png}
    \put(0,#6){%
    \begin{scorebox}[]%
    \scalebox{0.35}{%
        \begingroup%
        \setlength\tabcolsep{5pt}%
        \setsepchar{ }%
        \readlist\score{#5}%
        \fontfamily{phv}\selectfont%
        \begin{tabular}{@{}cccccc@{}}%
            & .2 & 1 & 2 & 4 & 8 \\
         #4 & \score[1] & \score[2] & \score[3] & \score[4] & \score[5] \\
        \end{tabular}%
        \endgroup%
    }%
    \end{scorebox}
    }%
\end{overpic}%
}}
\newcommand{\inlinebottom}[0]{0}
\newcommand{\inlinetop}[0]{81.5}
\newcommand{\myrowspace}{\vspace{3pt}}%
\newcommand{\myendspace}{\vspace{10pt}}%
\newcommand{\vtitle}[1]{\rotatebox[origin=c]{90}{\hspace{-.2cm}\scalebox{.7}{#1}}}
\newcommand{\pred}[2] {\predfull{\casedir/\casetimestamp}{#1}{#2}}
\newcommand{\predinline}[5]{\predinlinefull{\casedir/\casetimestamp}{#1}{#2}{#3}{#4}{#5}}

\newcommand{\predfulllarge}[4] {%
  \raisebox{-0.5\height}{%
    \begin{overpic}[scale=.2]{case_studies/#1_#2_#3.png}%
    #4%
    \end{overpic}
}}
\newcommand{\predlarge}[2] {\predfulllarge{\casedir/\casetimestamp}{#1}{#2}{}}

\newcommand{\predlargepin}[3] {\predfulllarge{\casedir/\casetimestamp}{#1}{#2}{#3}}

\begin{figure}
\centering
\begingroup

\newcommand{\caseprefix}{video_example2_1mm_rate_54eb14dc00000000}
\newcommand{\casetimestamp}{1546516800}
\newcommand{\firsthourpos}[0]{\inlinebottom}
\newcommand{\thirdhourpos}[0]{\inlinebottom}
\newcommand{\sixthhourpos}[0]{\inlinebottom}
\newcommand{\twelvedhourpos}[0]{\inlinebottom}

\begin{subfigure}{\textwidth}
\centering
\newcommand{\casedir}{\caseprefix _all_rates_3d}
\setlength\tabcolsep{1.5pt}
\begin{tabular}{@{}c c c c c r@{}}
  & 1 hr & 3 hr & 6 hr & 12 hr & \\ 
  \vtitle{Ground Truth}
      & \predlarge{MRMS}{00060}
      & \predlarge{MRMS}{00180}
      & \predlarge{MRMS}{00360}
      & \predlarge{MRMS}{00720}\myrowspace
      & \raisebox{-0.5\height}{\includegraphics[scale=.5]{case_studies/\casedir/colorbar_fixed.pdf}~\rotatebox[origin=l]{90}{\hspace{.4cm}\smallb{Precipitation (mm)}}} \\
  \vtitle{\mt{}}
      & \predlargepin{MetNet2}{00060}{
            \put(15,63){81 \%}%
            \put(24,25){\linethickness{0.25mm}\color{black}\line(0,1){35}}%
        }
      & \predlarge{MetNet2}{00180}
      & \predlarge{MetNet2}{00360}
      & \predlargepin{MetNet2}{00720}{
            \put(81,21){8 \%}%
            \put(53,23){\linethickness{0.25mm}\color{black}\line(1,0){25}}%
        }\myrowspace
      & \raisebox{-0.5\height}{\shortstack[r]{\includegraphics[scale=.5]{case_studies/\casedir/colorbar_probs_8.0.pdf}\\\includegraphics[scale=.5]{case_studies/\casedir/colorbar_probs_4.0.pdf}\\\includegraphics[scale=.5]{case_studies/\casedir/colorbar_probs_2.0.pdf}\\\includegraphics[scale=.5]{case_studies/\casedir/colorbar_probs_1.0.pdf}\\\includegraphics[scale=.5]{case_studies/\casedir/colorbar_probs_0.2.pdf}}~\rotatebox[origin=l]{90}{\hspace{-.35cm}\smallb{Prob. of Precipitation (mm)}}} \\
\end{tabular}

\caption{The varying intensities 
correspond to MetNet-2's predicted probability for the respective rate of precipitation showing the probabilistic structure of the forecast. For each precipitation rate we show intensities only for probabilities above the respective CSI threshold. MetNet-2 tends to become less certain over time about the amount and location of the precipitation. See Figure~\ref{fig:main_case_a_app} for MetNet-2 Postprocess and MetNet-2 Hybrid.}

\end{subfigure}
\begin{subfigure}[t]{.5\textwidth}%
\centering%
\newcommand{\casedir}{\caseprefix _all_rates}%
\caption{Decision boundaries based on thresholds optimally chosen for CSI. The numbers in each graphic correspond to the CSI score greater or equal to the respective rate. }
\end{subfigure}%
    ~ 
\begin{subfigure}[t]{.5\textwidth}%
\centering%
\newcommand{\casedir}{\caseprefix _rate_brier_1_00}%
\caption{Brier score maps that quantify the error of the probabilistic prediction.
  Lower error is better. NWP's forecast is assigned probability 1 everywhere. Only the region of high quality radar signal is visualized (Figure~\ref{fig:context_masked})}%
\end{subfigure}
\caption{Case study for Thu Jan 03 2019 12:00 UTC of the North West coast of the US with forecasts of instantaneous precipitation.}

\label{fig:main_case_a}
\endgroup
\end{figure}

\begin{figure}
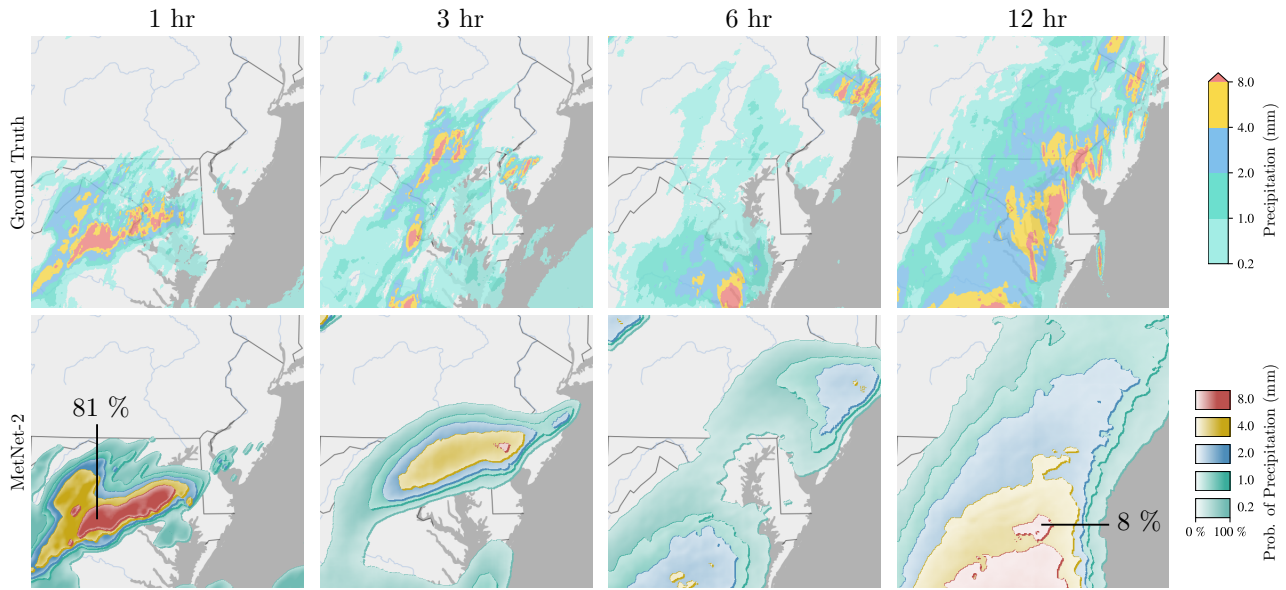
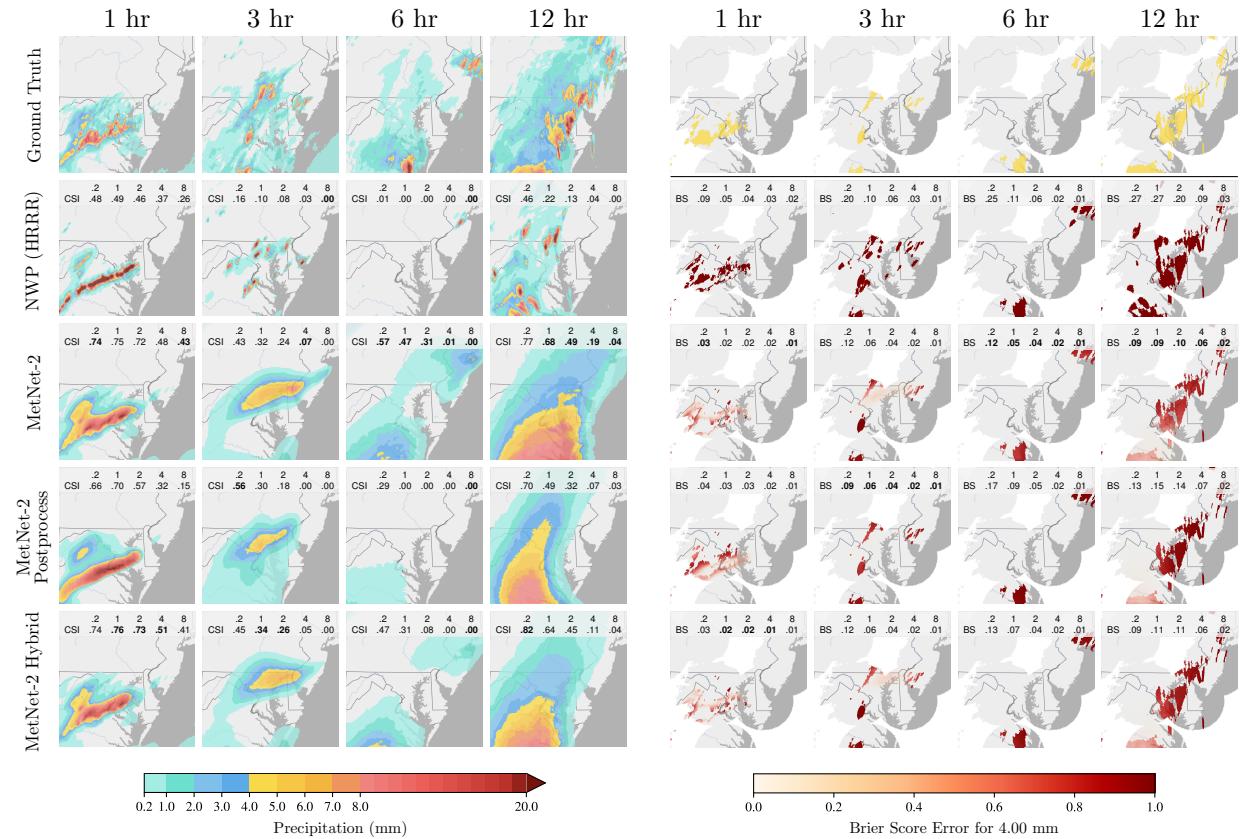

      \vspace*{-2\baselineskip}
\centering
\begingroup

\newcommand{\caseprefix}{video_hurricane_isaias_1mm_cumulative_89c80adc00000000}
\newcommand{\casetimestamp}{1596484800}
\newcommand{\firsthourpos}[0]{\inlinetop}
\newcommand{\thirdhourpos}[0]{\inlinetop}
\newcommand{\sixthhourpos}[0]{\inlinetop}
\newcommand{\twelvedhourpos}[0]{\inlinetop}

\begin{subfigure}{\textwidth}
\centering
\newcommand{\casedir}{\caseprefix _all_rates_3d}
\caption{MetNet-2's forecasted probabilities distinguish between a certain rate of precipitation that is predicted with high probability (e.g.\ at 1 hour) or with low probability (e.g.\ at 12 hours).  See Figure~\ref{fig:main_case_b_app} for MetNet-2~Postprocess and MetNet-2~Hybrid.}
\end{subfigure}
\begin{subfigure}[t]{.5\textwidth}%
\centering%
\newcommand{\casedir}{\caseprefix _all_rates}%
\caption{CSI decision boundaries for hourly cumulative precipitation and respective scores.}%
\end{subfigure}%
    ~ 
\begin{subfigure}[t]{.5\textwidth}%
\centering%
\newcommand{\casedir}{\caseprefix _rate_brier_4_00}%
\caption{Brier score maps that quantify the error of the probabilistic prediction.
  Lower error is better. NWP's forecast is assigned probability 1 everywhere. Only the region of high quality radar signal is visualized (Figure~\ref{fig:context_masked})}%
\end{subfigure}

\caption{Case study of Hurricane Isaias, a Category 1 hurricane, that caused widespread destruction and economic damage. The forecast time is Mon Aug 03 2020 20:00 UTC on the East coast of the United States. The measure is the gauge-corrected hourly cumulative precipitation.}
\label{fig:main_case_b}
\endgroup
\end{figure}

Many aspects of the framework play a central role in obtaining MetNet-2's performance. We find that the size of the input context of 2048~km~$\times$~2048~km improves performance over context sizes of 1536~km~$\times$~1536~km, 1024~km~$\times$~1024~km and 512~km~$\times$~512~km (Supplement~\ref{sup:abl_context}). The additional observations of the atmosphere that the assimilation process incorporates also have an impact on MetNet-2's performance, especially at later hours (Supplement \ref{sup:abl_subsets}). MetNet-2 is able to extract information from a broad range of observations and any additional observations are likely to improve MetNet-2's performance. Furthermore, both removing the special conditioning scheme for the lead time index (Supplement~\ref{sup:abl_lead_time}) and  limiting the maximum dilation factor to 16 or below also impact  MetNet-2's performance negatively (Supplement~\ref{sup:abl_dilation_factor}). 

The MetNet-2 variants Postprocess and Hybrid shed additional light onto MetNet-2. MetNet-2 Postprocess learns to estimate the uncertainty of the different parts of the NWP model HRRR's forecast. This allows MetNet-2 Postprocess to cleverly spread the uncertainty over HRRR's forecast and invariably increases the resulting forecast skill based on both the CSI and the CRPSS metric. Despite the marked performance increase, MetNet-2 still outperforms MetNet-2 Postprocess up to about seven hours of lead time, after which MetNet-2 Postprocess' performance inches somewhat higher. The information contributed to MetNet-2 Postprocess by NWP's atmospheric simulation becomes especially valuable in these later lead times. Finally, MetNet-2 Hybrid makes it possible to combine the best information from both approaches. Remarkably, even at 12 hours, the neural network MetNet-2 is able to provide valuable information on top of HRRR's forecast and outperform MetNet-2 Postprocess' forecast skill.

\begin{figure}[H]
    \centering
    \begin{subfigure}{.48\textwidth}
      \centering
      \includegraphics[width=1\linewidth]{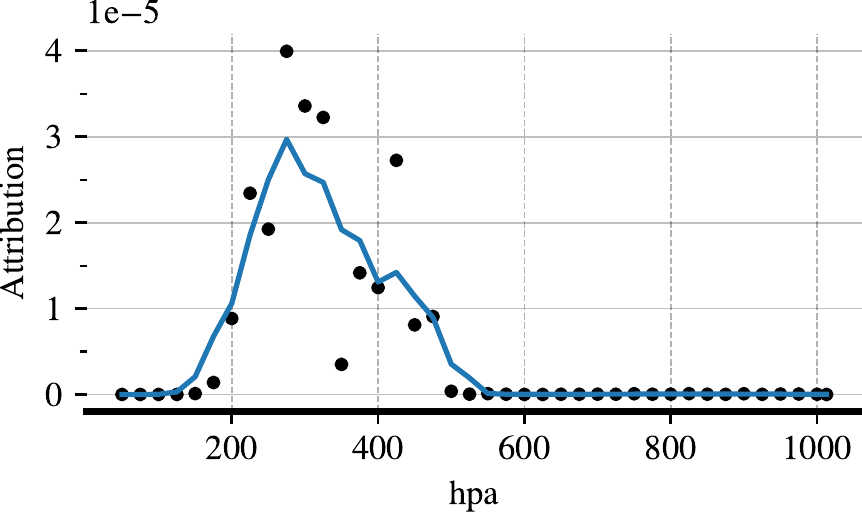}
      \caption{At different pressure levels.}
      \label{fig:absv}
    \end{subfigure}
    \begin{subfigure}{.48\textwidth}
      \centering
      \includegraphics[width=1\linewidth]{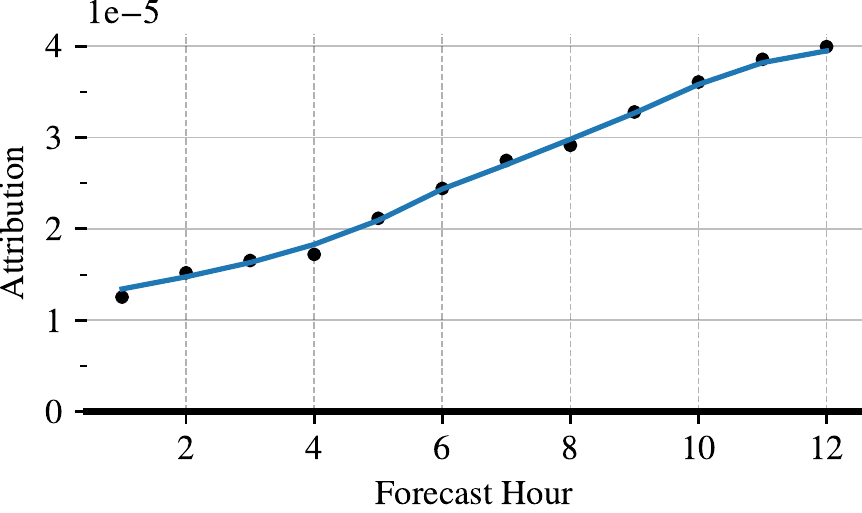}
      \caption{At 250 hPa of 1-12 hr forecasts.}
      \label{fig:absv250}
    \end{subfigure}
    \caption{Attribution of Absolute Vorticity, i.e., the amount a single input pixel contributes to a single pixel probability prediction. On the left, the solid line is 100 hPa moving average over the attributions and on the right, it is a 3 hour moving average. Attributions for this feature is relatively higher compared to majority of the input features.} 
    \label{fig:av}
\end{figure}

MetNet-2's remarkable performance makes it important to understand what the physics-free neural network is learning. This can help researchers gain new insight about interactions between different meteorological variables and ensure that the model conforms to our prior knowledge about weather physics. We adopt a state-of-the-art neural interpretation technique called Integrated Gradients to attribute predictions to the input variables~\citep{sundararajan2017axiomatic}. Among the notable findings, Figure~\ref{fig:absv} shows that the relative importance of absolute vorticity is small for near-term forecasts, but grows in importance as lead time increases all the way up to 12 hours. The importance of upper-level vorticity for a twelve hour forecast is consistent with what is known as quasi-geostrophic theory, a non-trivial set of simplifications and filtering of the equations of motion. A key result in the theory is that positive vorticity in the upper-troposphere is consistent with upward motion in the lower-troposphere~\cite{bluestein1992synoptic}.  This upward motion does not directly trigger precipitation, but prepares the atmosphere for convection. See Supplement \ref{sup:interpretation} for other key findings.

\section{Conclusion}

 MetNet-2's strong performance for both low and high levels of precipitation for both precipitation measures, its ability to estimate uncertainty, its independence from atmospheric simulation, its design simplicity and the rapid and different nature of MetNet-2's computation represent a step towards a fundamental shift in forecasting from physics-based models to learning-based ones. The results also show how neural networks can learn to emulate complex and large-scale physics paving the way for ever more ambitious applications of neural nets in the physical sciences.
Though designed for geo-spatial prediction, little in MetNet-2's architecture is specific to precipitation. This raises hopes that MetNet-2 could work well for many other weather variables possibly at once and even learn to transfer from one variable to the next and improve overall performance. Similarly, direct sensor data, although not readily available, can likely be used in place of the assimilation state and further reduce MetNet-2's total latency to essentially just the time required for observation. The flexibility and non-specificity of MetNet-2 may also enable broadening and combining the domain of the predicted variables beyond core weather and encompass general geo-spatial prediction.

\section*{Acknowledgements}
We would like to thank Amy McGovern, Rob Carver and Stephan Hoyer for insightful discussions and comments on the draft of the paper, as well as Zack Ontiveros, David McPeek, Ian Gonzalez, Claudio Martella, Samier Merchant, Fred Zyda, Daniel Furrer and Marcin Andrychowicz for project and technical contributions. The authors of the manuscript declare no competing interests.

\bibliography{main}
\bibliographystyle{plain}

\newpage
\appendix

\section{Supplement: Related Work}
\label{sup:related_work}

Operational weather forecasts based on NWP rely on decades-long research using laws of physics for atmospheric simulation. While the tremendous increase in observational data, scientific and computing advancements has resulted in an increase in the forecast skill by about one day every decade \citep{bauer2015quiet}, the recent success of deep learning in scientific domains has spurred an interest in its application to weather forecasting \citep{chantry2021opportunities,doi:10.1098/rsta.2020.0097}.
For the domain of nowcasting, Prudden et al.~\citep{prudden2020review} provide a detailed review of radar-based nowcasting techniques and several ML approaches used in the past. Shi et al.~\citep{xingjian2015convolutional} applied a recurrent neural network based approach using convolutional LSTMs~\citep{hochreiter}. Agrawal et al.~\citep{agrawal2019machine} use a U-Net model and turn the forecasting problem into an image-to-image translation problem. Trebing et al.~\citep{TREBING2021178} apply attention modules and show a reduction in model parameter size while maintaining performance. Ravuri et al.~\citep{DBLP:journals/corr/abs-2104-00954} use a generative model that is radar-based only. However, all of these techniques have been shown to be skillful only from 0 to at most 3 hours of lead time. The MetNet model~\citep{sonderby2020metnet} shows initial results for for skilful nowcasting of precipitation up to 7-8 hours over a HRRR baseline. On the longer forecast sides, deep learning has also been applied for seasonal forecasts of weather events including extreme ones~\citep{Ham2019DeepLF,Yan2020TemporalCN}.

As we move from nowcasting to short or medium range forecasting the biggest challenge is the lack of high resolution observational data that is typically obtained through data assimilation. There have also been efforts to postprocess NWP predictions using neural networks by incorporating them into the set of input features \citep{NeuralNetworksforPostprocessingEnsembleWeatherForecasts}; this is similar to our MetNet-2 Postprocess variant.

\section{Supplement: Dataset}
\label{sup:dataset}

The training data consists of 1,230,585 patches of size 2048~km~$\times$~2048~km at the input and targets of size 512~km~$\times$~512~km including all 360 (2 to 720 minutes) time slices. The training area covers a region of 7000 $\times$ 2500 kilometers. We sample target patches from the input context region minus an all around border of $\approx$512 km. The input context is padded for all regions outside of the 7000 $\times$ 2500 CONUS. The validation data used for developing the models consists of 11,991 patches and the test data of 39,864 patches. The training, validation and test data are drawn from non-overlapping ranges of hours, with ``black out'' periods of 12 hours in between, over a period of observations of 3 years from July 2017 to August 2020. This ensures that the model does not learn any spurious training and evaluation correlations within any single day. HRRR only generates forecasts starting at full hours.

\subsection{Data Types}
\label{sup:data_types}
\begin{figure}[H]
    \centering
    \begin{subfigure}{.48\textwidth}
      \centering
  \includegraphics[width=1\linewidth]{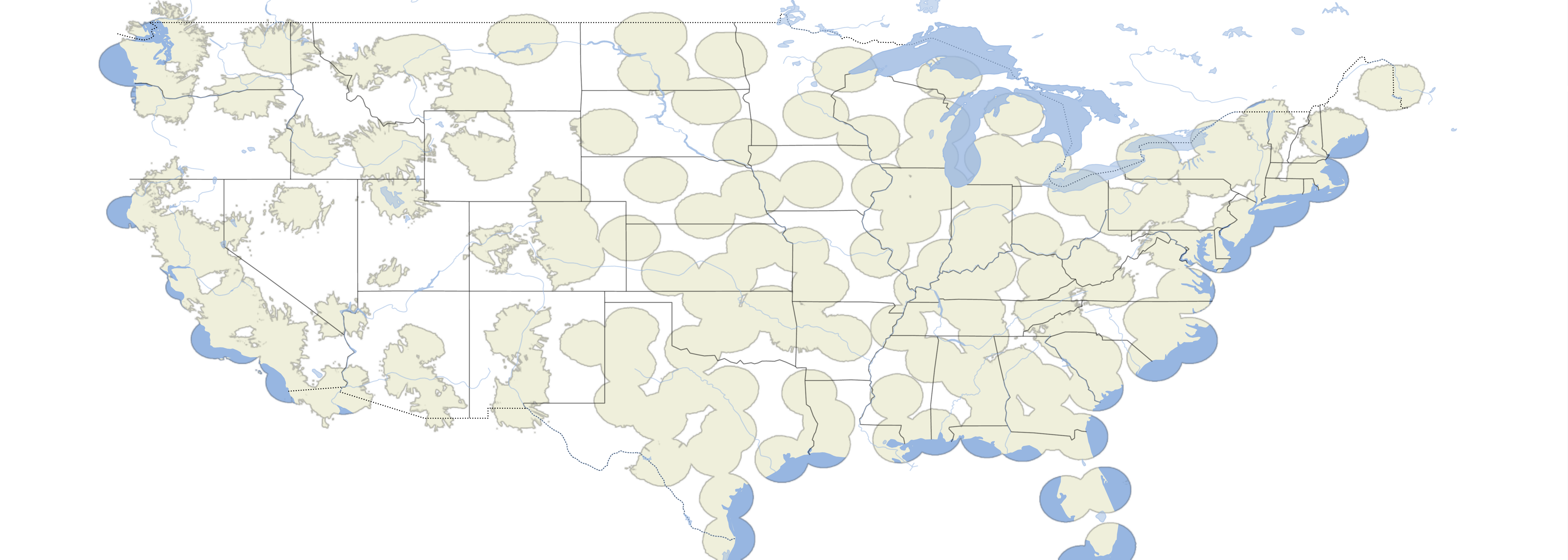}     
    \end{subfigure}
    \begin{subfigure}{.48\textwidth}
      \centering
  \includegraphics[width=1\linewidth]{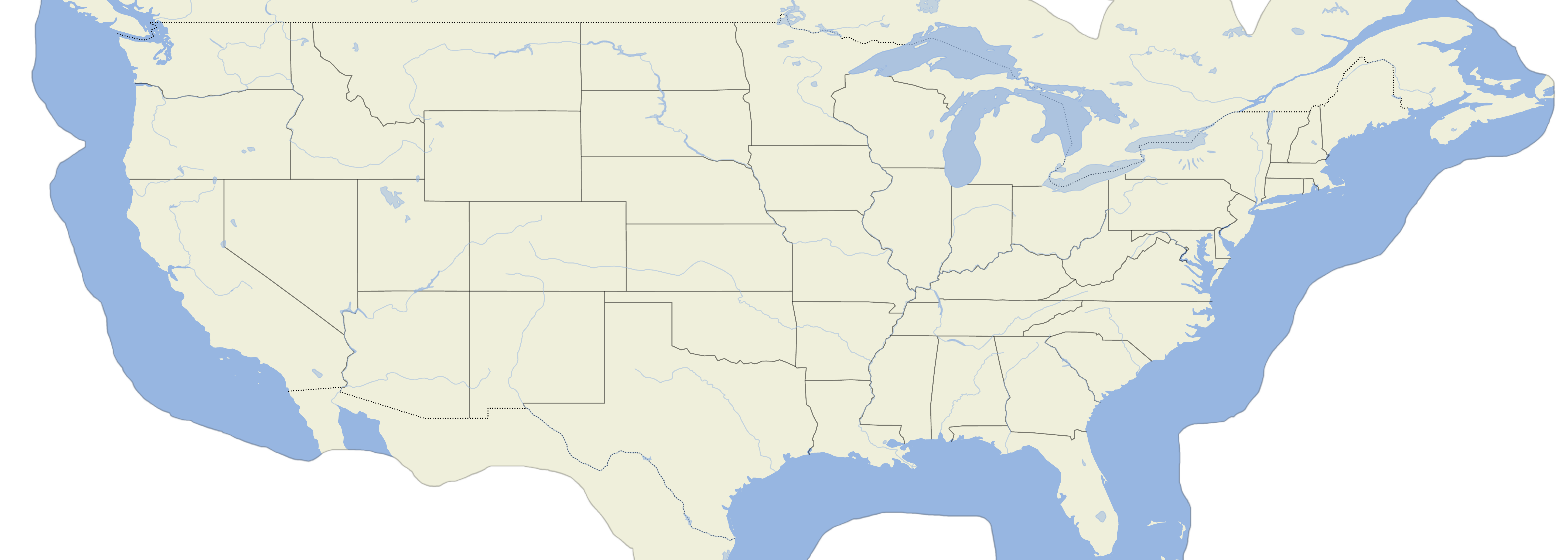}     
    \end{subfigure}
  \caption{Continental United States masked using the ``better" range of NOAA NEXRAD on the left for evaluation and the training mask ``roi'' on the right.}
  \label{fig:context_masked}
\end{figure}

MetNet-2's radar data comes from the Multi-Radar Multi-Sensor (MRMS)\cite{mrms}
data that uses the reflectivity of ground radars and processes it to estimate precipitation. The data has a spatial resolution of approximately 1 km $\times$ 1 km and a temporal resolution as low as 2 minutes. MRMS provides both of the precipitation measures that we use: the instantaneous rate of precipitation, that we refer to here as MRMS Instantaneous, and the hourly cumulative precipitation, that we refer to as MRMS Cumulative.
The MRMS Cumulative estimates use rain gauges at weather stations to further corroborate the radar measurements. For ensuring as precise evaluation as possible, we evaluate only regions where radar measurements are of high quality (Figure~\ref{fig:context_masked}). For training we use a larger mask corresponding to the full range of the radars.

The NWP model HRRR also provides atmospheric observations at forecast time $t_0$ that is a result of an hourly updated data assimilation system \citep{ANorthAmericanHourlyAssimilationandModelForecastCycleTheRapidRefresh}. We use 612 of these variables that capture atmospheric and oceanic variables such as pressure at various altitudes in the atmosphere, humidity, wind velocity and direction, temperature and current precipitation levels. The data  generally originates from sparse empirical measurements but the result of the data assimilation process is a dense grid with a spatial resolution of 3 km $\times$ 3 km. 

Additional input types include images from the GOES~\cite{goes} weather satellites. These come in 16 optical bands, have a resolution of approximately 1 km $\times$ 1 km and a temporal frequency of $\approx$15 minutes. Furthermore, simple geographical features for longitude, latitude, elevation~\cite{sonderby2020metnet} and temporal features that encode the forecast time $t_0$ are additional sources of inputs for the MetNet-2 models. 

Finally, the precipitation forecast from the NWP (HRRR) model is also used as input in the MetNet-2 Postprocess and Hybrid variants. These are two features that include an instantaneous rate forecast and an hourly cumulative forecast that correspond to the same measures, respectively, as MRMS Instantaneous and MRMS Cumulative. Table~\ref{tab:obs_characteristics} summarizes the sets of input features and their main characteristics.

\begin{table}
\begin{center}
\begin{tabular}{ l c c c c c c c  } 
\toprule
 {Input Sets} & {\#Features} & {Resolution} (km) & {Frequency} (min) \\
 \midrule
    Radar (MRMS Instantaneous) & 1 & 1 & ${\approx}2$ \\ 
    Gauge-corrected Radar (MRMS Cumulative) & 1 & 1 & 60 \\ 
    GOES Optical Images & 16 & 1 & ${\approx}15$  \\
    Assimilation & 612 & 3 & 60  \\
    Geospatial coordinates & 3 + 3 & 1 & 1   \\
    NWP Forecast for MetNet-2 Postprocess/Hybrid & 2 & 3 & 60 \\
\bottomrule

\end{tabular}
\caption{For each set of inputs to MetNet-2, we list the respective number of features, the native resolution of the data and the temporal frequency by which each set of features is generated.}
\label{tab:obs_characteristics}
\end{center}
\end{table}

\subsection{Distribution of Precipitation Events}
\label{sup:precip_events}
The target data on which we train and evaluate MetNet-2 and its variants comes from the MRMS Instantaneous and MRMS Cumulative radar measures. The distribution of the levels of precipitation is summarized in Table~\ref{tab:precipitation_distribution}. The specific numbers are from the test data within the evaluation mask (Figure~\ref{fig:context_masked}), but the training data distribution is similar. Around $\approx$93~\% of precipitation events corresponds to no precipitation of 0~mm/hr. Higher amounts of precipitation events get increasingly more rare. This has consequences for learning too, as the MetNet-2 needs to learn good estimates for higher levels of precipitation from a relative paucity of data. During training, we resample the data to contain more high amount precipitation.

\begin{table}
\begin{center}
 \begin{tabular}{l r r r r r r r}
 \toprule
 & \multicolumn{6}{c}{Precipitation Bucket (mm/hr)} \\
 \midrule
\textbf{Measure Type} & 0 & .2  & 1 & 2  & 4  & 8 & $\geq$20 \\ 
 \midrule
 {MRMS Instantaneous} &	93.83 \%	& 3.23 \% & 1.43 \% & .92 \% & .41 \% & .13 \% & .06 \% \\
 {MRMS Cumulative} & 92.74 \% & 4.24 \% & 1.49 \% & .92 \% & .42 \% & .17 \% & .03 \% \\
 \bottomrule
\end{tabular}
\caption{Distribution of precipitation levels in test data for the two precipitation measures.}
\label{tab:precipitation_distribution}
\end{center}
\end{table}

\section{Supplement: Evaluation Metrics}
\label{sup:metrics}
We evaluate the quality of the precipitation forecasts using three different metrics, the Critical Skill Index (CSI) \citep{donaldson_rj_objective_1975}, the Brier score \citep{brierscore} and the Continuous Ranked Probability Score (CRPS)~\citep{DecompositionoftheContinuousRankedProbabilityScoreforEnsemblePredictionSystems}. 

\subsection{Critical Success Index}

 The CSI score is a binary categorical score similar to the F1 score that goes beyond plain accuracy to take into account more aspects of the confusion matrix. CSI is defined as follows:
\begin{equation}
    CSI = TP/(TP+FN+FP)
\end{equation}
where TP are true positives, FN are false negatives and FP are false positives.
Like the F1 score, the CSI score is not directly applicable to the probability distributions that \mt{} produces.  To make a categorical decision, for a binary category corresponding to an amount of precipitation greater or equal to a given rate $r$, we calculate on held-out data a probability threshold between 0 and 1. If the total predicted probability mass for rates $\geq r$ exceeds the threshold, then we take it to be a positive prediction for this rate category. This is the same procedure as used for the F1 score in Sønderby et al.~\citep{sonderby2020metnet}. We calculate the CSI scores of \mt{} on multiple rates ranging from small amounts of precipitation (0.2 mm/hr) to high amounts of precipitation (20 mm/hr).

\subsection{Brier Score}

The Brier score by contrast is not categorical and measures the magnitude of the error between the ground truth rate and the probability that a model predicts for that rate:
\begin{equation}
    BS_r = \frac{1}{N} \sum_n (P(r) - \mathbbm{1}(y \geq r))^2
\end{equation}
where $y$ is the ground truth rate at the respective time and location and $\mathbbm{1}$ is the indicator function. We calculate the Brier score from the raw probabilities produced from the output of \mt{} and do not calibrate the learnt distribution after training, which can affect the absolute values of the Brier score. For HRRR's deterministic forecast, we can also compute the Brier score  by assigning probability 1 to the predicted rate value and 0 everywhere else.

\subsection{Continuous Ranked Probability Score}

CRPS in essence is the mean squared error between the cumulative density function of the prediction and that of the ground truth. It can be defined as the Brier Score integrated over all rates, and therefore:
\begin{equation}
    CRPS = \sum_{r=0.2}^{102.4} BS_r \times 0.2
\end{equation}
where we sum over all 512 discrete buckets of rate in increments of 0.2 mm/hr.

\subsection{Skill Scores}

We compute skill scores for both Brier Score and CRPS. For both, we obtain skill scores with respect to HRRR's performance, as follows:
\begin{align*}
\text{Brier Skill Score:}  &&  BSS = 1 - \frac{BS}{BS_{NWP}} \\
\text{Continuous Ranked Probability Skill Score:}   && CRPSS = 1 - \frac{CRPS}{CRPS_{NWP}} \\
\end{align*}

\section{Supplement: Architecture and Training Details}
\label{sup:arch_train}
We provide details about the hyper-parameters governing the architecture and the training procedure (see Table ~\ref{tab:hyperparams}). The MetNet-2s are trained in parallel on 64 TPU chips (128 TPU cores). It takes the models approximately 48 hours to train the convergence. We use weight decay to regularize the training and the models tend to converge faster on later lead times than on earlier lead times. Due to this imbalance, we perform per-lead-time-hour checkpointing, whereby we maintain the best checkpoints according to validation set performance for each hour of the 12 hours of lead time.

\begin{table}
\begin{center}
\begin{tabular}{ l c } 
 \toprule
 \textbf{Architecture} & \textbf{Value} \\
 \midrule
 LSTM Channels & 128 \\
Encoder Blocks & 18 \\
Encoder Channels & 384 \\
Upsampler Blocks & 2 \\
Upsampler Channels & 512 \\
Lead Time Network - Layers & 2 \\
Lead Time Network - Features & 2048 \\
\bottomrule
\end{tabular}\hspace{1cm}%
\begin{tabular}{ l c } 
\toprule
 \textbf{Training Hyperparamters} & \textbf{Value} \\
 \midrule
 Weight Decay & 0.1 \\
 Polyak Decay & 0.9999 \\
 Optimizer & Adam \\
 Beta & 0.9 \\
 Batch size & 16 \\
 Learning rate & 2e-5 \\
 Training steps & 500K \\
 \bottomrule
\end{tabular}
\caption{Hyperparameters for MetNet-2 governing the architecture and the training procedure.}
\label{tab:hyperparams}
\end{center}
\end{table}

\begin{table}
\begin{center}
\begin{tabular}{ l c  c  } 
\toprule
 \textbf{Observations} & \textbf{Time Slices} \\ \midrule
    {Radar MRMS} &  13\\ 
    {Satellite GOES} &  3 \\
    {Assimilation} &  3  \\
    {Geospatial Coordinates} &  1 \\
    {NWP Forecast} & 1 \\
    \bottomrule
\end{tabular}
\caption{Number of time slices used for each observation set.}
\label{tab:obs_slices}
\end{center}
\end{table}

\begin{figure}
  \centering
  \includegraphics[width=1\linewidth]{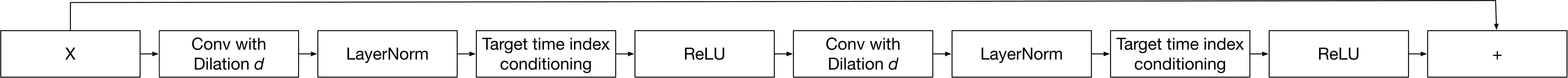}
  \caption{Residual block composition in the MetNet-2 architecture.}
  \label{fig:residual_block_appendix}
\end{figure}

\subsection{Rich Conditioning with Lead Time}
\label{app:lead_time_cond}

The rich form of conditioning that MetNet-2 uses works as follows. After every convolutional layer in MetNet-2's encoding and upsampling residual blocks, two dense projections map the continuous representation of the lead time index into a bias and a scale vector. The bias and scale vector are difference for each residual block. We add the bias to the output of every convolution at every position of the tensor and then multiply the result at each position with the scale vector. 
This ensures that the output of each convolutional layer now depends directly and very strongly on the lead time.  

\subsection{Lead Time Conditioning for Hourly Cumulative Precipitation}

For instantaneous precipitation, MetNet-2 gets a single lead time index to indicate the lead time minutes. When predicting hourly cumulative precipitation, the encoding is not a single one-hot embedding of just one index, but the 30 one-hot embeddings that correspond to all the 2 minute intervals in the hour preceding the lead time. This lets MetNet-2 share the same time indices for both targets and simultaneously training on the two targets while still being able to distinguish between the two targets. Per step, the combined model is on par with the individually trained models, signaling transfer learning between the targets.

\section{Supplement: Forecasting Results}
\label{sup:results}

We perform an extensive evaluation of the models using the various metrics and for multiple levels of precipitation, both for the instantaneous measure and for the hourly cumulative. The NWP and MetNet-2 Postprocess models are evaluated at every full hour, whereas the MetNet-2 and MetNet-2 Hybrid models at two minute intervals. One important thing to note is that we do not count the delay of the computation in our evaluations. Despite the fact that NWP, MetNet-2 Postprocess and MetNet-2 Hybrid technically take approximately 60 minutes longer than MetNet-2 to generate their forecasts due to the time used by the atmospheric simulation, we ignore this additional delay and do not consider it when comparing the performance based on lead time. NWP's prediction for $n$ minutes is compared with MetNet-2's prediction for $n$ minutes independently of the time it took to generate it and even if the generation time itself is longer than $n$ minutes. The same holds for the MetNet-2 variants that rely on the atmospheric simulation.

\begin{figure}
    \centering
    \begin{subfigure}[b]{0.48\textwidth}
      \centering
      \includegraphics[width=1\linewidth]{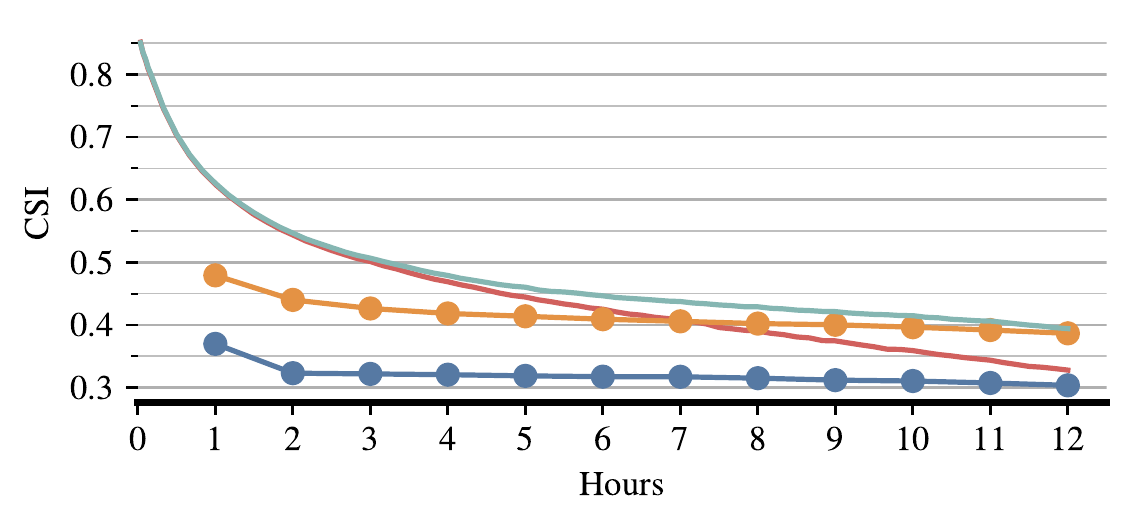}
      \caption{.2 mm}
    \end{subfigure}
    \begin{subfigure}[b]{0.48\textwidth}
      \centering
      \includegraphics[width=1\linewidth]{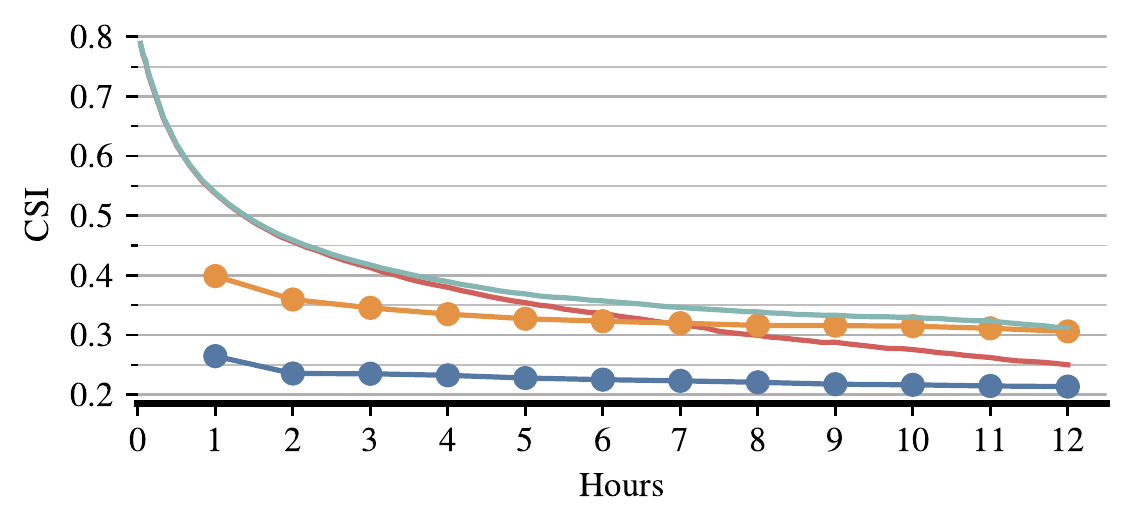}
      \caption{1 mm}
    \end{subfigure}

    \begin{subfigure}{.48\textwidth}
      \centering
      \includegraphics[width=1\linewidth]{plots/rate_large_test_2.00_CSI.pdf}
      \caption{2 mm}
    \end{subfigure}
    \begin{subfigure}{.48\textwidth}
      \centering
      \includegraphics[width=1\linewidth]{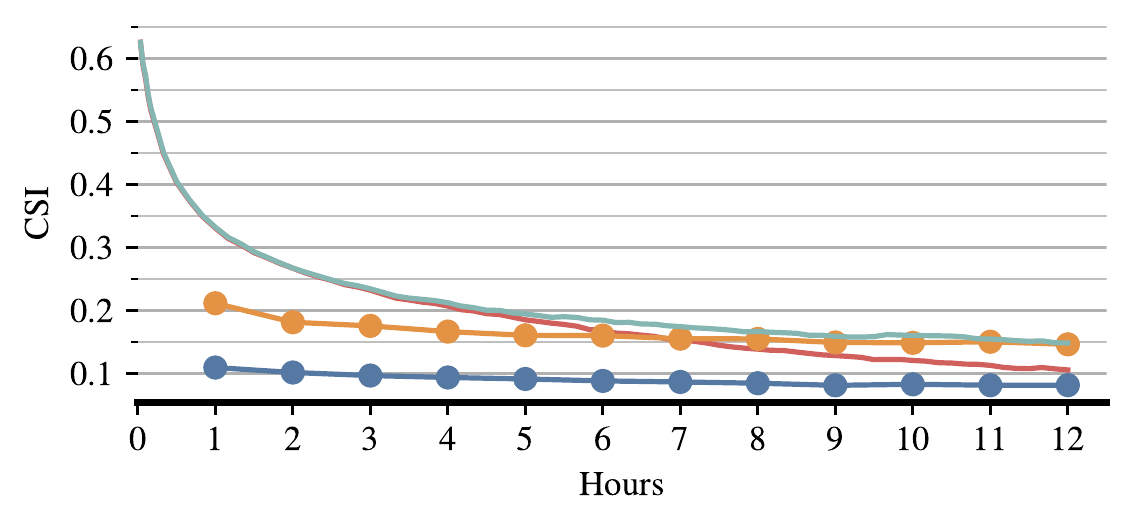}
      \caption{4 mm}
    \end{subfigure}
    
    \begin{subfigure}{.48\textwidth}
      \centering
      \includegraphics[width=1\linewidth]{plots/rate_large_test_8.00_CSI.pdf}
      \caption{8 mm}
    \end{subfigure}
    \begin{subfigure}{.48\textwidth}
      \centering
      \includegraphics[width=1\linewidth]{plots/rate_large_test_20.00_CSI.pdf}
      \caption{20 mm}
    \end{subfigure}
    \includegraphics[scale=.6]{plots/rate_large_test_legend_horizontal_CSI.pdf}

    \caption{CSI performance of the models for various low to high instantaneous precipitation rates of $\geq$.2 mm/hr, $\geq$1 mm/hr, $\geq$2 mm/hr, $\geq$4 mm/hr, $\geq$8 mm/hr and $\geq$20 mm/hr.}
    \label{fig:appendix_csi_rate}
\end{figure}

\begin{figure}
    \centering
    \begin{subfigure}[b]{0.48\textwidth}
      \centering
      \includegraphics[width=1\linewidth]{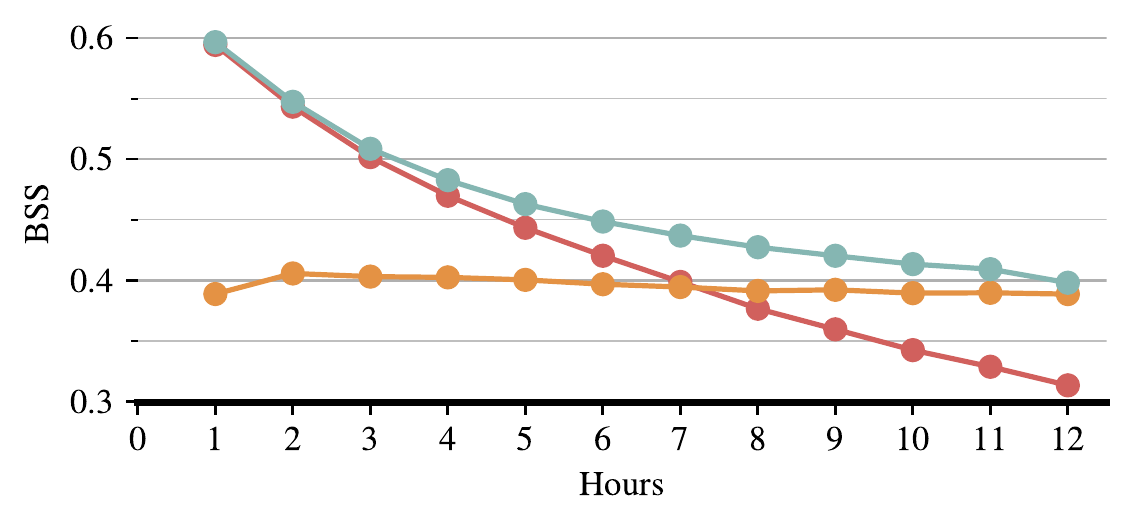}
      \caption{.2 mm}
    \end{subfigure}
    \begin{subfigure}[b]{0.48\textwidth}
      \centering
      \includegraphics[width=1\linewidth]{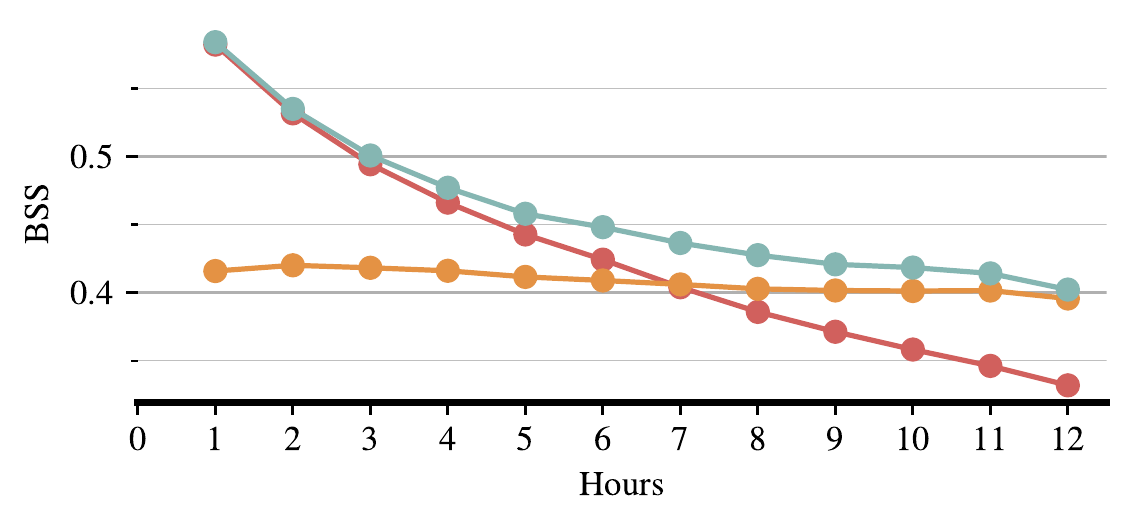}
      \caption{1 mm}
    \end{subfigure}

    \begin{subfigure}{.48\textwidth}
      \centering
      \includegraphics[width=1\linewidth]{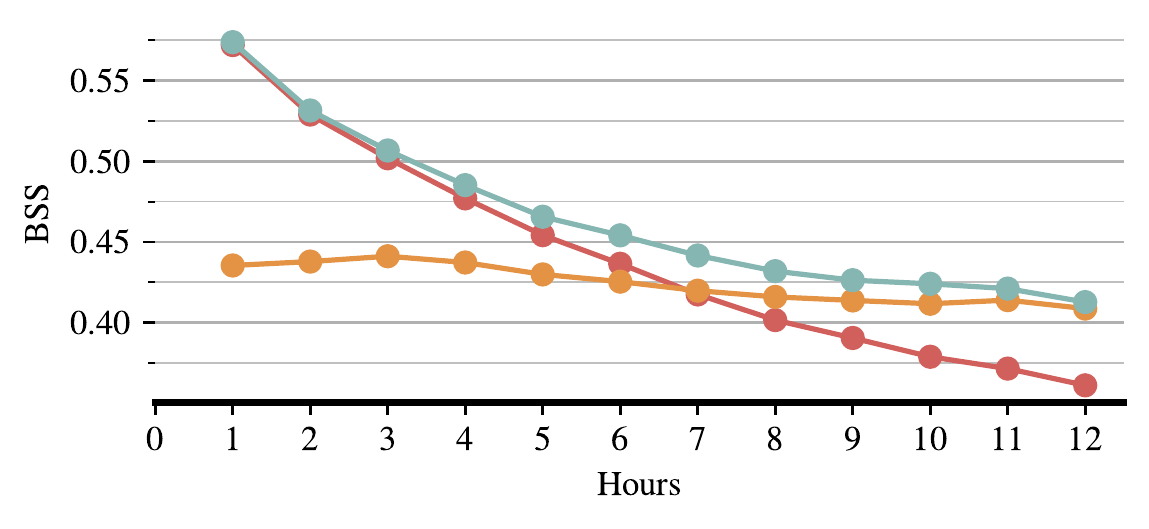}
      \caption{2 mm}
    \end{subfigure}
    \begin{subfigure}{.48\textwidth}
      \centering
      \includegraphics[width=1\linewidth]{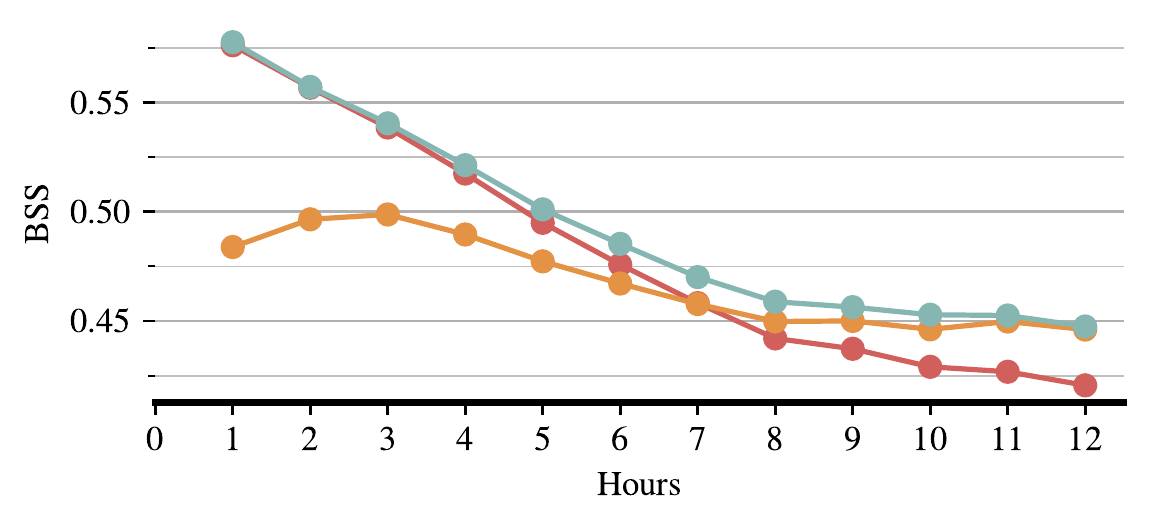}
      \caption{4 mm}
    \end{subfigure}
    
    \begin{subfigure}{.48\textwidth}
      \centering
      \includegraphics[width=1\linewidth]{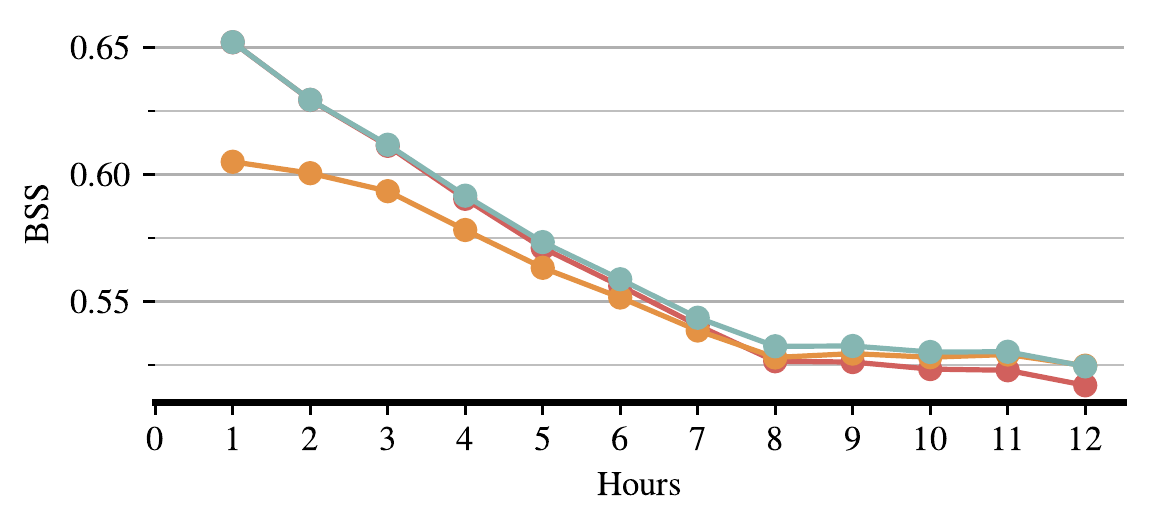}
      \caption{8 mm}
    \end{subfigure}
    \begin{subfigure}{.48\textwidth}
      \centering
      \includegraphics[width=1\linewidth]{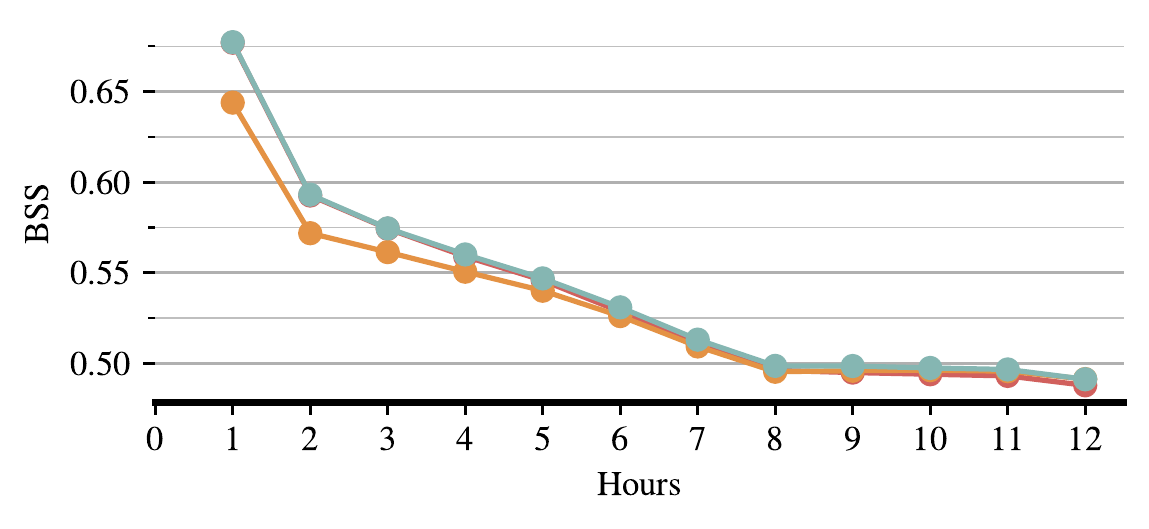}
      \caption{20 mm}
    \end{subfigure}
    \includegraphics[scale=.6]{plots/rate_large_test_legend_horizontal_BSS.pdf}

    \caption{Brier Skill Score relative to NWP of MetNet-2 and variants for various instantaneous precipitation rates of $\geq$.2 mm/hr, $\geq$1 mm/hr, $\geq$2 mm/hr, $\geq$4 mm/hr, $\geq$8 mm/hr and $\geq$20 mm/hr. NWP's predictions are assigned probability 1.}
    \label{fig:appencix_bss_rate}
\end{figure}

\begin{figure}
    \centering
    \begin{subfigure}[b]{0.48\textwidth}
      \centering
      \includegraphics[width=1\linewidth]{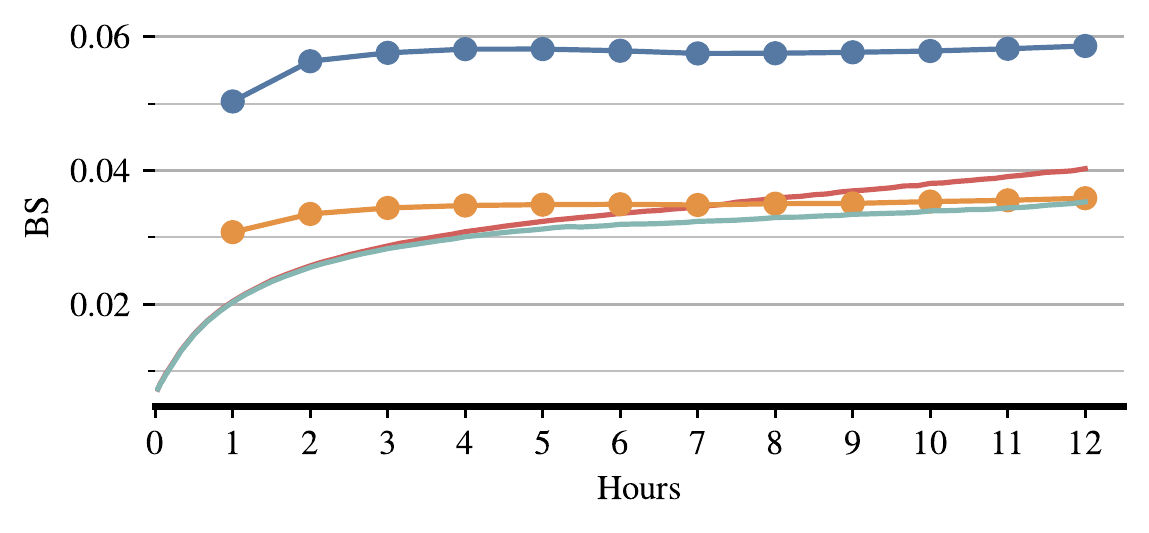}
      \caption{.2 mm}
    \end{subfigure}
    \begin{subfigure}[b]{0.48\textwidth}
      \centering
      \includegraphics[width=1\linewidth]{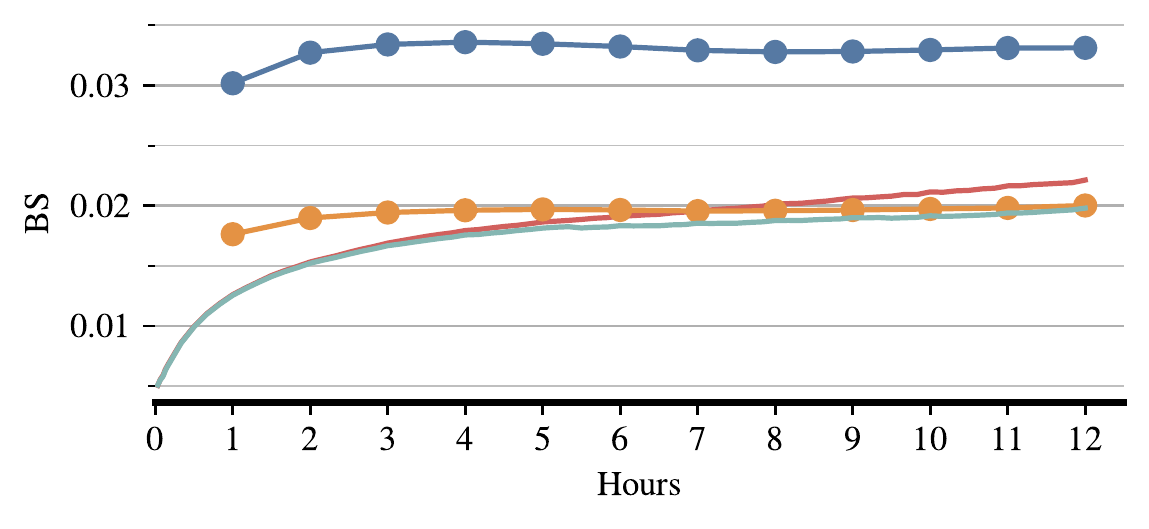}
      \caption{1 mm}
    \end{subfigure}

    \begin{subfigure}{.48\textwidth}
      \centering
      \includegraphics[width=1\linewidth]{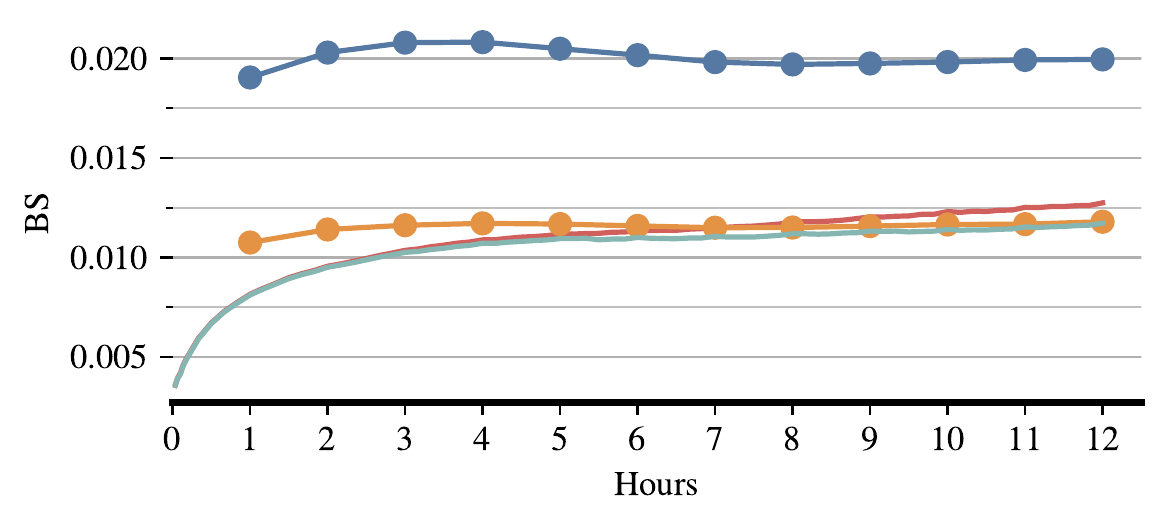}
      \caption{2 mm}
    \end{subfigure}
    \begin{subfigure}{.48\textwidth}
      \centering
      \includegraphics[width=1\linewidth]{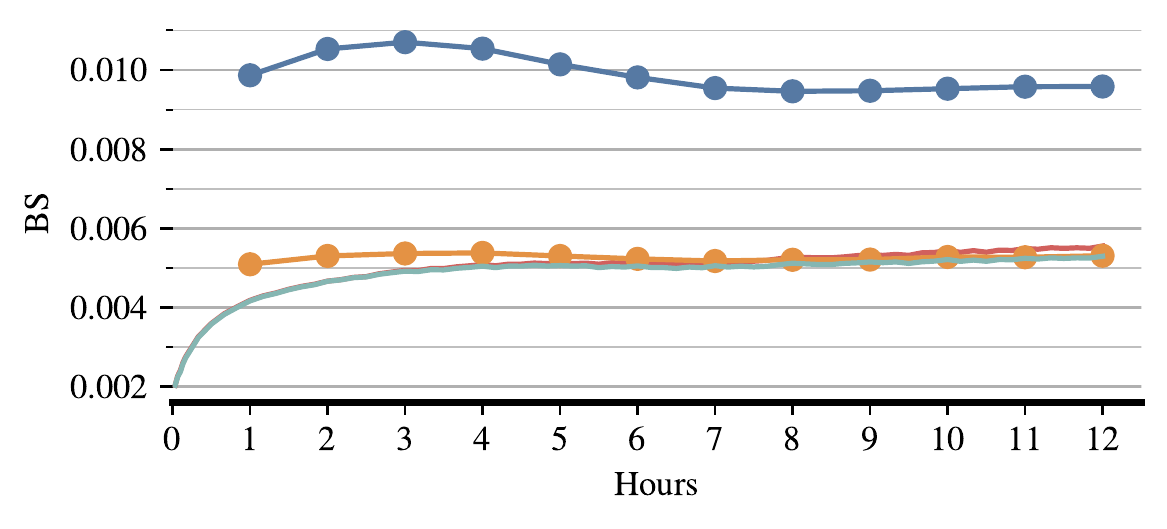}
      \caption{4 mm}
    \end{subfigure}
    
    \begin{subfigure}{.48\textwidth}
      \centering
      \includegraphics[width=1\linewidth]{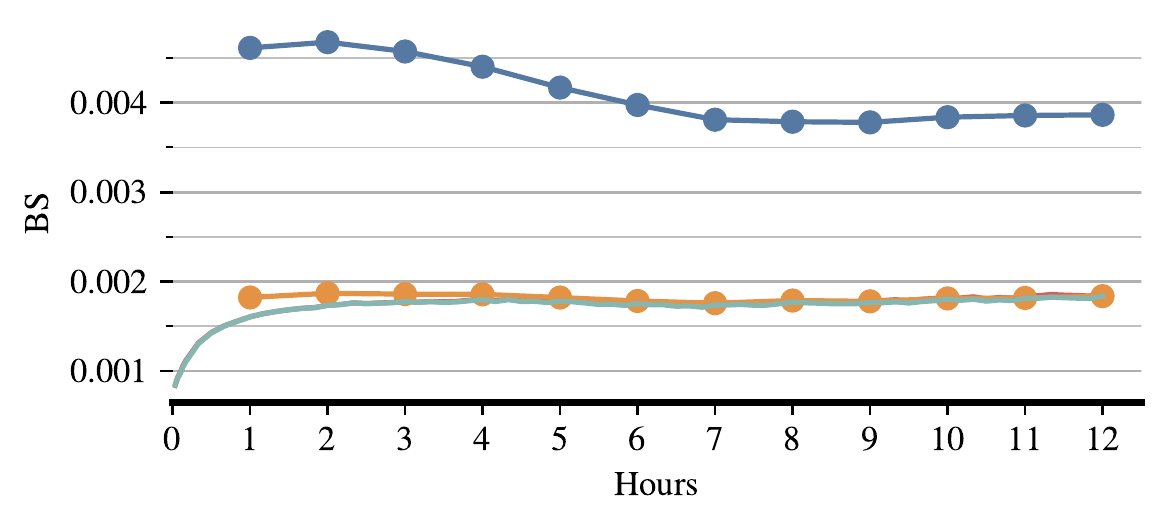}
      \caption{8 mm}
    \end{subfigure}
    \begin{subfigure}{.48\textwidth}
      \centering
      \includegraphics[width=1\linewidth]{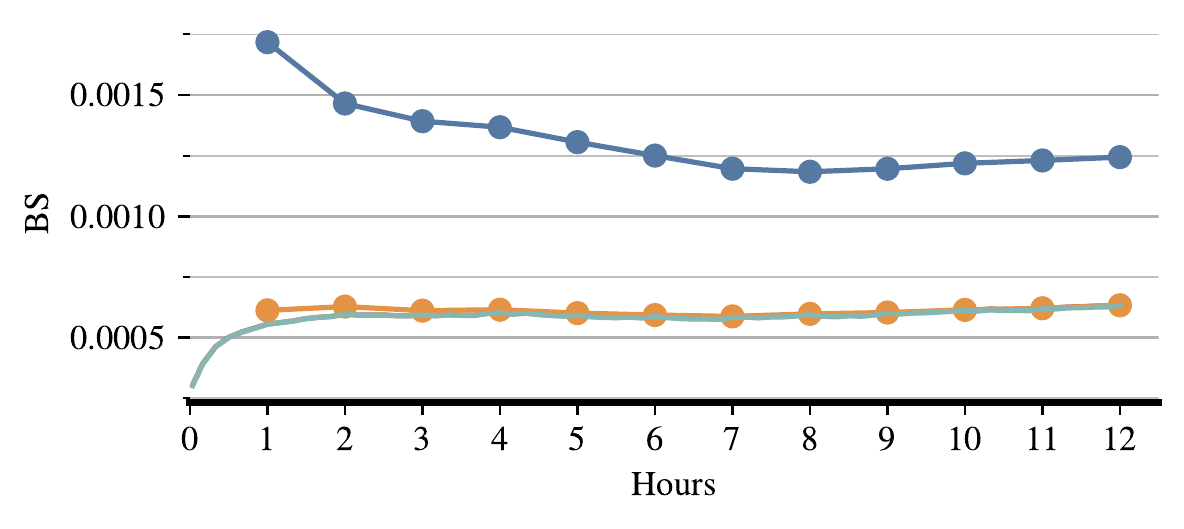}
      \caption{20 mm}
    \end{subfigure}
    \includegraphics[scale=.6]{plots/rate_large_test_legend_horizontal_CSI.pdf}
    \caption{Brier Score of NWP, MetNet-2 and variants for various instantaneous precipitation rates of $\geq$.2 mm/hr, $\geq$1 mm/hr, $\geq$2 mm/hr, $\geq$4 mm/hr, $\geq$8 mm/hr and $\geq$20 mm/hr. NWP's predictions are assigned probability 1.}
    \label{fig:appendix_bs_rate}
\end{figure}

\begin{figure}
    \centering
    \begin{subfigure}[b]{0.48\textwidth}
      \centering
      \includegraphics[width=1\linewidth]{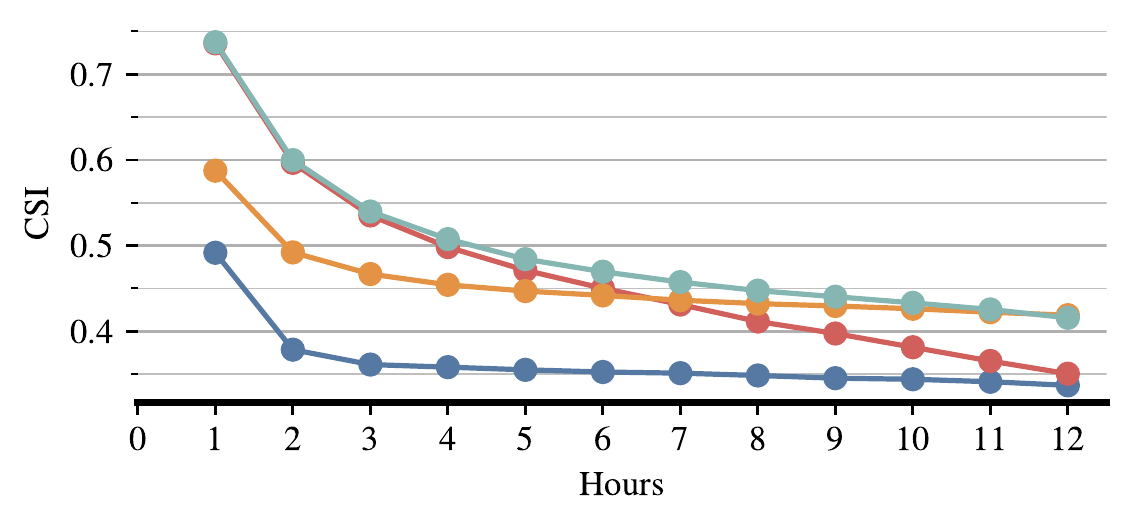}
      \caption{.2 mm}
    \end{subfigure}
    \begin{subfigure}[b]{0.48\textwidth}
      \centering
      \includegraphics[width=1\linewidth]{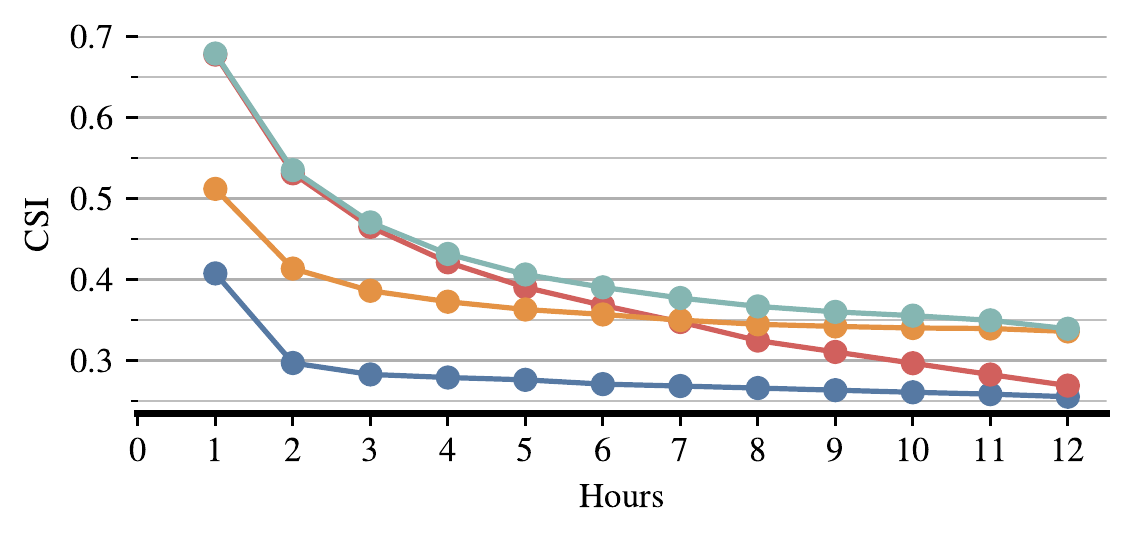}
      \caption{1 mm}
    \end{subfigure}

    \begin{subfigure}{.48\textwidth}
      \centering
      \includegraphics[width=1\linewidth]{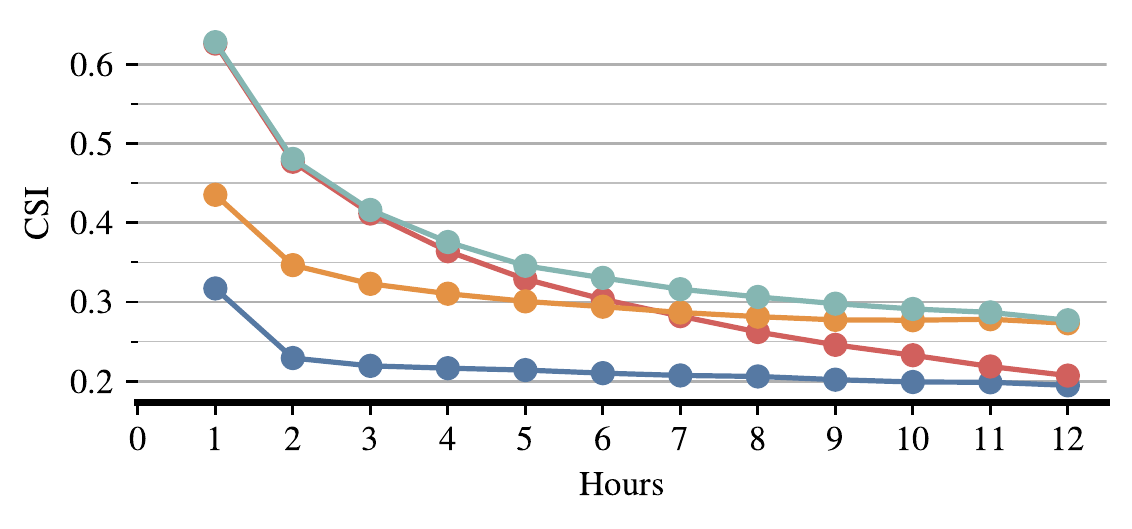}
      \caption{2 mm}
    \end{subfigure}
    \begin{subfigure}{.48\textwidth}
      \centering
      \includegraphics[width=1\linewidth]{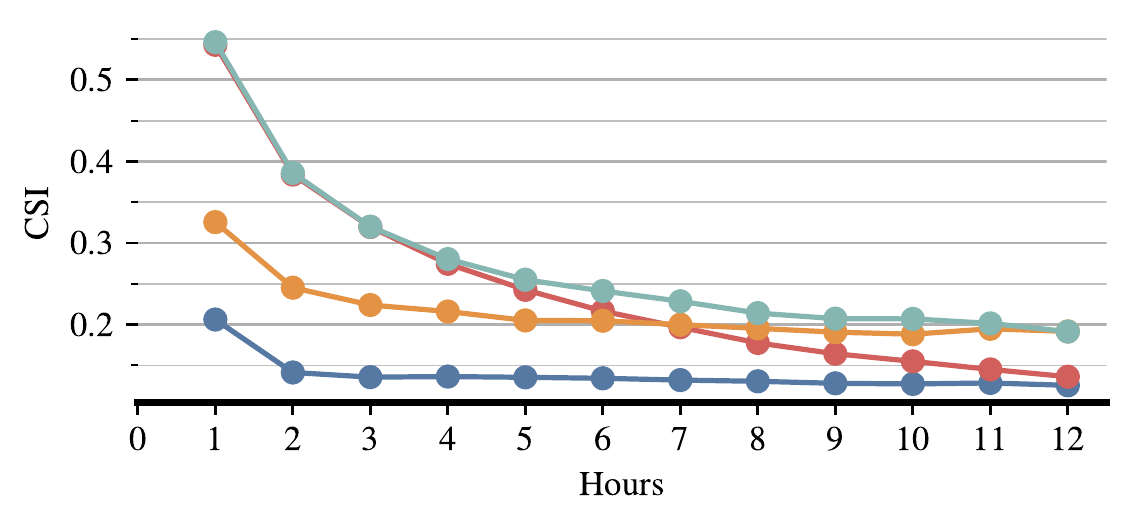}
      \caption{4 mm}
    \end{subfigure}
    
    \begin{subfigure}{.48\textwidth}
      \centering
      \includegraphics[width=1\linewidth]{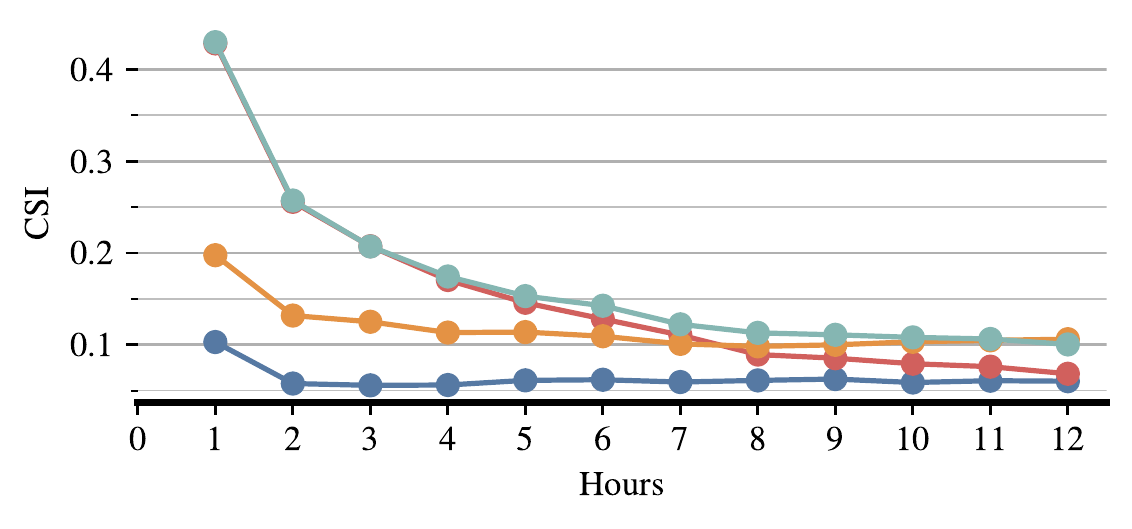}
      \caption{8 mm}
    \end{subfigure}
    \begin{subfigure}{.48\textwidth}
      \centering
      \includegraphics[width=1\linewidth]{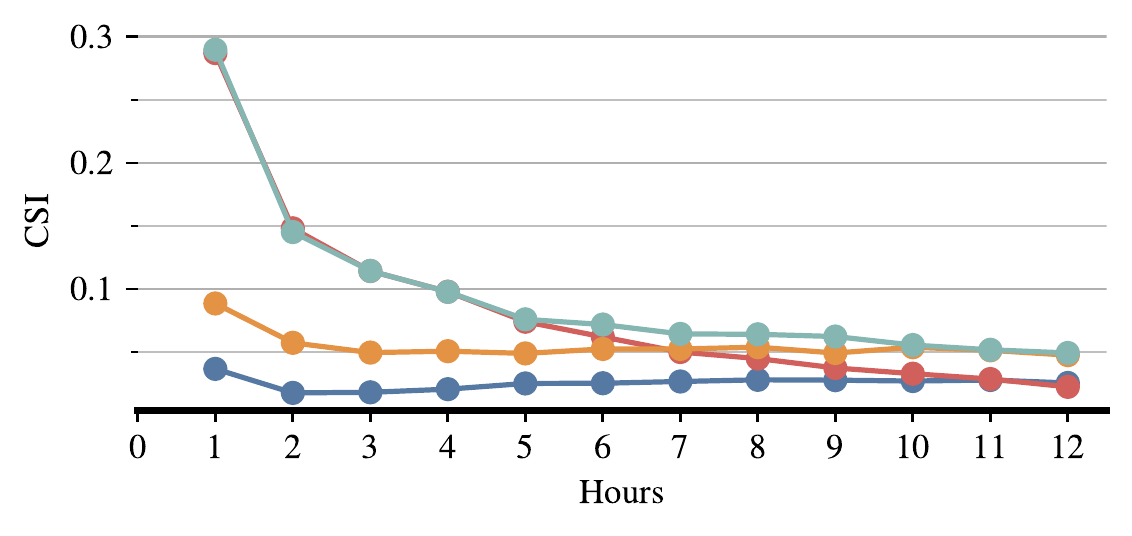}
      \caption{20 mm}
    \end{subfigure}
    \includegraphics[scale=.6]{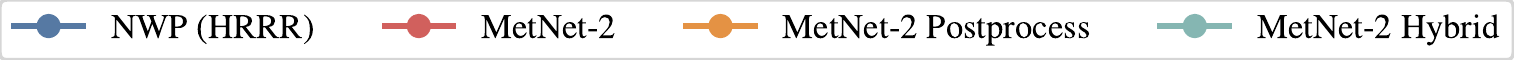}
    
    \caption{CSI performance of NWP, MetNet-2 and variants for hourly cumulative precipitation at rates of $\geq$.2 mm/hr, $\geq$1 mm/hr,$\geq$2 mm/hr,$\geq$4 mm/hr, $\geq$8 mm/hr and $\geq$20 mm/hr.}
    \label{fig:appendix_csi_cumulative}
\end{figure}

\begin{figure}
    \centering
    \begin{minipage}{.6\textwidth}
    \centering
    \includegraphics[width=1\textwidth]{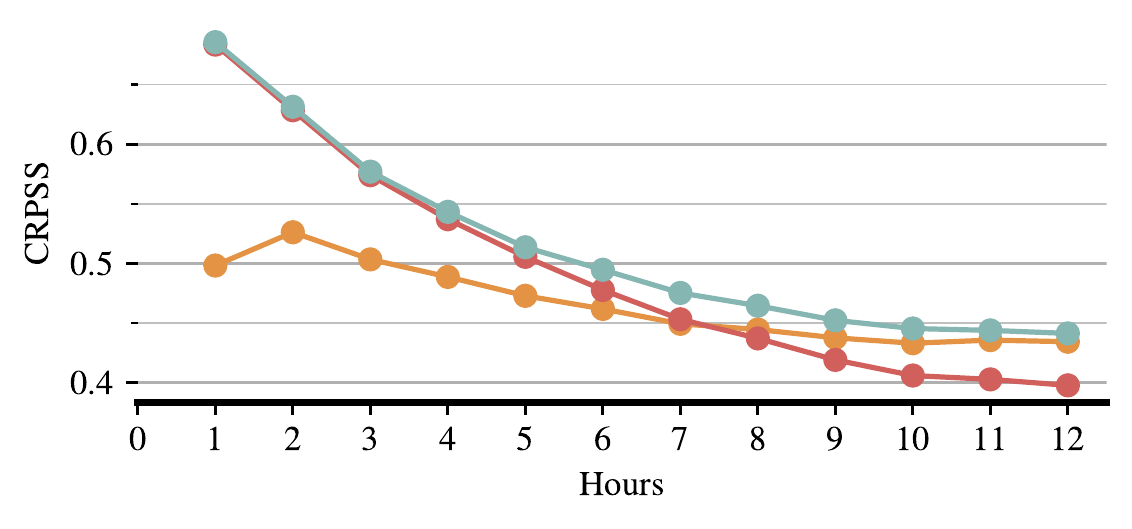}\\
    \includegraphics[scale=.6]{plots/rate_large_test_legend_horizontal_BSS.pdf}
    \end{minipage}%
    \begin{minipage}[b]{.38\textwidth}
    \caption{Continuous Ranked Probability Score Skill for hourly cumulative precipitation. The score tracks the relative improvement of MetNet-2, MetNet-2 Postprocess and MetNet-2 Hybrid over NWP (HRRR).}
    \label{fig:cumulative_crpss}
    \end{minipage}
\end{figure}

The results are summarized as follows. Table~\ref{tbl:main_rate_2mm} in the main text gives numerical values for CSI score for instantaneous rate $\geq$2 mm/hr. The main text also contains Figure~\ref{fig:main_csi_rate} with full graphs for instantaneous rates of $\geq$2 mm/hr, $\geq$8 mm/hr and $\geq$20 mm/hr and Figure~\ref{fig:main_crpss_a} gives the CRPS skill score at once for all levels of precipitation for the instantaneous measure. In this supplement, CSI, Brier Score and BSS for the instantaneous measure for additional levels of precipitation appear in Figure~\ref{fig:appendix_csi_rate}, Figure~\ref{fig:appendix_bs_rate} and Figure~\ref{fig:appencix_bss_rate}, respectively. For the hourly cumulative measure, the CSI score for additional levels of precipitation appears in~Figure~\ref{fig:appendix_csi_cumulative} and CRPS skill in Figure~\ref{fig:cumulative_crpss}. Note the difference in evaluation frequency for the various models.

\begin{figure}
\centering
\begingroup
\newcommand{\caseprefix}{video_example2_1mm_rate_54eb14dc00000000}
\newcommand{\casetimestamp}{1546516800}
\newcommand{\casedir}{\caseprefix _all_rates_3d}
\setlength\tabcolsep{1.5pt}
\begin{tabular}{@{}c c c c c r@{}}
  & 1 hr & 3 hr & 6 hr & 12 hr & \\ 
  \vtitle{Ground Truth}
      & \predlarge{MRMS}{00060}
      & \predlarge{MRMS}{00180}
      & \predlarge{MRMS}{00360}
      & \predlarge{MRMS}{00720}\myrowspace
      & \raisebox{-0.5\height}{\includegraphics[scale=.5]{case_studies/\casedir/colorbar_fixed.pdf}~\rotatebox[origin=l]{90}{\hspace{.4cm}\smallb{Precipitation (mm)}}} \\
  \vtitle{\mt{}}
      & \predlarge{MetNet2}{00060}
      & \predlarge{MetNet2}{00180}
      & \predlarge{MetNet2}{00360}
      & \predlarge{MetNet2}{00720}\myrowspace
      & \raisebox{-0.5\height}{\shortstack[r]{\includegraphics[scale=.5]{case_studies/\casedir/colorbar_probs_8.0.pdf}\\\includegraphics[scale=.5]{case_studies/\casedir/colorbar_probs_4.0.pdf}\\\includegraphics[scale=.5]{case_studies/\casedir/colorbar_probs_2.0.pdf}\\\includegraphics[scale=.5]{case_studies/\casedir/colorbar_probs_1.0.pdf}\\\includegraphics[scale=.5]{case_studies/\casedir/colorbar_probs_0.2.pdf}}~\rotatebox[origin=l]{90}{\hspace{-.35cm}\smallb{Prob. of Precipitation (mm)}}} \\
  \vtitle{\mt{} Postprocess}
      & \predlarge{PP-HRRR}{00060}
      & \predlarge{PP-HRRR}{00180}
      & \predlarge{PP-HRRR}{00360}
      & \predlarge{PP-HRRR}{00720}\myrowspace
      &  \\
  \vtitle{\mt{} Hybrid}
      & \predlarge{MetNet2_Hybrid}{00060}
      & \predlarge{MetNet2_Hybrid}{00180}
      & \predlarge{MetNet2_Hybrid}{00360}
      & \predlarge{MetNet2_Hybrid}{00720}\myrowspace
      &  \\
\end{tabular}
\caption{Case study for Thu Jan 03 2019 12:00 UTC of the North West coast of the US with forecasts of instantaneous precipitation.}
\label{fig:main_case_a_app}
\endgroup
\end{figure}

\begin{figure}
\centering
\begingroup

\newcommand{\caseprefix}{video_hurricane_isaias_1mm_cumulative_89c80adc00000000}
\newcommand{\casetimestamp}{1596484800}
\newcommand{\casedir}{\caseprefix _all_rates_3d}
\setlength\tabcolsep{1.5pt}
\begin{tabular}{@{}c c c c c r@{}}
  & 1 hr & 3 hr & 6 hr & 12 hr & \\ 
  \vtitle{Ground Truth}
      & \predlarge{MRMS}{00060}
      & \predlarge{MRMS}{00180}
      & \predlarge{MRMS}{00360}
      & \predlarge{MRMS}{00720}\myrowspace
      & \raisebox{-0.5\height}{\includegraphics[scale=.5]{case_studies/\casedir/colorbar_fixed.pdf}~\rotatebox[origin=l]{90}{\hspace{.4cm}\smallb{Precipitation (mm)}}} \\
  \vtitle{\mt{}}
      & \predlarge{MetNet2}{00060}
      & \predlarge{MetNet2}{00180}
      & \predlarge{MetNet2}{00360}
      & \predlarge{MetNet2}{00720}\myrowspace
      & \raisebox{-0.5\height}{\shortstack[r]{\includegraphics[scale=.5]{case_studies/\casedir/colorbar_probs_8.0.pdf}\\\includegraphics[scale=.5]{case_studies/\casedir/colorbar_probs_4.0.pdf}\\\includegraphics[scale=.5]{case_studies/\casedir/colorbar_probs_2.0.pdf}\\\includegraphics[scale=.5]{case_studies/\casedir/colorbar_probs_1.0.pdf}\\\includegraphics[scale=.5]{case_studies/\casedir/colorbar_probs_0.2.pdf}}~\rotatebox[origin=l]{90}{\hspace{-.35cm}\smallb{Prob. of Precipitation (mm)}}} \\
  \vtitle{\mt{} Postprocess}
      & \predlarge{PP-HRRR}{00060}
      & \predlarge{PP-HRRR}{00180}
      & \predlarge{PP-HRRR}{00360}
      & \predlarge{PP-HRRR}{00720}\myrowspace
      &  \\
  \vtitle{\mt{} Hybrid}
      & \predlarge{MetNet2_Hybrid}{00060}
      & \predlarge{MetNet2_Hybrid}{00180}
      & \predlarge{MetNet2_Hybrid}{00360}
      & \predlarge{MetNet2_Hybrid}{00720}\myrowspace
      &  \\
\end{tabular}
\caption{Case study of Hurricane Isaias, a Category 1 hurricane, that caused widespread destruction and economic damage. The forecast time is Mon Aug 03 2020 20:00 UTC on the East coast of the United States. The measure is the gauge-corrected hourly cumulative precipitation.}
\label{fig:main_case_b_app}
\endgroup
\end{figure}

\subsection{NWP (HRRR)}

Two prominent features of HRRR's results is the relatively low performance for the first 2 hours of lead time and the positive flattening of the performance for the lead hours after that. HRRR's non-probabilistic forecasts (see Figures~\ref{fig:main_case_a}~and~\ref{fig:main_case_b}) aim at capturing large scale weather structure that may not be optimized for  making forecasts for a specific location and time $x,y,t$. HRRR's results on hourly rate forecasts paint a similar picture. The hourly cumulative precipitation estimates are calibrated with measurements from precipitation gauges making them less susceptible to biases directly linked to ground radar data. But this does not significantly affect the broad trend in HRRR's results, with a relatively low initial performance that flattens out over longer lead times into a stronger performance.

\subsection{\mt{}} 

Since it does not perform explicit atmospheric simulation, \mt{} needs to learn to approximate the underlying physics in order to make a good forecast. At longer lead times one requires a larger spatial context at the input patch and \mt{} must be able to capture that large context effectively. We can see the performance that \mt{} reaches in the tables of results. For instantaneous rate, \mt{}'s performance is substantially better than HRRR's performance throughout the 12 hour range of lead time. This does not include that additional 60 minutes of delay that the NWP model HRRR incurs due to atmospheric simulation compared to MetNet-2. 
The relative performances of the NWP model and MetNet-2 remain similar across both at low, medium and high instantaneous rates of precipitation up to 20~mm/hr, suggesting that \mt{} can learn to forecast also the much rarer levels for which data is much scarcer. Figure~\ref{fig:appendix_csi_cumulative} also shows the CSI scores for the hourly cumulative precipitation where \mt{}'s performance also exceeds that of HRRR over the 12 hour evaluation range.

\subsection{\mt{} Postprocess}

MetNet-2 Postprocess's performance steadily improves on that of NWP across the full range. 
This results in CSI scores that are up to 50~$\%$ higher than those of the NWP and much better BS and CRPS. Despite this, MetNet-2's performance is still substantially better than MetNet-2 Postprocess' performance for the first seven hours of lead time; after the seven hours mark, MetNet-2 Postprocess performance tends higher than that of MetNet-2. MetNet-2 Postprocess requires however both running the atmospheric simulation as well as training the  neural model.

\subsection{\mt{} Hybrid} 

\mt{} learns to combine all the benefits of the atmospheric simulation with those of the neural computation and this shows in the results. MetNet-2 Hybrid  matches the performance of MetNet-2 up to approximately three hours of lead time and then gradually exceeds it over the remaining range. Remarkably, the Hybrid variant exceeds also \mt{} Postprocess all the way up to twelve hours of lead time. This shows the ability of the neural computation to contribute additional useful information to a precipitation forecast of a very long lead time of twelve hours. Both the CSI and the Brier scores reflect this behavior. For the hourly cumulative precipitation, the results paint a similar performance picture too: \mt{} Hybrid matches or exceeds all other variants on the full range, making the most of both the neurally learnt  forecast and NWP's own simulation.

\newtcolorbox{scorebox_2}[1][]{
    width=70.5pt,
    height=20.5pt,
    arc=0,
    boxsep=0cm,
    toprule=0pt,
    leftrule=0pt,
    bottomrule=0pt,
    rightrule=0pt,
    colframe=gray,
    breakable,
    nobeforeafter,
    left=2pt,
    right=0pt,
    top=1pt,
    bottom=0pt,
    enhanced jigsaw,
    opacityframe=0.5,
    opacityback=0.8
}
\renewcommand{\predfull}[3] {%
  \raisebox{-0.5\height}{%
    \includegraphics[scale=.1]{case_studies/#1_#2_#3.png}}}
\renewcommand{\predinlinefull}[6] {
\raisebox{-0.5\height}{%
\begin{overpic}[scale=.14]{case_studies/#1_#2_#3.png}
    \put(0,#6){%
    \begin{scorebox_2}[]%
    \scalebox{0.5}{%
        \begingroup%
        \setlength\tabcolsep{5pt}%
        \setsepchar{ }%
        \readlist\score{#5}%
        \fontfamily{phv}\selectfont 
        \begin{tabular}{@{}cccccc@{}}%
            & .2 & 1 & 2 & 4 & 8 \\
         #4 & \score[1] & \score[2] & \score[3] & \score[4] & \score[5] \\
        \end{tabular}%
        \endgroup%
    }%
    \end{scorebox_2}
    }%
\end{overpic}%
}}
\renewcommand{\inlinebottom}[0]{0}
\renewcommand{\inlinetop}[0]{81.5}
\renewcommand{\myrowspace}{\vspace{3pt}}%
\renewcommand{\myendspace}{\vspace{10pt}}%
\renewcommand{\vtitle}[1]{\rotatebox[origin=c]{90}{\hspace{-.2cm}\scalebox{1}{#1}}}
\renewcommand{\pred}[2] {\predfull{\casedir/\casetimestamp}{#1}{#2}}
\renewcommand{\predinline}[5]{\predinlinefull{\casedir/\casetimestamp}{#1}{#2}{#3}{#4}{#5}}

\renewcommand{\predfulllarge}[3] {%
  \raisebox{-0.5\height}{%
    \includegraphics[scale=.14]{case_studies/#1_#2_#3.png}}}  
\renewcommand{\predlarge}[2] {\predfulllarge{\casedir/\casetimestamp}{#1}{#2}}

\section{Supplement: Ablations}
\label{sup:ablations}

We provide a series of ablation studies to shed light on which aspects of MetNet-2's design drive its performance.

\subsection{Spatial Context} 
\label{sup:abl_context}

\begin{figure}[H]
    \centering
    \begin{subfigure}[b]{0.48\textwidth}
      \centering
      \includegraphics[width=1\linewidth]{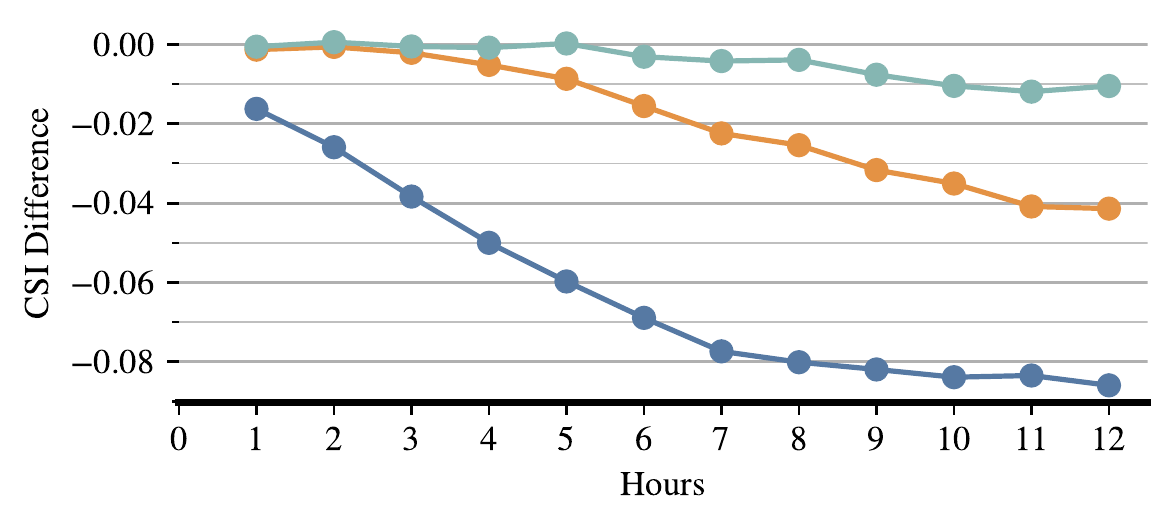}
      \caption{.2 mm}
    \end{subfigure}
    \begin{subfigure}[b]{.48\textwidth}
      \centering
      \includegraphics[width=1\linewidth]{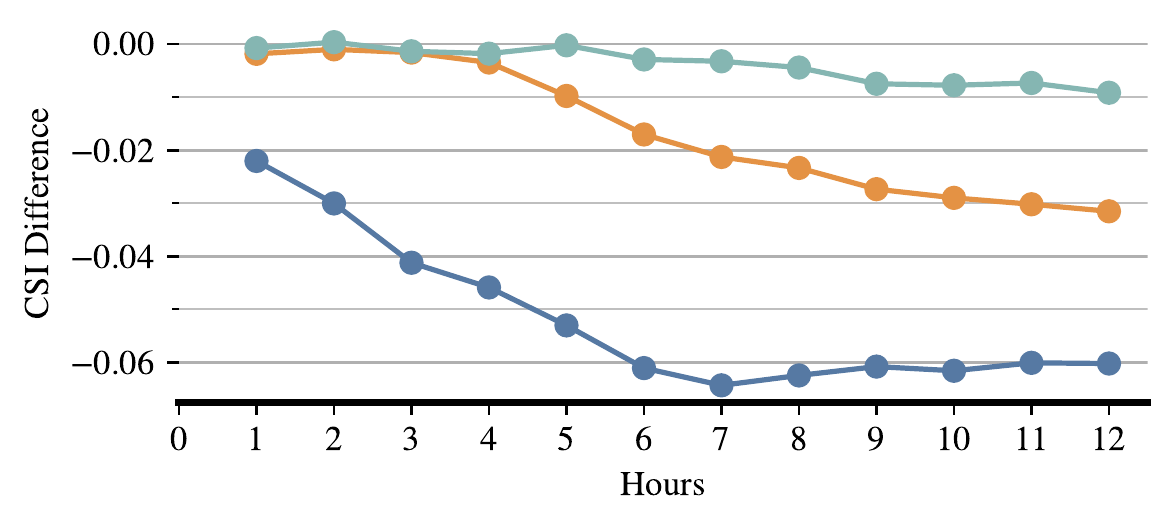}
      \caption{2 mm}
    \end{subfigure}
    \centering
    \includegraphics[scale=.6]{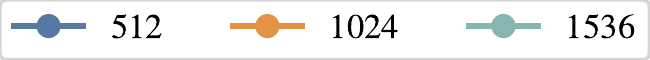}
    \caption{Ablation experiments for the size of the input spatial context to MetNet-2, computed on instantaneous precipitation. Smaller context size hurts performance especially at later hours.}
    \label{fig:ablation_spatial}
\end{figure}

A direct way of measuring the effect that context size has on forecast performance is by limiting directly the context size at the network's input. We do so for \mt{} by reducing the spatial context size from the default 2048~km~$\times$~2048~km centered around the target patch of 512~km~$\times$~512~km to 1536~km~$\times$~1536~km, 1024~km~$\times$~1024~km and 512~km~$\times$~512~km. Figure~\ref{fig:ablation_spatial} reports the results as a difference from the full context. We can see that reducing spatial input context incurs a marked performance drop that increases with the hours of lead time. The drop at 512~km $\times$ 512 km of spatial context already occurs at the earliest hours of lead time, suggesting that the model requires some context beyond the target patch even for short lead times. After 3~-~4~hours, MetNet-2 benefits from a spatial context that is larger than 1024 that corresponds to borders of 256 km around the target patch. After 5~-~6~hours, MetNet-2 benefits from an even larger spatial context of at least 1536 km that corresponds to 512 km around the target patch.

\subsection{Subsets of Assimilated Variables}
\label{sup:abl_subsets}

\begin{figure}[H]
    \centering
    \begin{subfigure}[b]{0.48\textwidth}
      \centering
      \includegraphics[width=1\linewidth]{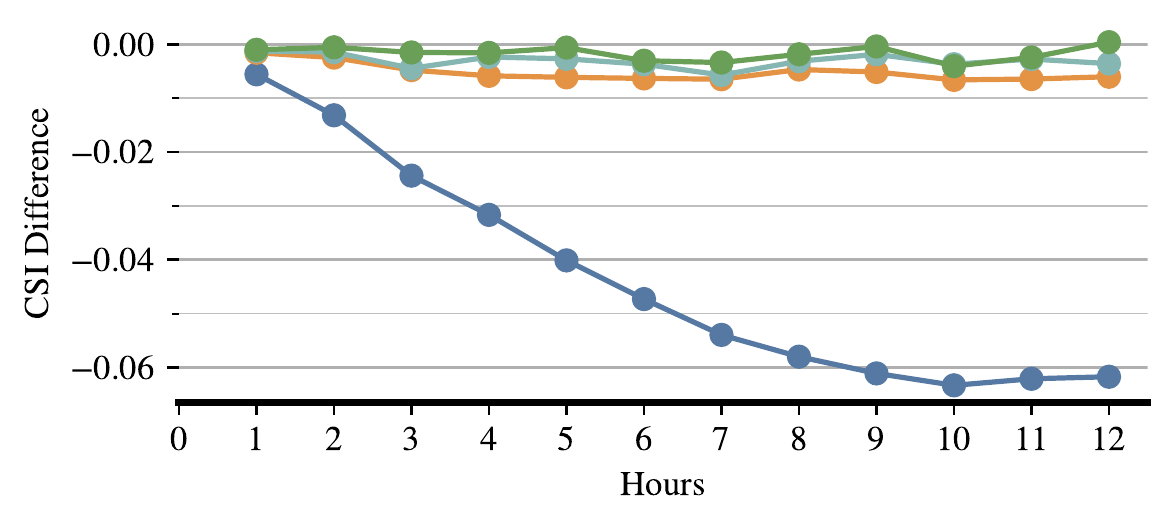}
      \caption{.2 mm}
    \end{subfigure}
    \begin{subfigure}[b]{.48\textwidth}
      \centering
      \includegraphics[width=1\linewidth]{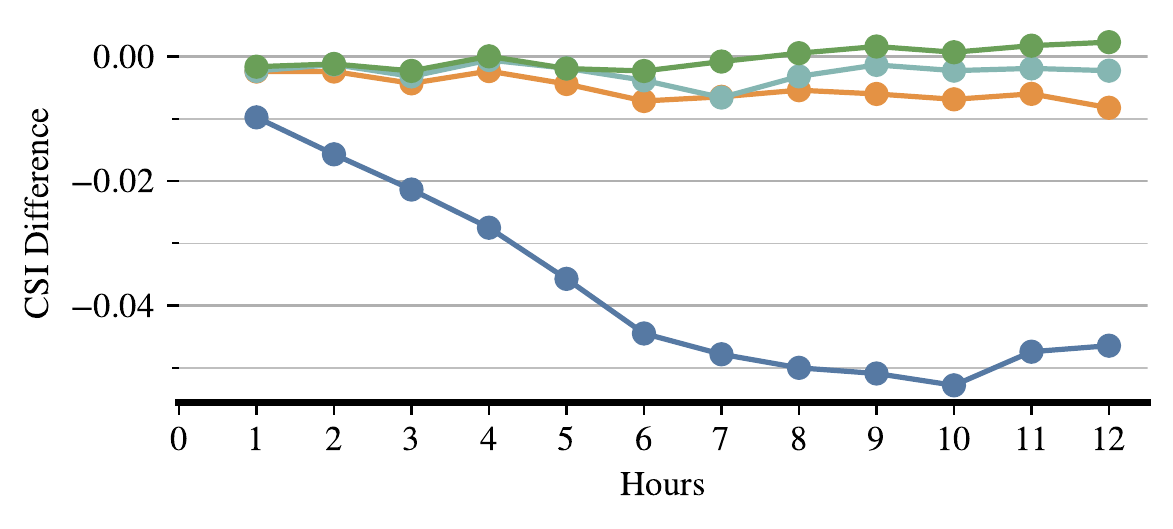}
      \caption{2 mm}
    \end{subfigure}
    \centering
    \includegraphics[scale=.6]{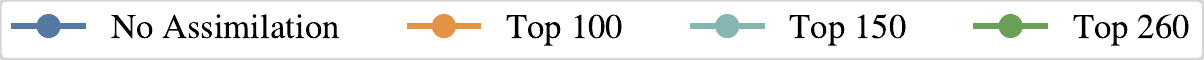}

    \caption{CSI score difference relative to MetNet-2 when using only subsets of the 612 Assimilation features: the top 260 features, the top 150 features, the top 100 features and none of the features (No Assimilation). Results on instantaneous precipitation.}
    \label{fig:ablation_hrrr}
\end{figure}

Besides ablating spatial context, we also run experiments by only using subsets of the Assimilation variables. The interpretability analysis (see Section~\ref{sup:interpretation}) allows us to approximately identify the importance of each variable and rank them accordingly. We then keep the top 260, top 150 or top 100 variables from the Assimilation state and evaluate their contributions. We also compare to the case of using none of the variables. Figure~\ref{fig:ablation_hrrr} reports the results from the ablation experiments. Removing the less important Assimilation inputs degrades performance, again especially at later hours, suggesting the ability of \mt{} to gain useful signal even from less relevant variables. The case of not using any of the Assimilation variables  incurs a significant drop in performance with respect to the default setting. 

\subsection{Comparison with MetNet}

\begin{figure}[H]
    \centering
    \begin{subfigure}[b]{0.48\textwidth}
      \centering
      \includegraphics[width=1\linewidth]{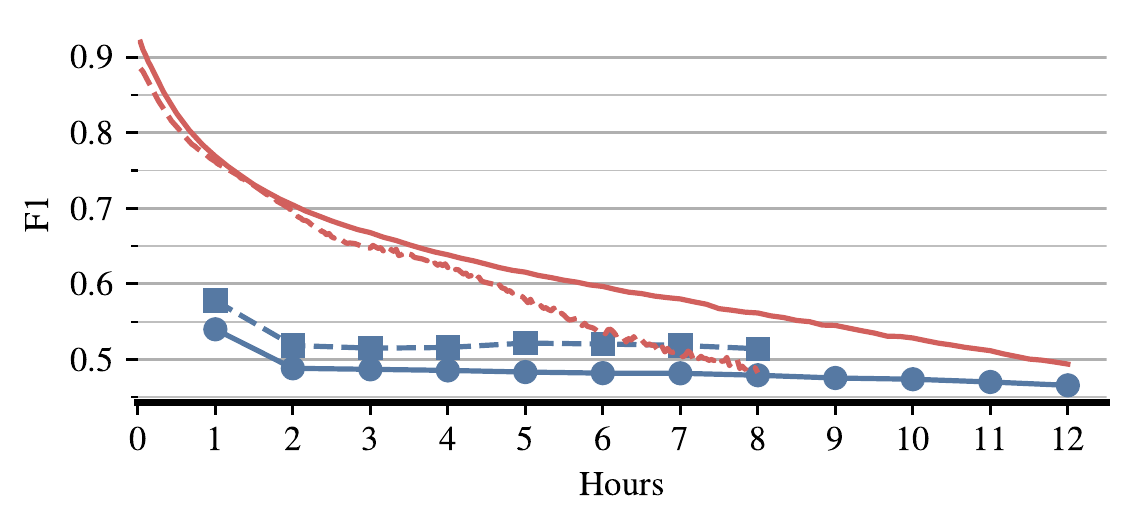}
      \caption{.2 mm}
    \end{subfigure}
    \begin{subfigure}[b]{.48\textwidth}
      \centering
      \includegraphics[width=1\linewidth]{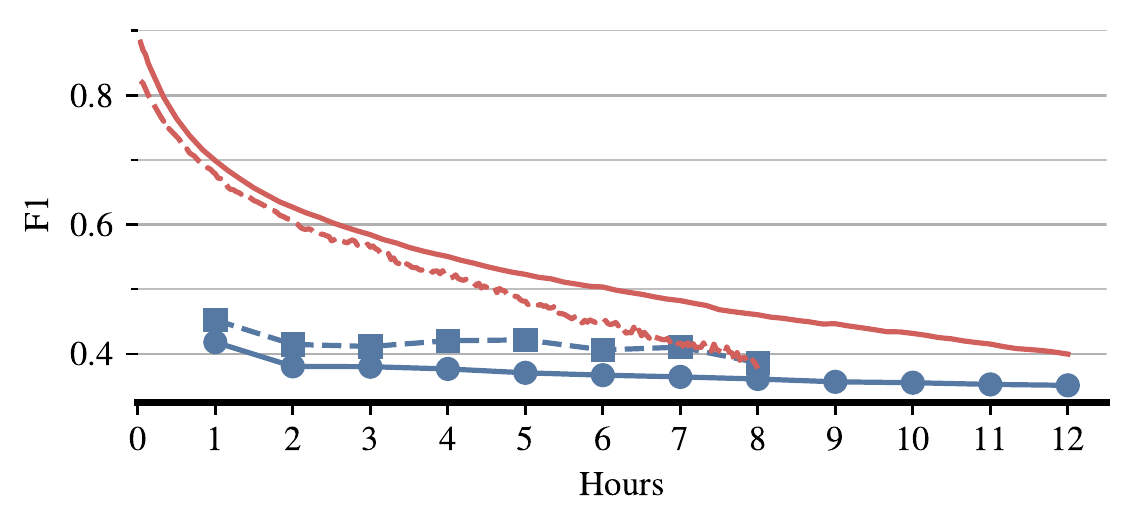}
      \caption{1 mm}
    \end{subfigure}
    \begin{subfigure}[b]{.48\textwidth}
      \centering
      \includegraphics[width=1\linewidth]{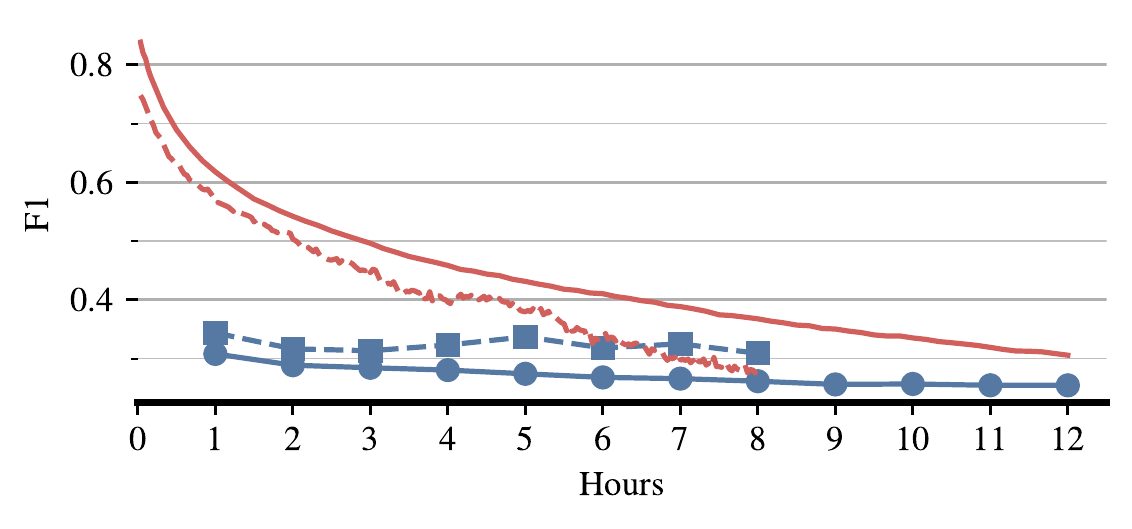}
      \caption{2 mm}
    \end{subfigure}
    \begin{subfigure}[b]{.48\textwidth}
    \centering
    \includegraphics[scale=.6]{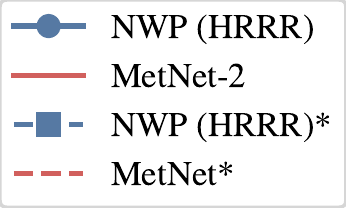}
    \vspace{2cm}
    \end{subfigure}

    \caption{Comparison using similar but distinct test sets of MetNet-2 and MetNet~\citep{sonderby2020metnet} on the originally used F1 metric, that is very similar to CSI. MetNet-2 outperforms MetNet substantially despite the fact that the data that MetNet-2 uses are harder for the NWP HRRR.}
    \label{fig:metnet1}
\end{figure}

We do an approximate comparison of MetNet-2 also with the MetNet model~\cite{sonderby2020metnet}. An exact comparison is difficult due to changes in the projection of the input data that has architectural implications, as well as other differences in the test data. For this reason, we simply plot MetNet-2' and MetNet' performances on the same plot, despite the fact that they are evaluated on slightly different test datasets. MetNet was shown to exceed HRRR's performance up to 7 to 8 hours of lead time. \mt{} without the assimilation variables exceeds HRRR's performance for up to nearly 12 hours of lead time, as can be seen in Figure~\ref{fig:ablation_hrrr}. This extension to 12 hours comes purely from the better architecture of MetNet-2 and an improved training regime. Using the assimilation variables as input, MetNet-2 provides a further improvement and its performance is even higher with respect to that of MetNet across the full range (Figure~\ref{fig:metnet1}). This is despite the fact that the baseline HRRR achieves higher F1 scores on MetNet's test data than on MetNet-2's test data and finds MetNet's test data easier to predict well.

\subsection{Forms of Lead Time Conditioning}
\label{sup:abl_lead_time}

\begin{figure}[H]
    \centering
    \begin{subfigure}[b]{0.48\textwidth}
      \centering
      \includegraphics[width=1\linewidth]{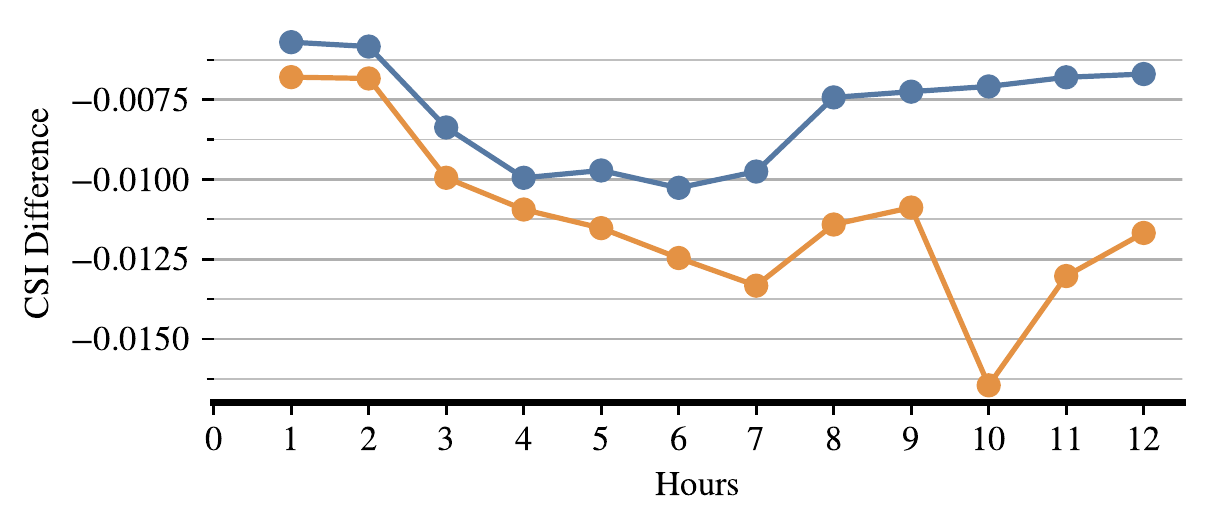}
      \caption{.2 mm}
    \end{subfigure}
    \begin{subfigure}[b]{.48\textwidth}
      \centering
      \includegraphics[width=1\linewidth]{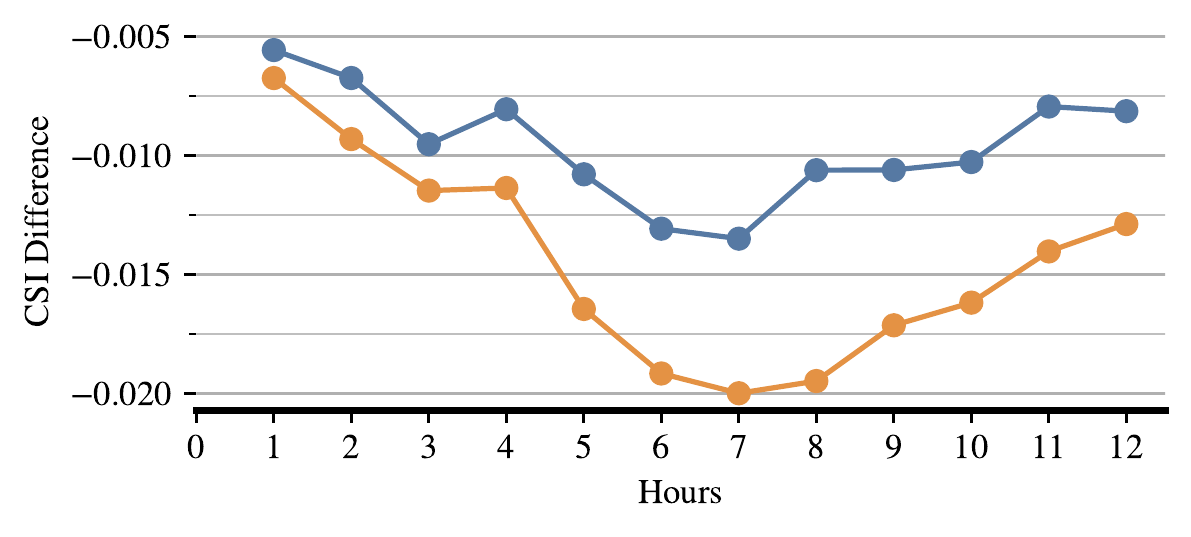}
      \caption{2 mm}
    \end{subfigure}
    \centering
    \includegraphics[scale=.6]{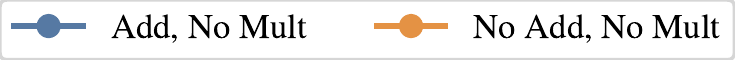}

    \caption{Ablation experiments with the adopted additive and multiplicative conditioning as baseline compared to only an additive variant (Add, No Mult) and the default concatenative variant (No Add, No Mult) used in Sønderby~et~al.~\cite{sonderby2020metnet}. Results on instantaneous precipitation.}
    \label{fig:ablation_target_conditioning}
\end{figure}

The importance of the proposed, richer form of lead time conditioning also shows in the evaluation (Figure~\ref{fig:ablation_target_conditioning}). We compare it with the vanilla form of conditioning of simply concatenating the lead time index at the input \cite{sonderby2020metnet} as well as part of the richer form of conditioning that only includes the additive bias component that shifts the activations in the convolutional layers. The proposed richer form of conditioning significantly improves performance with up to 1.5 CSI points over the default baseline and both the additive and the multiplicative components are important to the final performance.

\subsection{Dilation Factors}
\label{sup:abl_dilation_factor}

\begin{figure}[H]
    \centering
    \begin{subfigure}[b]{0.48\textwidth}
      \centering
      \includegraphics[width=1\linewidth]{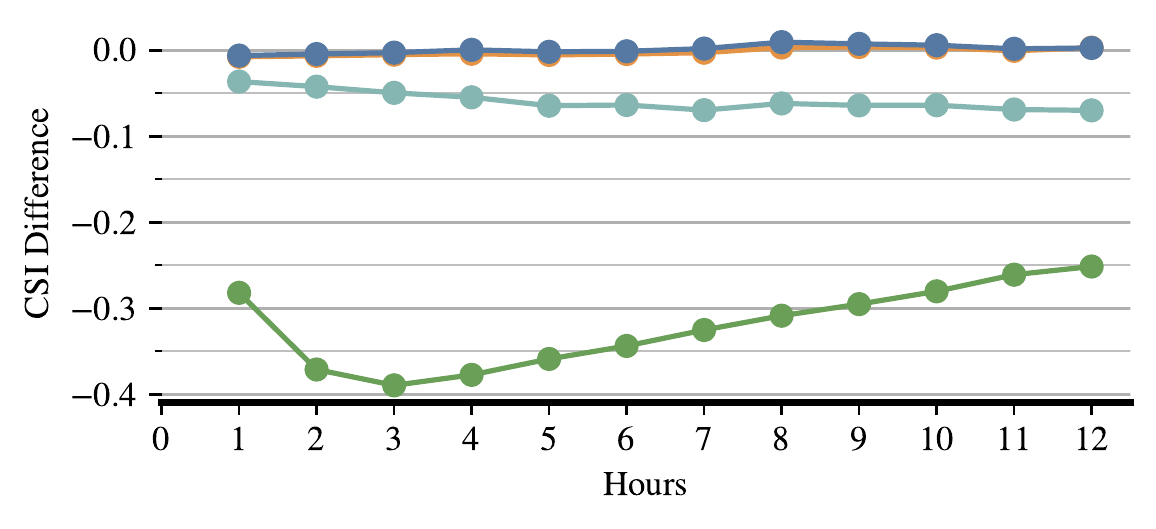}
      \caption{.2 mm}
    \end{subfigure}
    \begin{subfigure}[b]{.48\textwidth}
      \centering
      \includegraphics[width=1\linewidth]{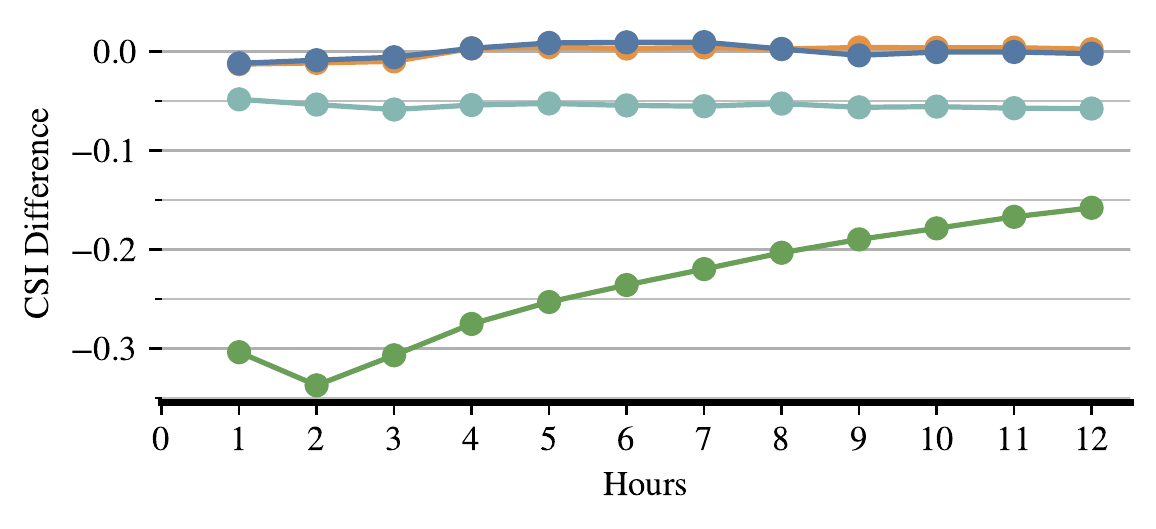}
      \caption{2 mm}
    \end{subfigure}
    \centering
    \includegraphics[scale=.6]{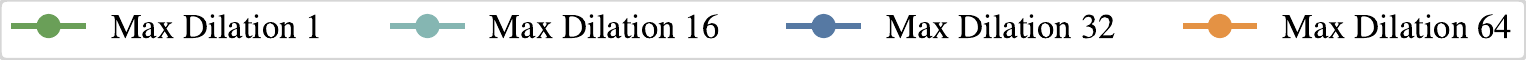}
    \caption{Ablation experiments for the size of the maximal dilation in the convolutions of MetNet-2's encoders. At maximal dilation of 16 one already sees a substantial drop in performance. Total depth of model and number of parameters is kept the same across all experiments.}
    \label{fig:ablation_dilation}
\end{figure}

One important feature of the \mt{} system and architecture is the necessity to process information from a large spatial context. On the one hand, the network's input must correspond to a large spatial context and, on the other, the network's connectivity must ensure that each point on the target patch has access to the portion of the input spatial context that it needs for a good forecast. The key aspect of the architecture that aims at capturing a large context for each of the target patch locations is the pattern of exponentially increasing factors of dilation. In this set of experiments, we ablate the maximal size of these factors, from 128 that is the default to 64, 32 and 16, as well as no dilation at all. Note that the difference between 64 and 128 is just one more residual block of larger dilation, which in the 64 version is replaced by additional layers with smaller dilation. Thus substantial differences in captured context size only appear when the maximum dilation is significantly smaller than 128. Figure~\ref{fig:ablation_dilation} summarizes the respective results for the \mt{} model. We see that the performance of the architecture tends to decrease as the maximum dilation factors get smaller. This is especially marked at later lead times when processing a large spatial context effectively becomes necessary for a good forecast. These results indicate that the unusually large dilation factors rarely used in spatial convolutions succeed at capturing large context effectively and are an important factor in \mt{}'s model performance.

\section{Supplement: Interpretation}
\label{sup:interpretation}

The 100s of weather features used as input into \mt{} make it  important to understand what a physics-free machine learning model, such as \mt{}, is learning. 

We aim to interpret \mt{} in order to explain individual predictions (local interpretability) as well as explaining phenomena learnt by the model globally, i.e. across all samples. Local interpretability helps us answer questions such as which spatial region of an input feature was crucial for forecasting the output. Whereas, global interpretability can help us identify useful minimal subsets of features and their physical interactions.

We obtain attributions using a technique called Integrated Gradients~\citep{sundararajan2017axiomatic} that allows attributing the prediction $f(x)$ to the inputs $x$. We use the minimum value of each input feature as its baseline input $x'$ since at this minimum valued baseline we get a prediction close to zero probability of rain. We compute gradients of each prediction at points along the straightline path from the baseline to the actual input. More formally,
\begin{equation}
    IntegratedGrads(x) ::= (x - x') \times \int_{\alpha=0}^1 \frac{\partial f(x_{\alpha})}{\partial x} d\alpha
\end{equation}

where $x_{\alpha} = x' + \alpha \times (x - x')$, is a point on the straightline path. 

Attribution maps have been typically used for singular classification outputs, and this is be one of the first applications of this technique to a model with a large multi-dimensional output. McGovern et al.~\citep{MakingtheBlackBoxMoreTransparentUnderstandingthePhysicalImplicationsofMachineLearning} discuss various techniques for explaining black box ML models on various weather prediction tasks. However, all of the applications involve a single predicted output for each class. In our case, while we can obtain the attributions for each predicted pixel that represents a $1~\textrm{km} \times 1~\textrm{km}$, it will be humanly impossible to consume and make sense of the attribution maps for each pixel in the output separately. Additionally, the complexity of most of the existing interpretability tools increases greatly if they were to be applied to every single predicted pixel in a multi-dimensional output, separately. We, therefore, use Integrated Gradients to explain not just a single predicted pixel but the entire multi-dimensional output of predictions for the given input. For global interpretability, we aggregate per-feature attributions across space and over all the predictions and samples. This essentially gives us a single attribution value for each feature that represents the relative importance of that feature to the model. 

In simpler terms, an attribution value for any given single input pixel represents how much it contributes to a single pixel probability prediction. The feature importance analysis computes how much on average, across all samples, each input contributes to a single predicted pixel. When doing a per-sample analysis, we compute how much on average, the input contributes to a predicted pixel for that specific sample. The attribution values for each of the features are, therefore, comparable to each other.

\subsection{Analysis of MRMS Radar and GOES Satellites}
\label{sec:interpret_mrms_goes}

\begin{figure}[H]
    \centering
    \begin{subfigure}[b]{0.48\textwidth}
      \centering
      \includegraphics[width=1\textwidth]{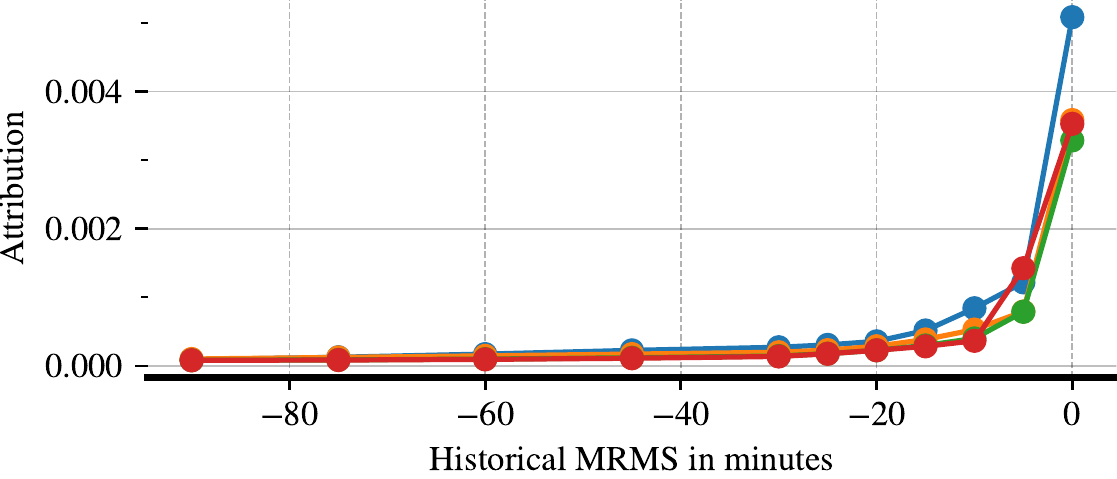}
      \caption{MRMS}
      \label{fig:mrms}
    \end{subfigure}
    \begin{subfigure}[b]{0.48\textwidth}
      \centering
      \includegraphics[width=1\textwidth]{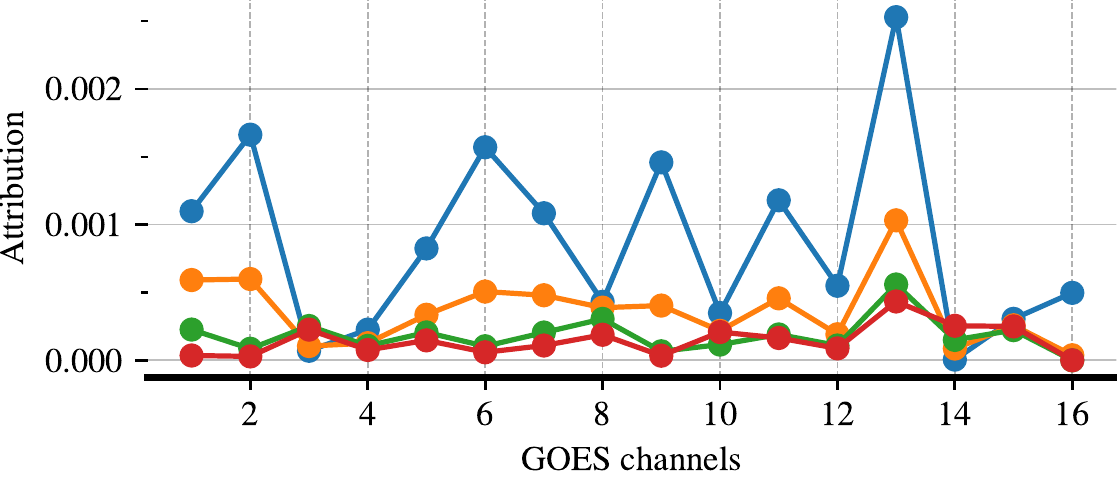}
      \caption{GOES}
      \label{fig:goes}
    \end{subfigure}
    \centering
    \includegraphics[scale=.6]{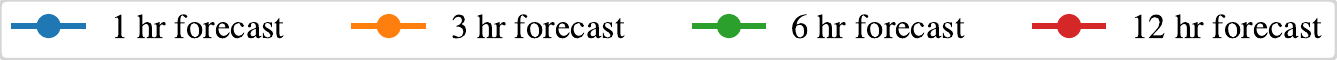}
    \caption{Importance of MRMS and GOES at different forecast hours. The attribution values correspond to how much on average, across all samples, each input pixel contributes to a single predicted pixel.}
    \label{fig:mrmsgoes}
\end{figure}

Figure~\ref{fig:mrmsgoes} shows the attribution of MRMS and GOES historical data for different forecast hours. It is evident that both MRMS and GOES are highly important for nowcasting, i.e., 1-3 hr forecasts, and decrease in importance for the longer forecast hours. In alignment with our expectation, the model learns most from the latest (0th minute) MRMS data, and there is a sharp drop for all other historical data points.

In Figure~\ref{fig:goes}, we see that ABI Band 13 in GOES, the \textit{clean infrared window} 10.3~\textmugreek m band, gets the highest attribution. According to NOAA~\citep{noaa_attribution}, this band is least sensitive to water vapor absorption than other infrared bands. It therefore improves atmospheric moisture corrections, aids in cloud and other atmospheric feature identification/classification. This shows that the model has learnt from features that conforms with our knowledge of what the most important GOES band should be. It is interesting to note that ABI Band 2, the \textit{Red Visible} 0.64~\textmugreek m band, is important for the 1 hour forecast, but rapidly decreases in importance for even the 3 hour forecast, and has no significance to any of the longer hours. This could be due to the very fine resolution (0.5 km) of this band which allows detection of boundaries and small clouds making it useful for nowcasting. In contrast, ABI Band 3, the \textit{Veggie} 0.86 textmugreek m band, has higher attribution for the longer range forecasts than for the 1 hr nowcast, implying that the land characteristics maybe an important feature predicting rain a few hours out. The \textit{Mid-level water vapour} 6.9 textmugreek m band, ABI Band 9, is used for tracking middle tropospheric winds, monitoring severe weather potential, estimating mid-level moisture, and other purposes \citep{schmit_closer_2017}. That the prediction of \mt{} relies on this feature is another sanity check that the model is learning meaningful information.    

\begin{figure}[H]
    \centering
    \begin{subfigure}[b]{0.48\textwidth}
      \centering
      \includegraphics[width=1\linewidth]{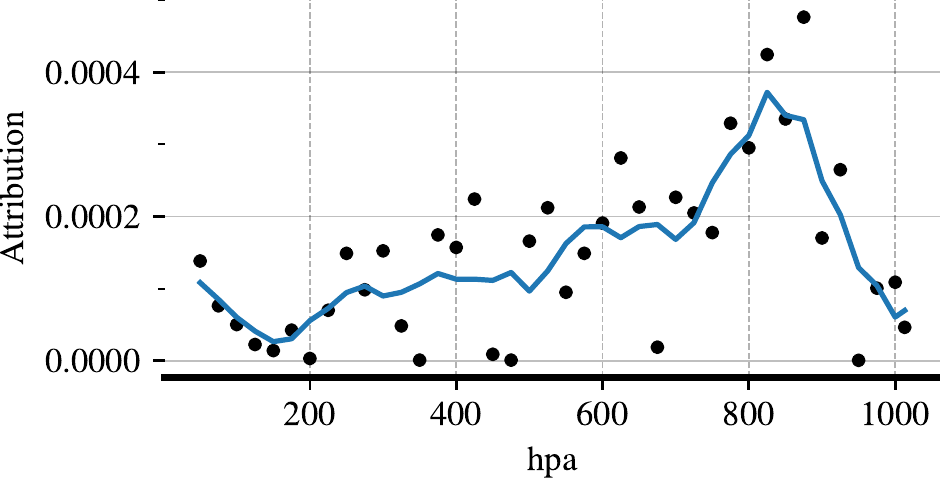}
      \caption{Temperature}
      \label{fig:temp}
    \end{subfigure}
    \begin{subfigure}[b]{0.48\textwidth}
      \centering
      \includegraphics[width=1\linewidth]{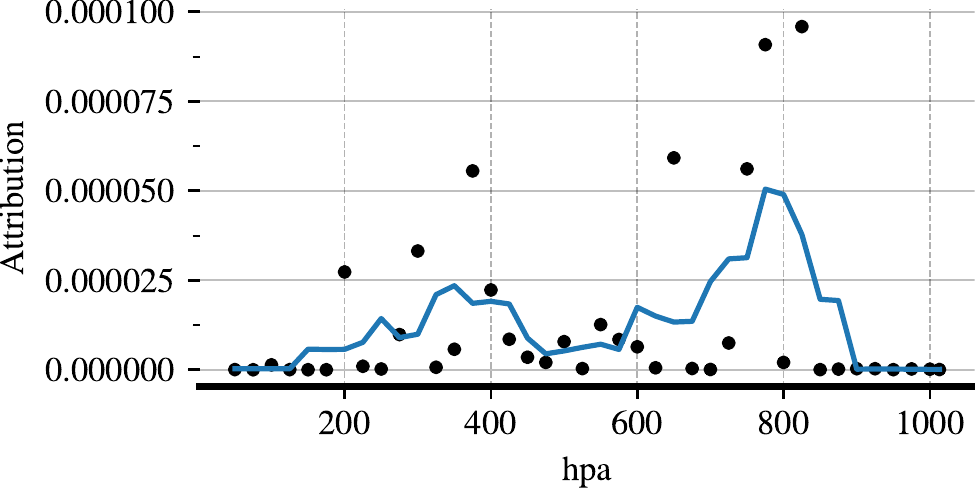}
      \caption{Dew Point Temperature}
      \label{fig:dpt}
    \end{subfigure}

    \begin{subfigure}{.48\textwidth}
      \centering
      \includegraphics[width=1\linewidth]{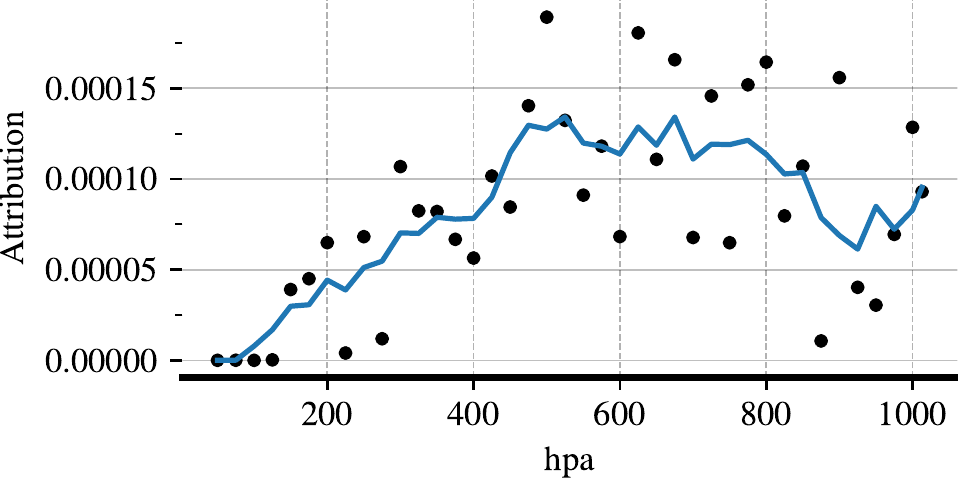}
      \caption{Relative Humidity}
      \label{fig:rh}
    \end{subfigure}
    \begin{subfigure}{.48\textwidth}
      \centering
      \includegraphics[width=1\linewidth]{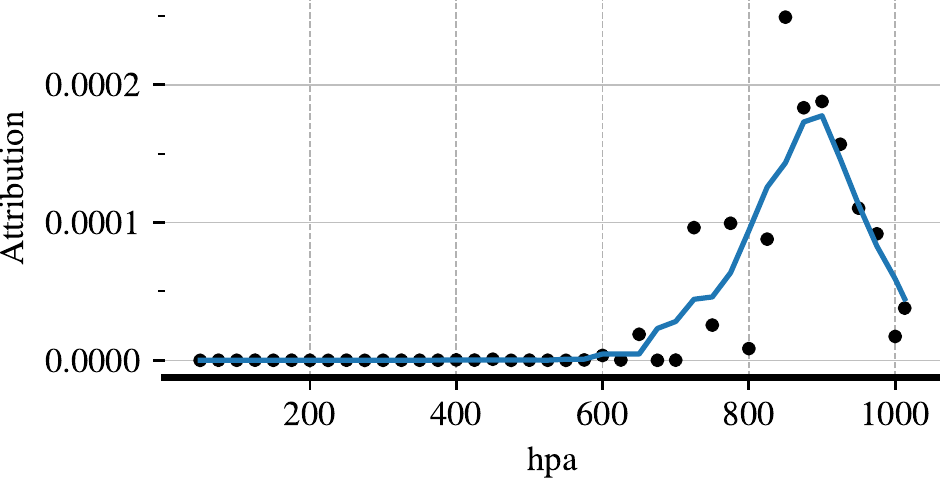} 
      \caption{Specific Humidity}
      \label{fig:sh}
    \end{subfigure}
    \centering
    \includegraphics[scale=.6]{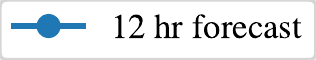}
    \caption{Attribution of different weather features at 12 hour forecast.}
    \label{fig:hrrr}
\end{figure}

\subsection{Analysis of Assimilation}

Including the Assimilaton features as an input was significantly beneficial to \mt{} (Figure~\ref{fig:ablation_hrrr}). These features included the surface level weather features as well as the data at pressure levels ranging from 50~hPa to 1013~hPa. Figure~\ref{fig:hrrr} plots the attribution of some of these features at 40 different pressure levels in this range, from just above the troposphere to sea level.

In a noteworthy observation, Figure~\ref{fig:absv} shows that absolute vorticity at pressure levels near 250 hPa is important for twelve hour precipitation forecasts. Absolute vorticity is the curl of the horizontal winds, so it can be considered a feature combining the effects of the u- and v-components of the wind. Figure~\ref{fig:absv} shows that the relative importance of absolute vorticity is small for near-term forecasts, but grows in importance as lead time increases.  
The importance of upper-level vorticity for a twelve hour forecast is consistent with quasi-geostrophic theory (QG theory).  QG theory is a set of simplifications and filtering of the equations of motion, and a key result is that positive vorticity in the upper-troposphere is consistent with upward motion in the lower-troposphere \cite{bluestein1992synoptic}.  This upward motion, does not directly trigger precipitation, but prepares the atmosphere for convection.  

Figure~\ref{fig:temp} is a plot of the attribution of temperature at different pressure levels, with a peak at $\approx$775~-~850~hPa. Temperature at 850~hPa, approximately 1.5~km above sea level and over the atmospheric boundary layer, generally used for detection of warm and cold fronts. At this height, the temperature sees no daily variations and effects of cooler surfaces such as the ocean are minimal. 

The attribution plot of dew point temperature, Figure~\ref{fig:dpt}, shows two peaks at 375 hPa and 775 hPa, indicating the moisture in the upper level and lower level of the atmosphere is predictive of precipitation 12 hours out.

In Figure~\ref{fig:rh} and Figure~\ref{fig:sh}, we can see a comparison of attributions between relative humidity and specific humidity, respectively. Specific humidity levels in the lower levels of the atmosphere indicate how much moisture there is for formation of storms, higher the moisture, higher the precipitation rates will be. Relative humidity is likely indicating large-scale cloudiness.

\begin{figure}[H]
    \centering
    \begin{subfigure}{.48\textwidth}
      \centering
      \includegraphics[width=1\linewidth]{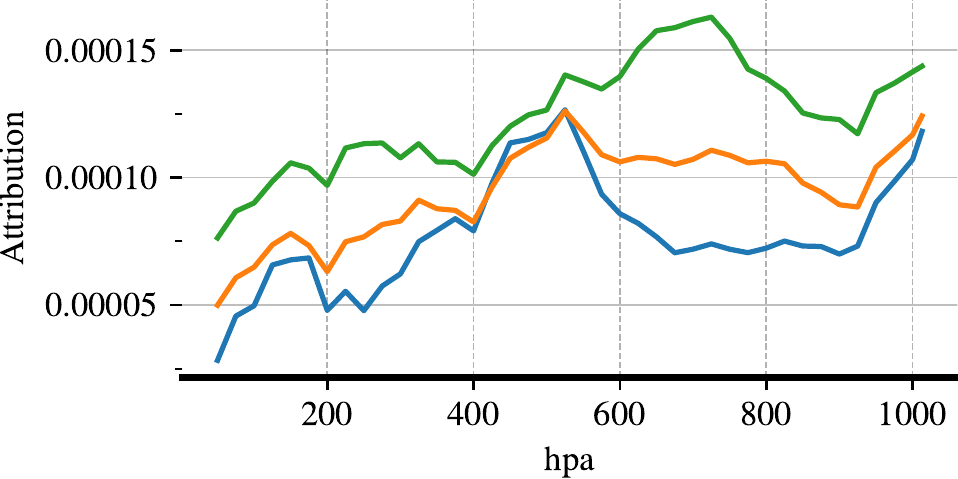}
      \caption{U-component of wind}
      \label{fig:uwind}
    \end{subfigure}
    \begin{subfigure}{.48\textwidth}
      \centering
      \includegraphics[width=1\linewidth]{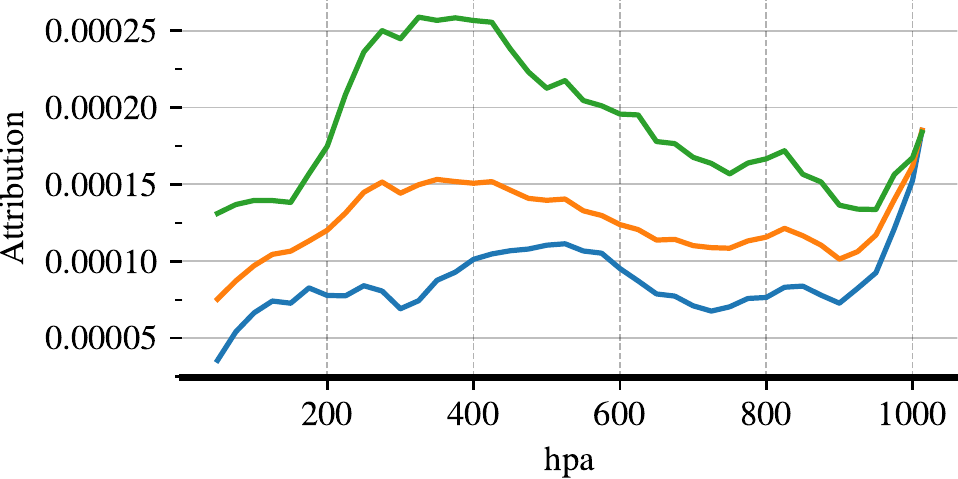}
      \caption{V-component of wind}
      \label{fig:vwind}
    \end{subfigure}
    \centering
    \includegraphics[scale=.6]{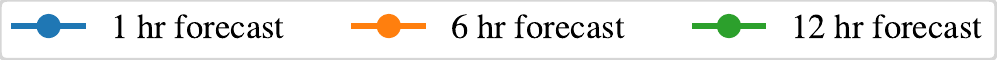}
    \caption{Attribution of wind over different forecast hours and pressure.}
    \label{fig:wind}
\end{figure}

In Figure \ref{fig:wind}, we look at how the attributions change for the u and v-component of wind.

For a 1hr forecast, surface level wind is most important for both components. But beyond the 4hr forecast, notice that v-component of the surface wind has substantially lesser importance than that in the upper troposphere (200-400 hPa). The v-component, the meridional flow of winds, has higher importance in the upper troposphere. Significant values of this component in the upper level indicates a low that is favorable for producing precipitation. Meridional flow aloft can be found near upper-level lows, which are favorable for large scale precipitation.  

It is expected that the u-component, flow of horizontal wind towards North, is not as important in the upper troposphere since it represents how fast the wind blowing.

We also noticed that for 1 hour nowcasting, features such as snow mixing ratio, cloud water mixing ratio had greater relative importance than at later hours. From this the model essentially learns that if it is raining now then it will continue to rain in the near future. \mt{} has learned the technique of persistence.

\begin{figure}[H]
    \begin{subfigure}{.48\textwidth}
      \centering
      \includegraphics[width=1\linewidth]{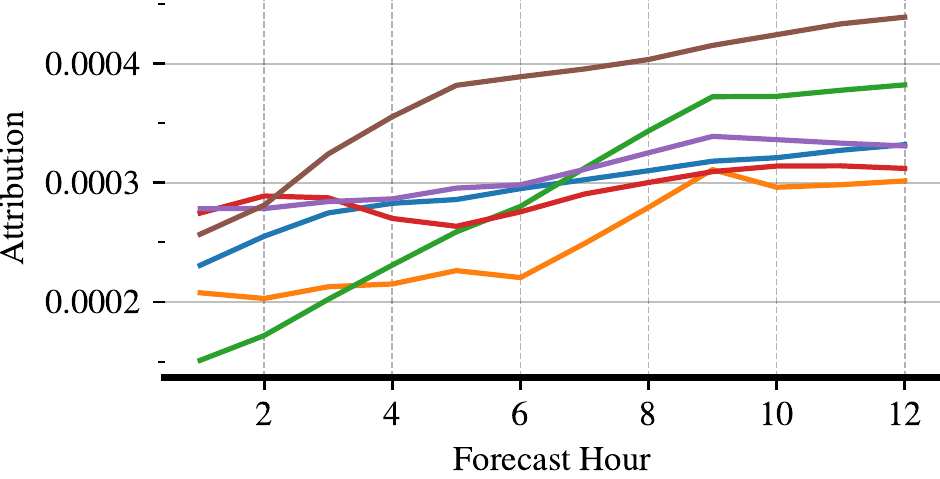}
      \caption{}
      \label{fig:inc_attr}
    \end{subfigure}
    \centering
    \includegraphics[scale=.6]{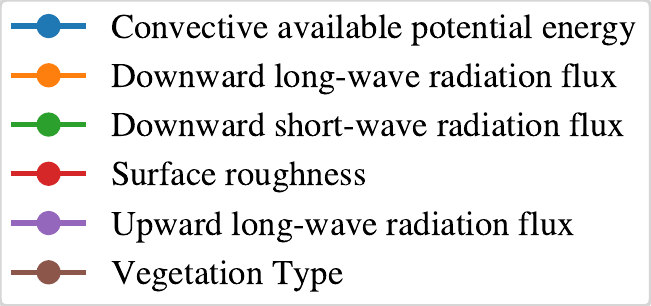}
    \begin{subfigure}{.48\textwidth}
      \centering
      \includegraphics[width=1\linewidth]{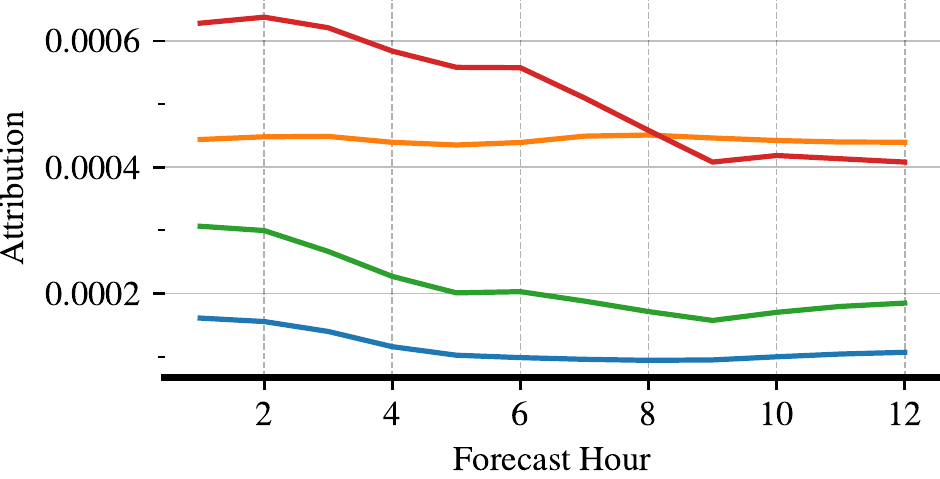}
      \caption{}
      \label{fig:dec_attr}
    \end{subfigure}
    \centering
    \includegraphics[scale=.6]{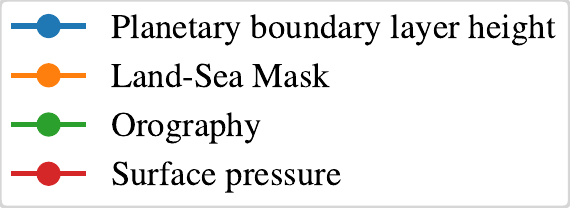}
    \caption{Surface features increasing (a) or decreasing (b) in importance with increasing forecast hour.}
    \label{fig:attr_hr}
\end{figure}

We also looked at behaviors of various surface level features. Figure~\ref{fig:inc_attr} plots features that increased in importance with increasing length of forecast and, in contrast, Figure~\ref{fig:dec_attr} plots features that decreased in importance. Downward short-wave radiation flux together with the upward long wave radiation flux indicate how much sunlight is reaching the surface, thereby warming it. When there is not enough sunlight and surface temperatures remain cool, deep convective storms are unlikely. Downward long-wave radiation, on the other hand, is the heat emitted back to Earth from the atmosphere, indicating how much cloud cover is currently present, which is useful for nowcasting.
Planetary boundary layer height varies throughout the day and is therefore less reliable for what can happen a 12 hours later. Similarly, the surface pressure indicates where the storms are forming in the present, being more useful for nowcasting.

\subsection{Local Interpretability Results}

\newcommand{\interpretinput}[1] {%
  \raisebox{-0.5\height}{%
    \includegraphics[scale=.3]{interpretability_plots/#1.png}}}

\newcommand{\interpretattr}[1] {%
  \raisebox{-0.5\height}{%
    \includegraphics[scale=.3]{interpretability_plots/#1_attr.png}}}

\begin{figure}
\centering
\begingroup
\newcommand{\firsthourpos}[0]{\inlinebottom}
\newcommand{\thirdhourpos}[0]{\inlinebottom}
\newcommand{\sixthhourpos}[0]{\inlinebottom}
\newcommand{\twelvedhourpos}[0]{\inlinebottom}
\setlength\tabcolsep{1.5pt}
\begin{tabular}{@{}c c c c c r@{}}
  & \smallb{Composite radar reflectivity} & \smallb{Absolute vorticity(275hPa)} & \smallb{V-component wind(375 hPa)} & \smallb{Dew-point Temp(825 hPa)} & \\ 
  \vtitle{Input}
      & \interpretinput{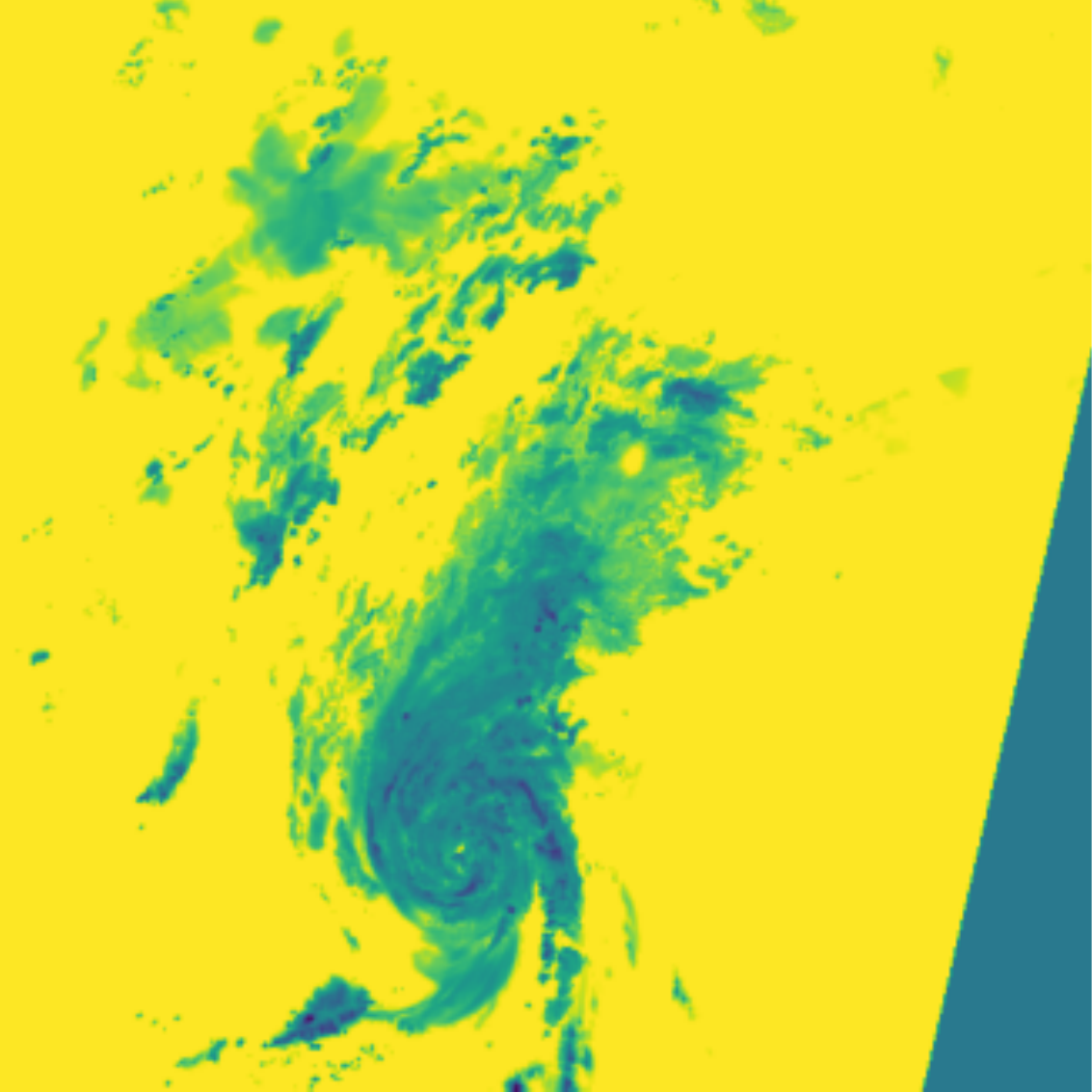}
      & \interpretinput{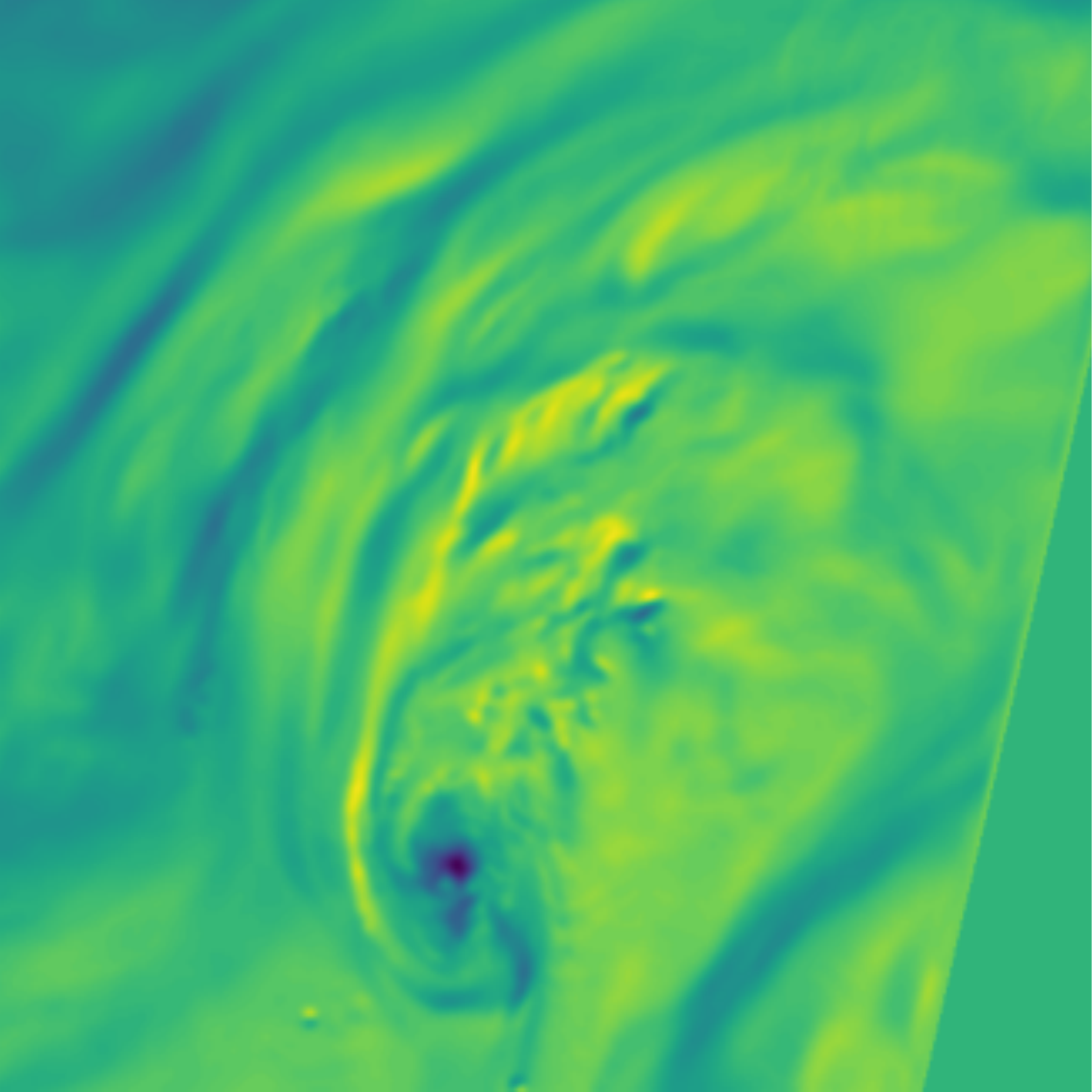}
      & \interpretinput{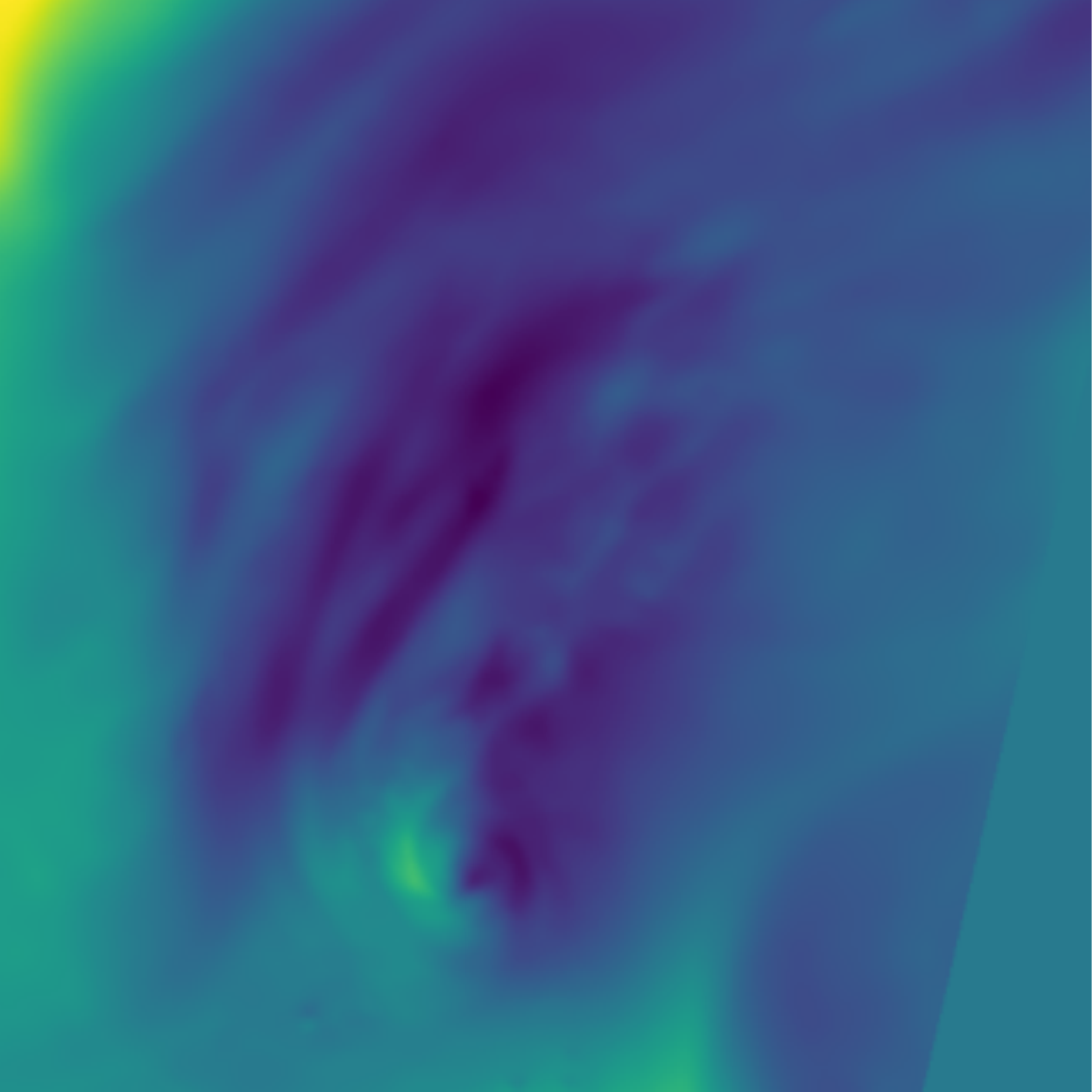}
      & \interpretinput{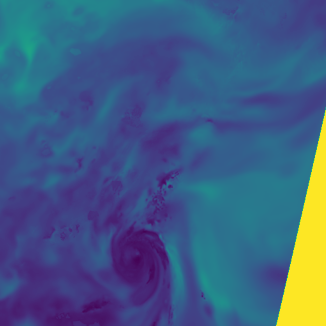}\myrowspace
      & \raisebox{-0.5\height}{\includegraphics[scale=.3]{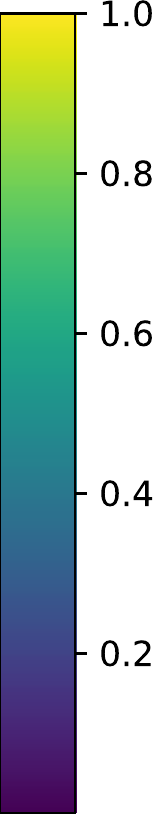}} \\
  \vtitle{Attribution}
      & \interpretattr{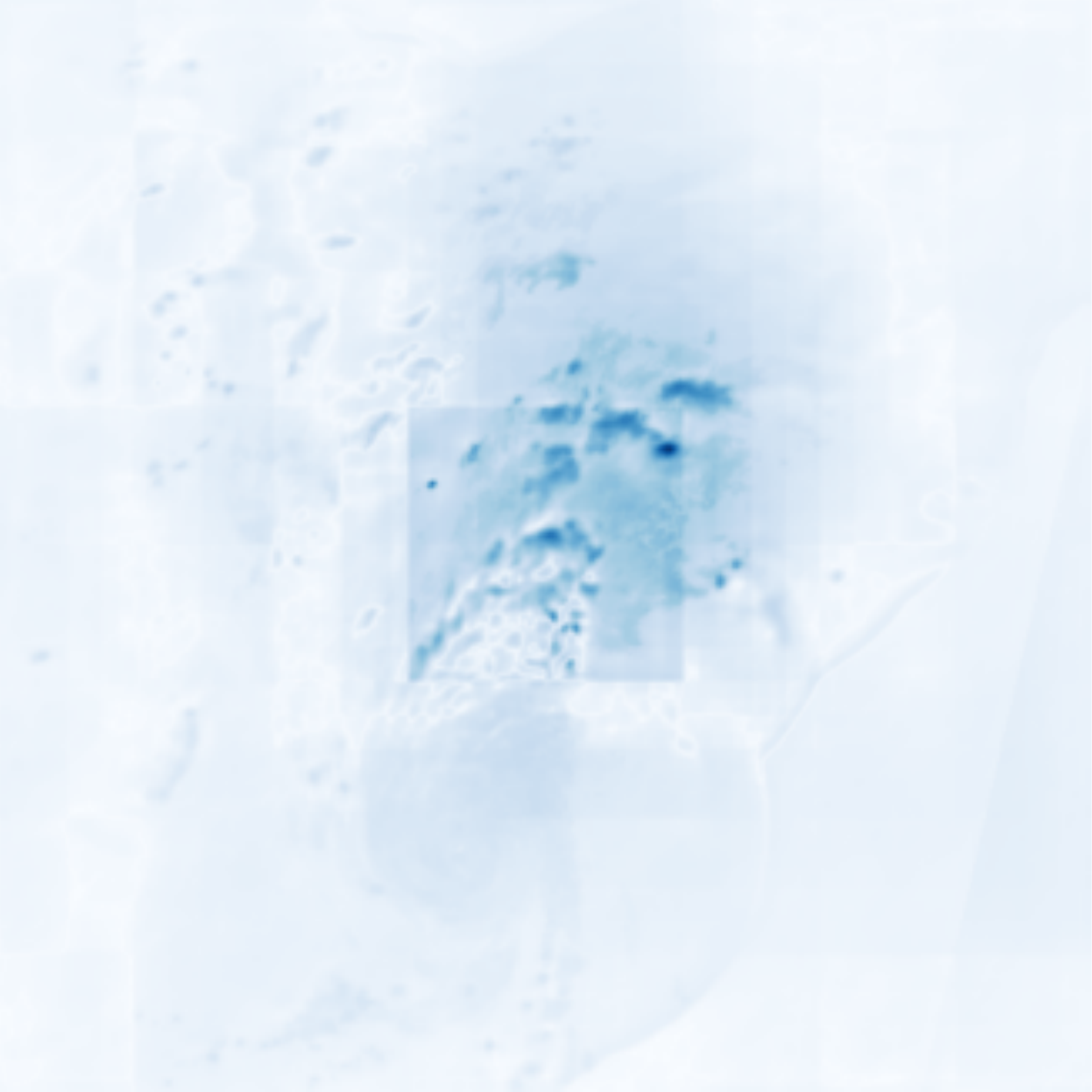}
      & \interpretattr{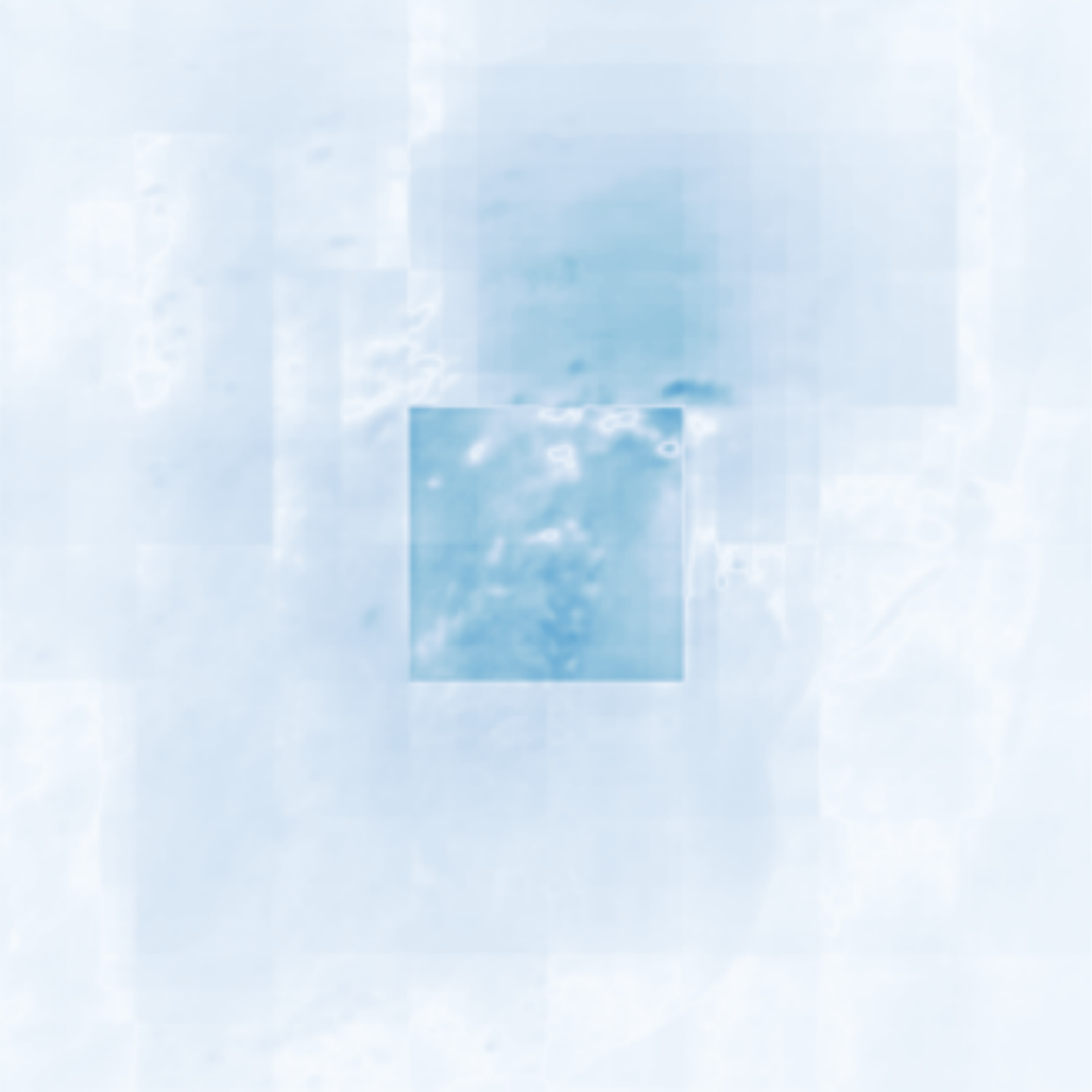}
      & \interpretattr{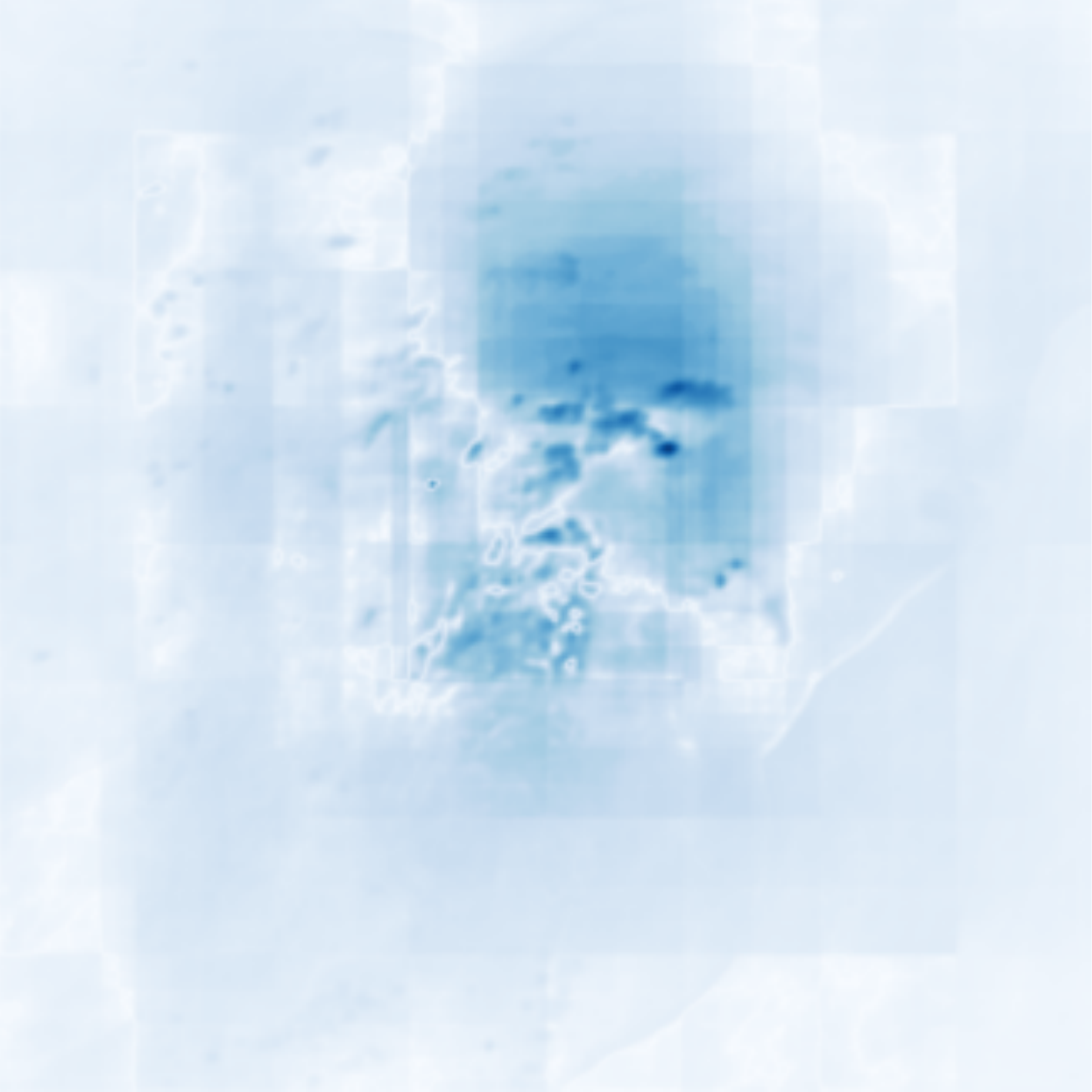}
      & \interpretattr{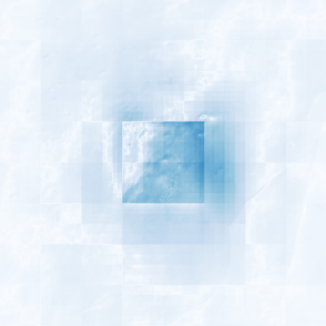}\myrowspace
      & \raisebox{-0.5\height}{\includegraphics[scale=.3]{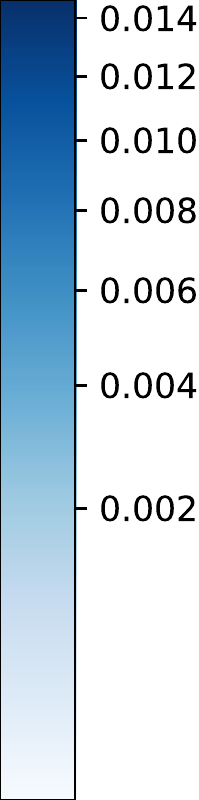}} \\
\end{tabular}
\caption{Attribution maps for MetNet-2's prediction of Hurricane Isaias shown in Figure~\ref{fig:main_case_b}.}
\label{fig:local_hurricane}
\endgroup
\end{figure}

To analyze the pixels in each sample that MetNet-2 was keying in on, we plotted attribution maps for every input feature. Figure~\ref{fig:local_hurricane}, shows attributions of the 12 hour forecast for features--radar reflectivity, absolute vorticity at 275~hPa and the v-component of wind at 375~hPa. For all of these, the entire context was influential, with pixels closer to the center of the sample being more important than those at the borders. The weather features that were unimportant to the prediction has attribution maps with values close to zero.

\newcommand\newsubcap[1]{\phantomcaption%
    \caption*{\figurename~\thefigure\ (\thesubfigure): #1}}
    
\section{Supplement: Regional Evaluation}
\label{sup:regional_eval}

We evaluate all our models also on a diverse set of sub-regions of CONUS using the same radar targets that we use on the full CONUS evaluation. We choose this set of regions because they represent different climatological regimes of precipitation within CONUS while also broadly covering different geographical parts of CONUS. 

\begin{figure}
\centering
\begin{subfigure}{.333\textwidth}
    \centering
    \smallb{.2 mm}
    \\
    \includegraphics[width=1\textwidth]{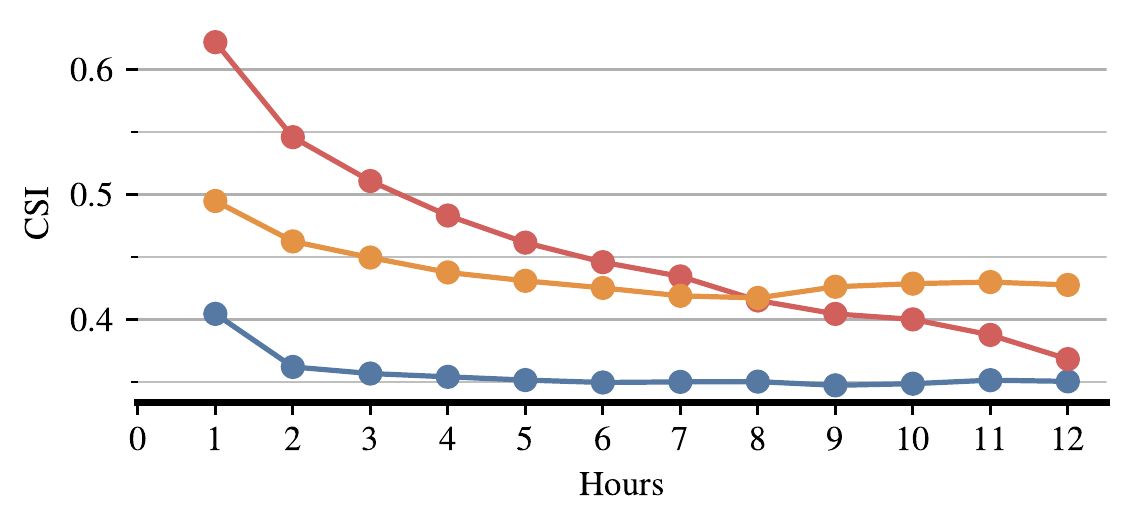}
    \\
    \smallb{1 mm}
    \\
    \includegraphics[width=1\textwidth]{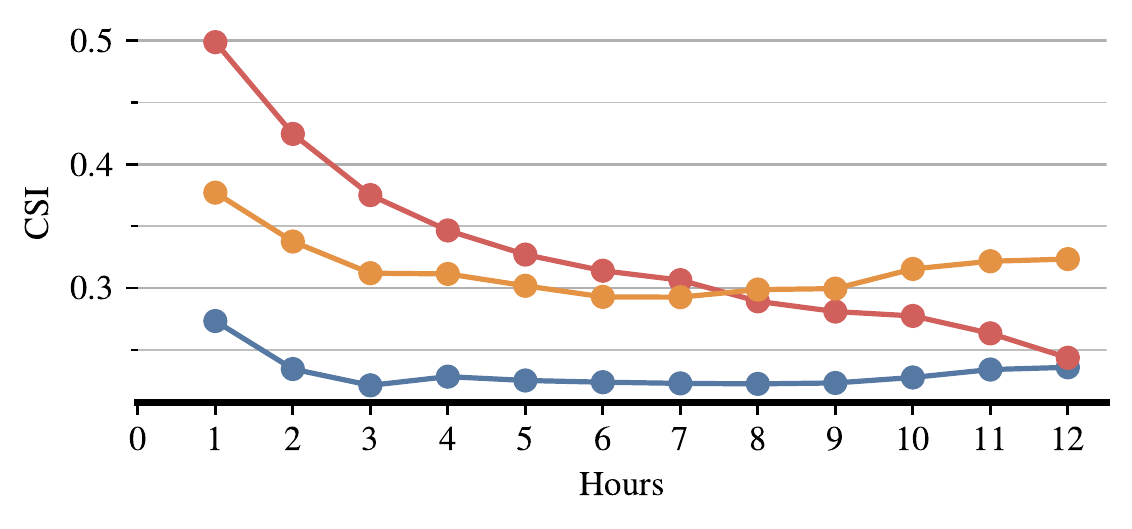}
    \\
    \smallb{2 mm}
    \\
    \includegraphics[width=1\textwidth]{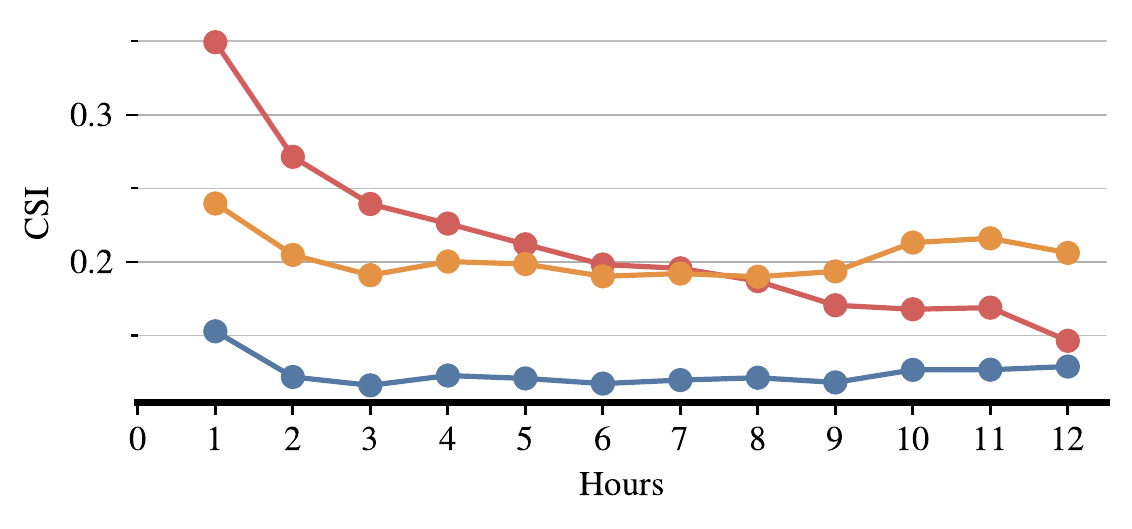}
    \\
    \smallb{4 mm}
    \\
    \includegraphics[width=1\textwidth]{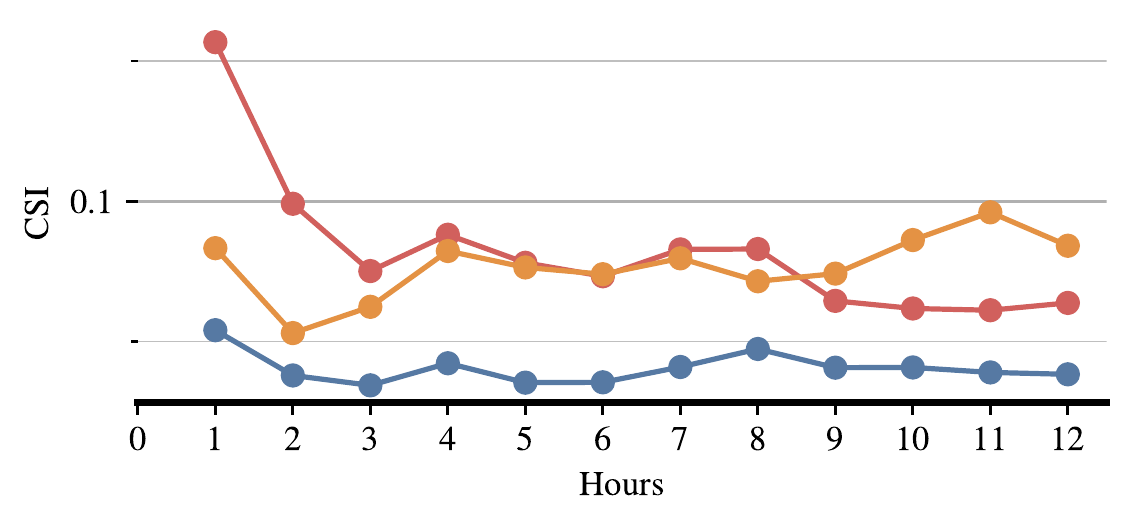}
    \\
    \smallb{8 mm}
    \\
    \includegraphics[width=1\textwidth]{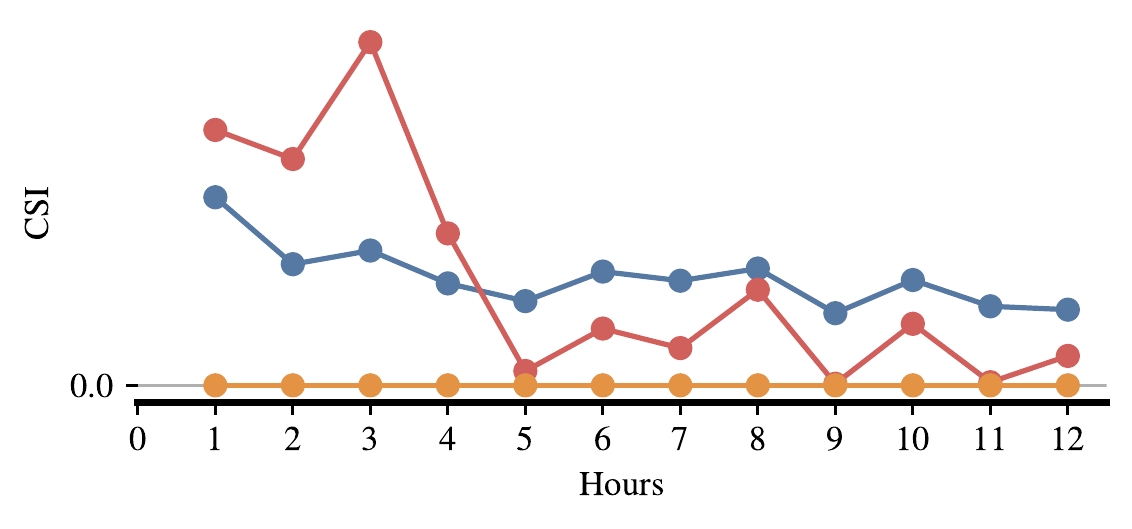}

    \newsubcap{Pacific Northwest}
    \label{fig:nwpacific}
\end{subfigure}%
\begin{subfigure}{.333\textwidth}
    \centering
    \smallb{.2 mm}
    \\
    \includegraphics[width=1\textwidth]{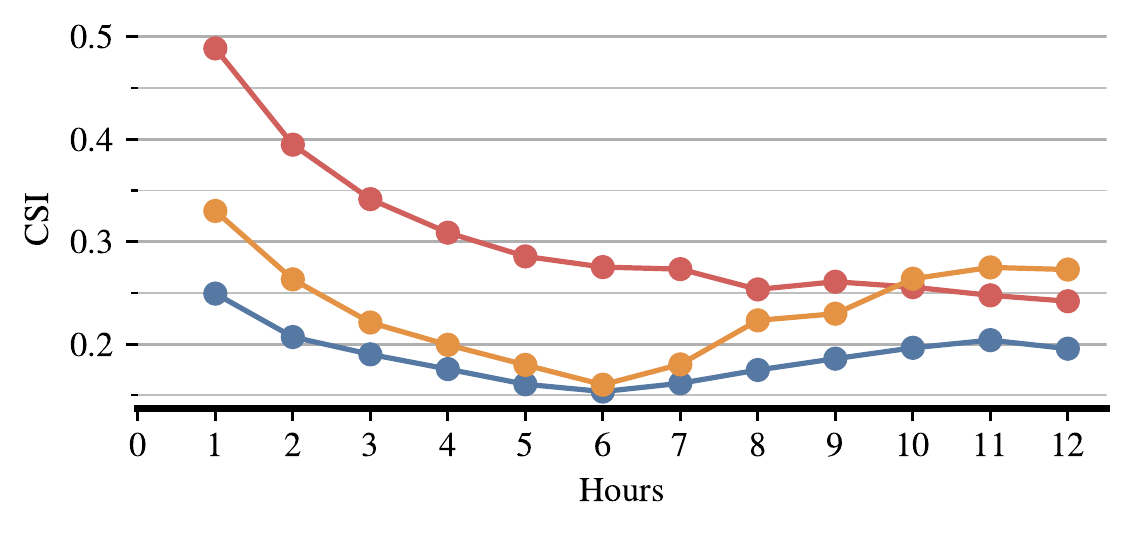}
    \\
    \smallb{1 mm}
    \\
    \includegraphics[width=1\textwidth]{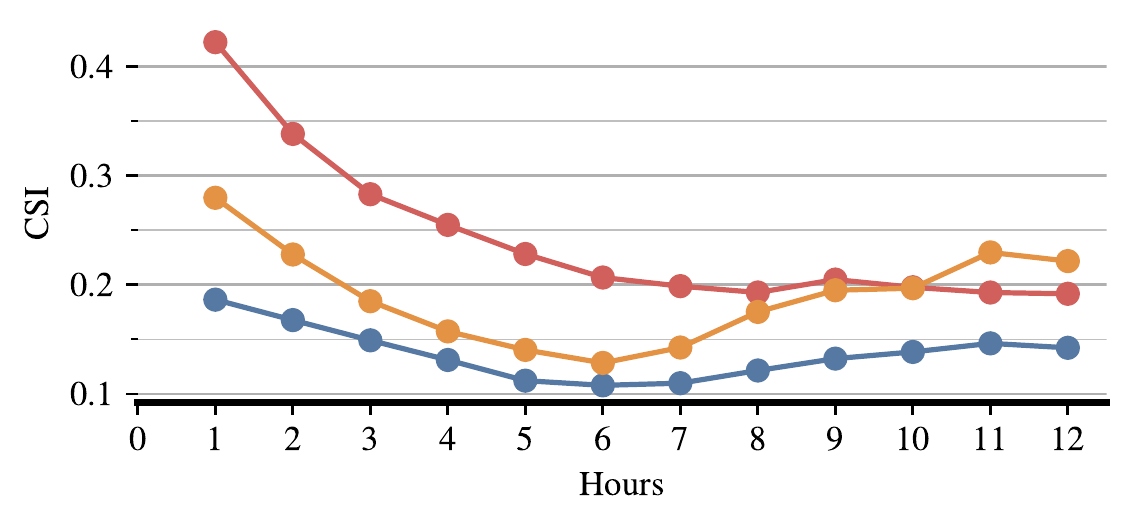}
    \\
    \smallb{2 mm}
    \\
    \includegraphics[width=1\textwidth]{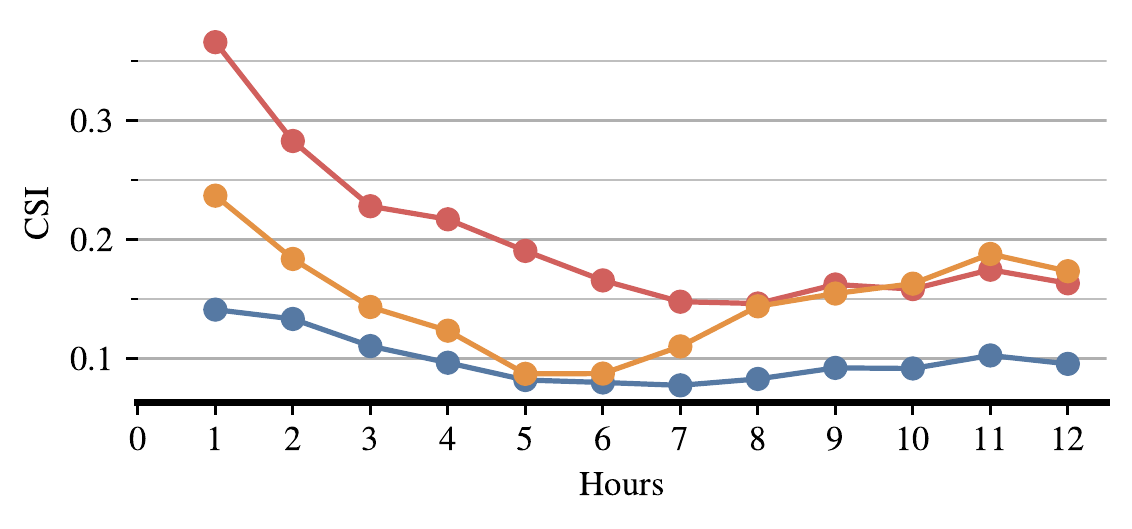}
    \\
    \smallb{4 mm}
    \\
    \includegraphics[width=1\textwidth]{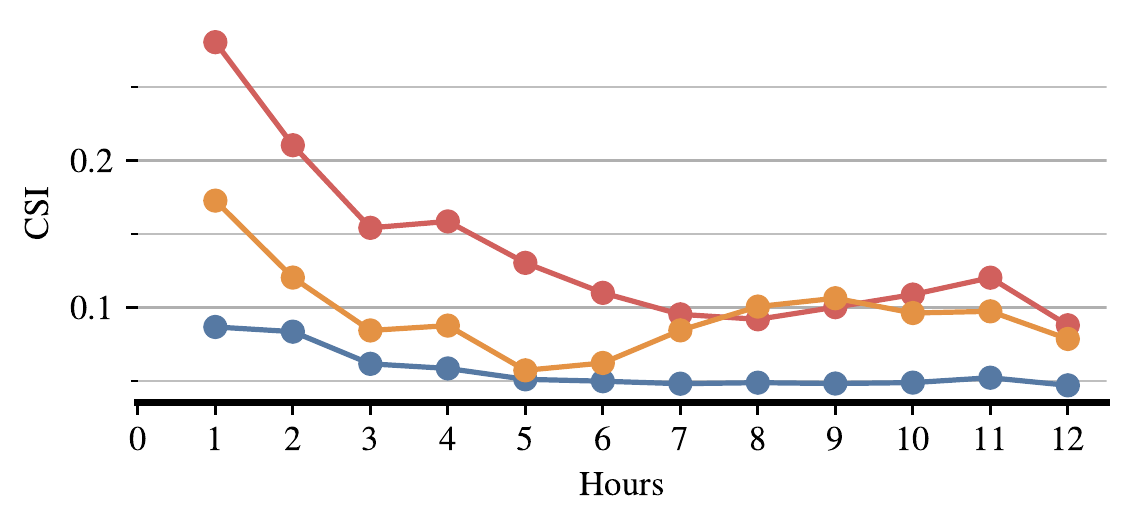}
    \\
    \smallb{8 mm}
    \\
    \includegraphics[width=1\textwidth]{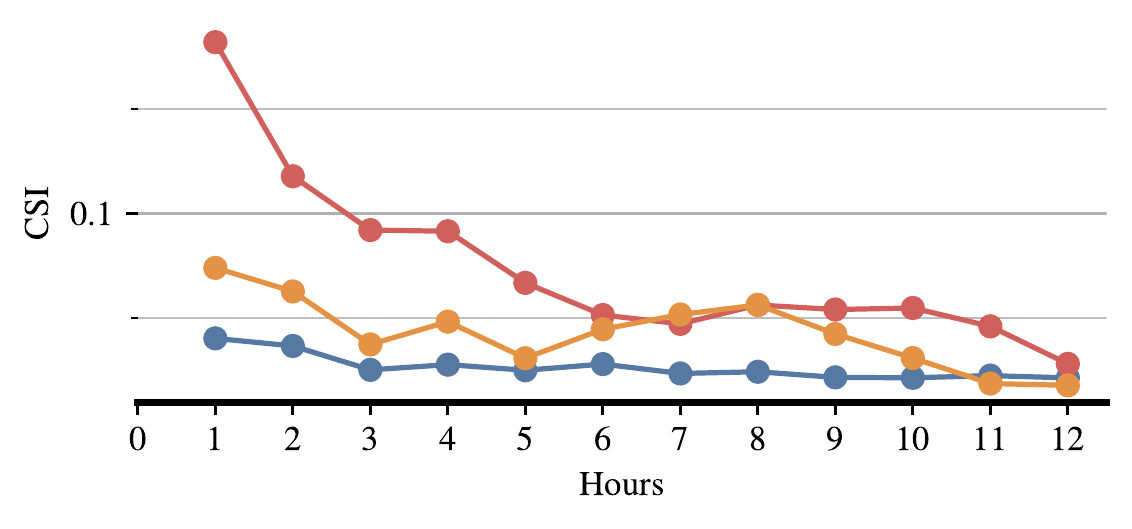}

    \newsubcap{Florida}
    \label{fig:florida}
\end{subfigure}%
\begin{subfigure}{.333\textwidth}
    \centering
    \smallb{.2 mm}
    \\
    \includegraphics[width=1\textwidth]{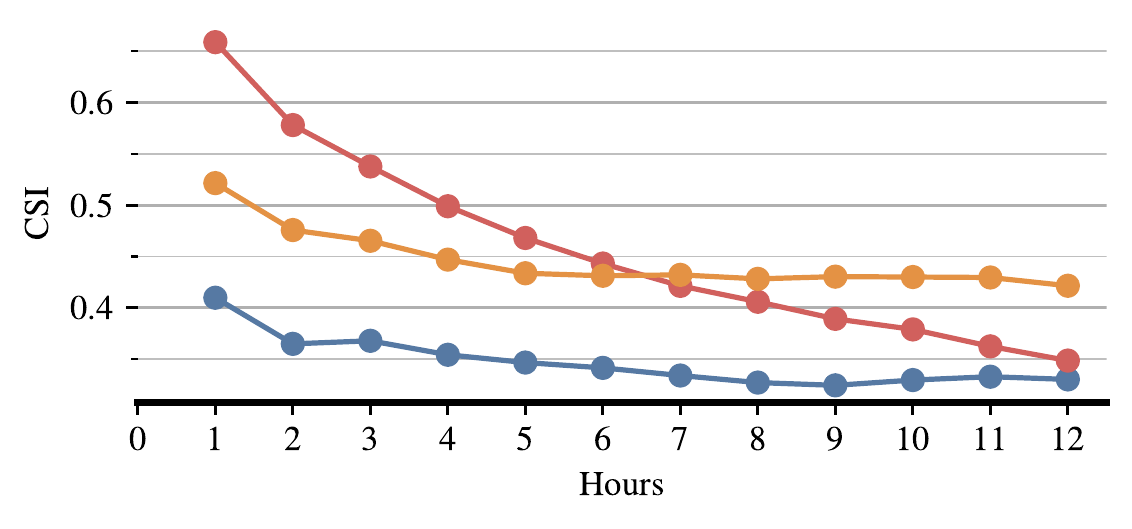}
    \\
    \smallb{1 mm}
    \\
    \includegraphics[width=1\textwidth]{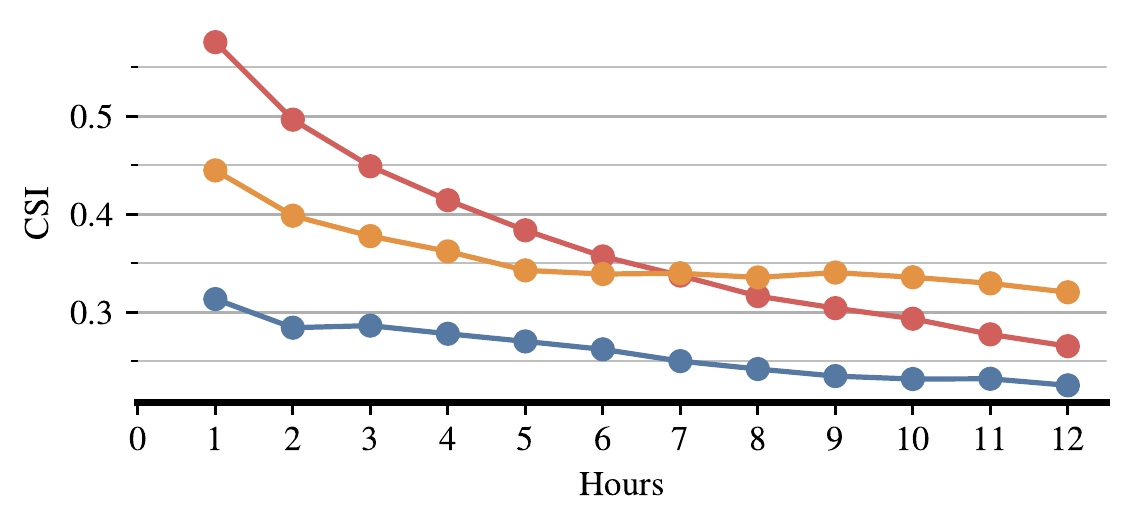}
    \\
    \smallb{2 mm}
    \\
    \includegraphics[width=1\textwidth]{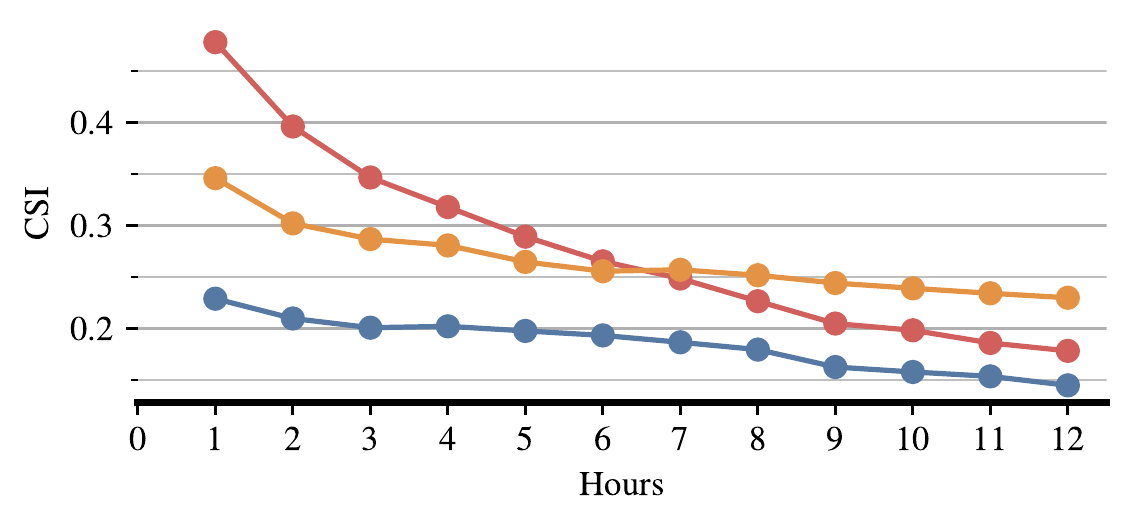}
    \\
    \smallb{4 mm}
    \\
    \includegraphics[width=1\textwidth]{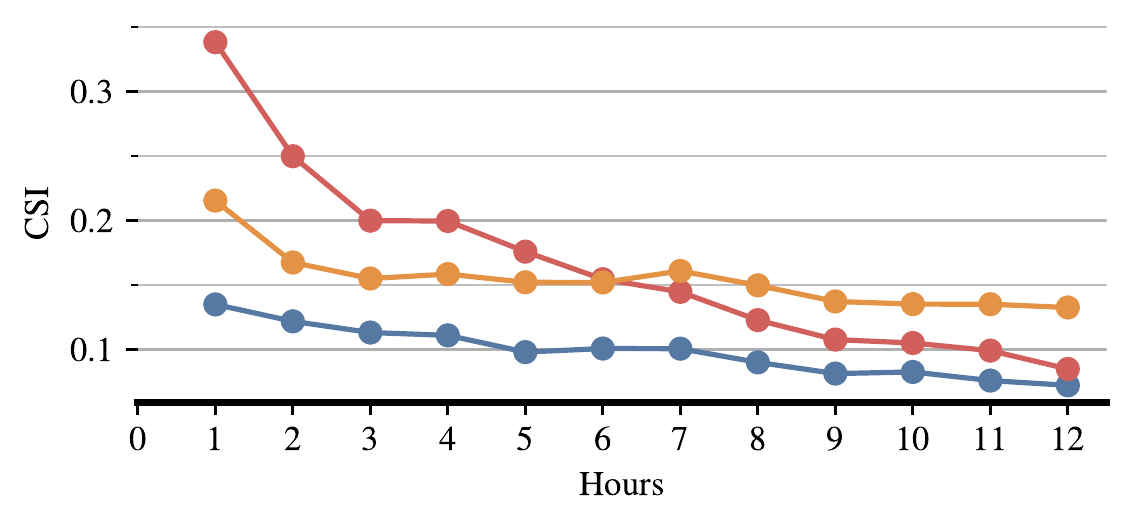}
    \\
    \smallb{8 mm}
    \\
    \includegraphics[width=1\textwidth]{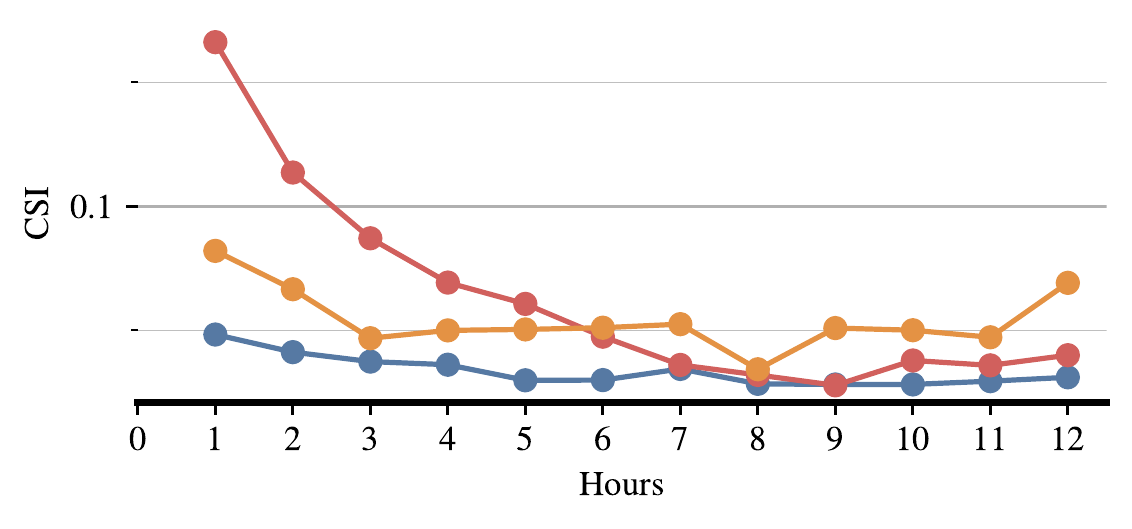}

    \newsubcap{East Coast}
    \label{fig:eastcoast}
\end{subfigure}
\\
\vspace{.5cm}
\includegraphics[scale=.6]{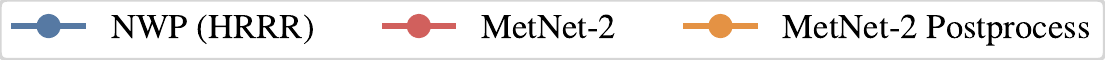}
\\
\vspace{.5cm}
\includegraphics[width=0.5\textwidth]{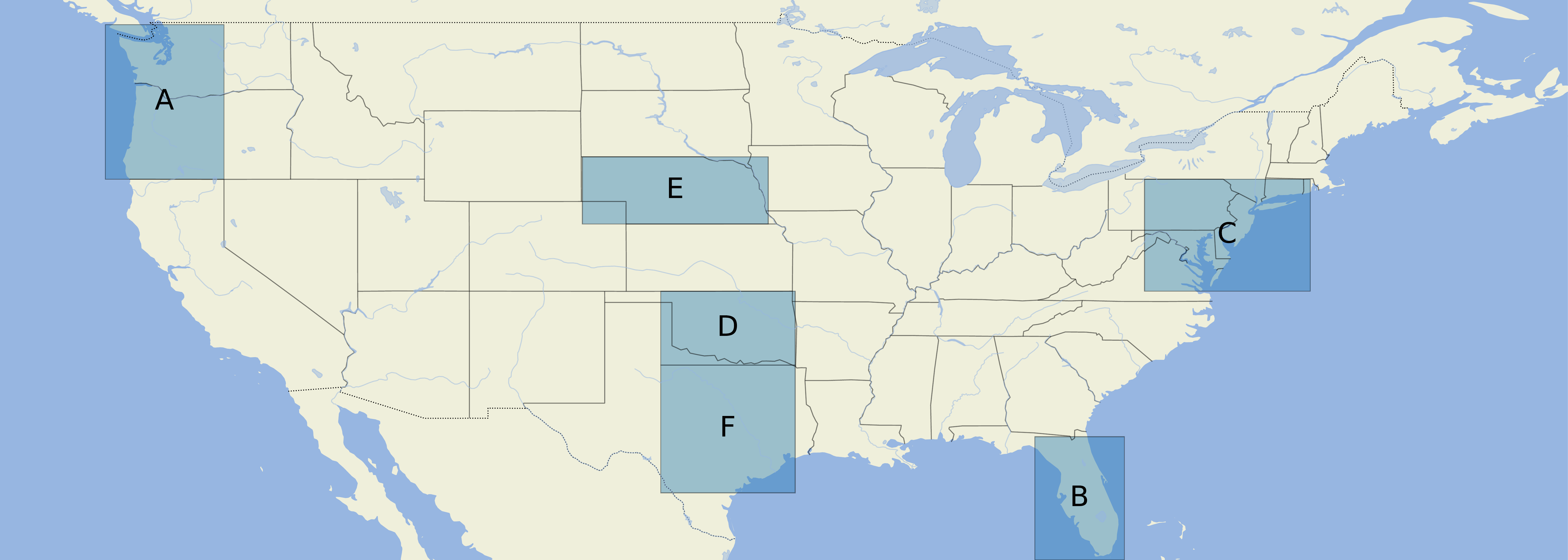}
\captionsetup{labelformat=empty}
\caption{}
\label{fig:regional}
\end{figure}

\begin{figure}\ContinuedFloat
\centering
\begin{subfigure}{.333\textwidth}
    \centering
    \smallb{.2 mm}
    \\
    \includegraphics[width=1\textwidth]{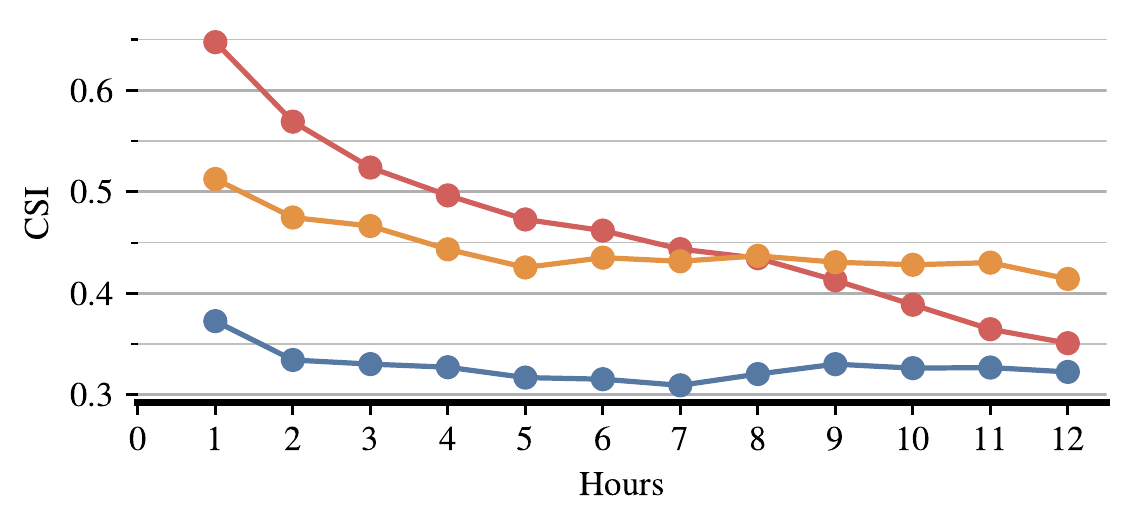}
    \\
    \smallb{1 mm}
    \\
    \includegraphics[width=1\textwidth]{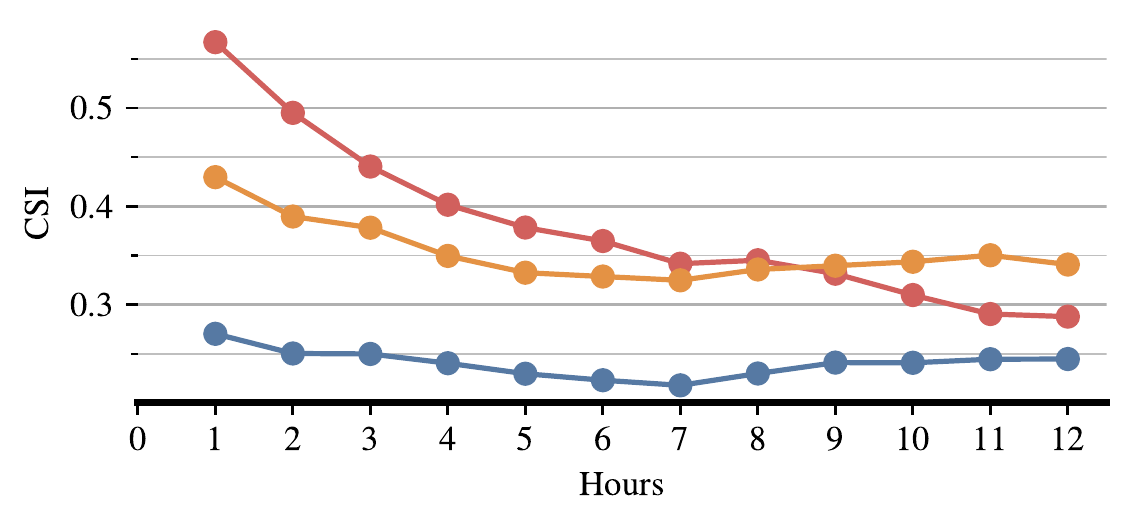}
    \\
    \smallb{2 mm}
    \\
    \includegraphics[width=1\textwidth]{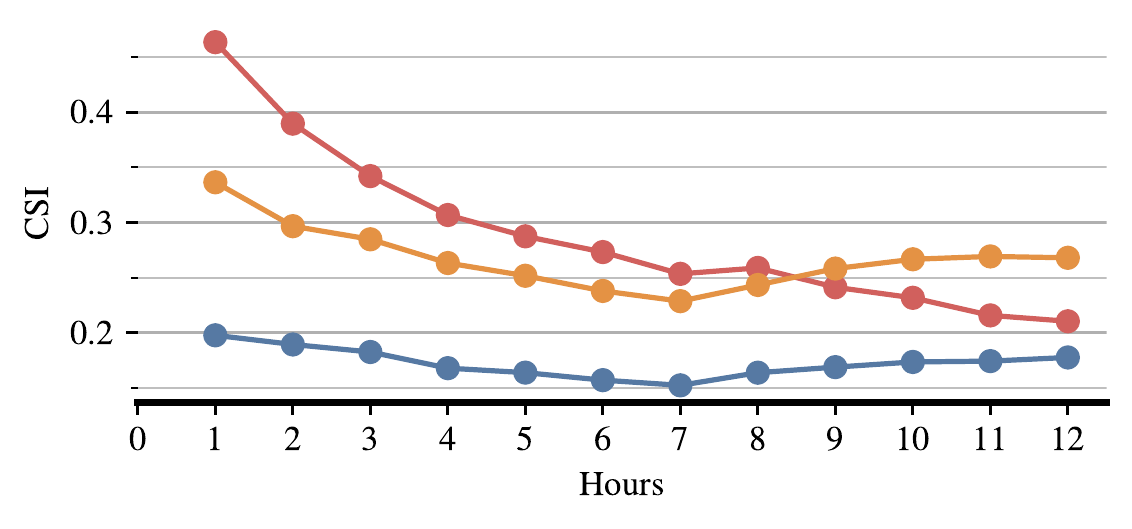}
    \\
    \smallb{4 mm}
    \\
    \includegraphics[width=1\textwidth]{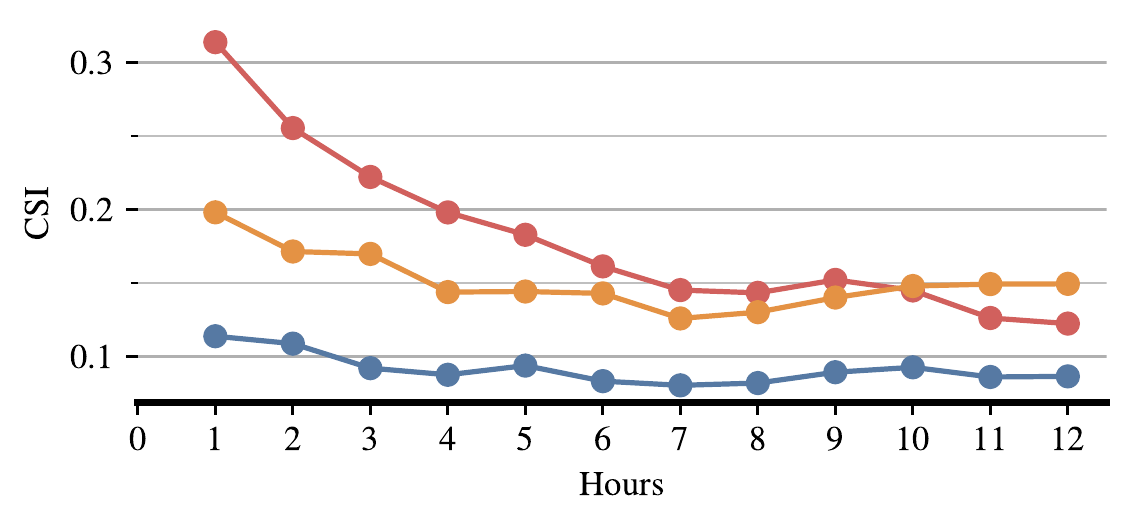}
    \\
    \smallb{8 mm}
    \\
    \includegraphics[width=1\textwidth]{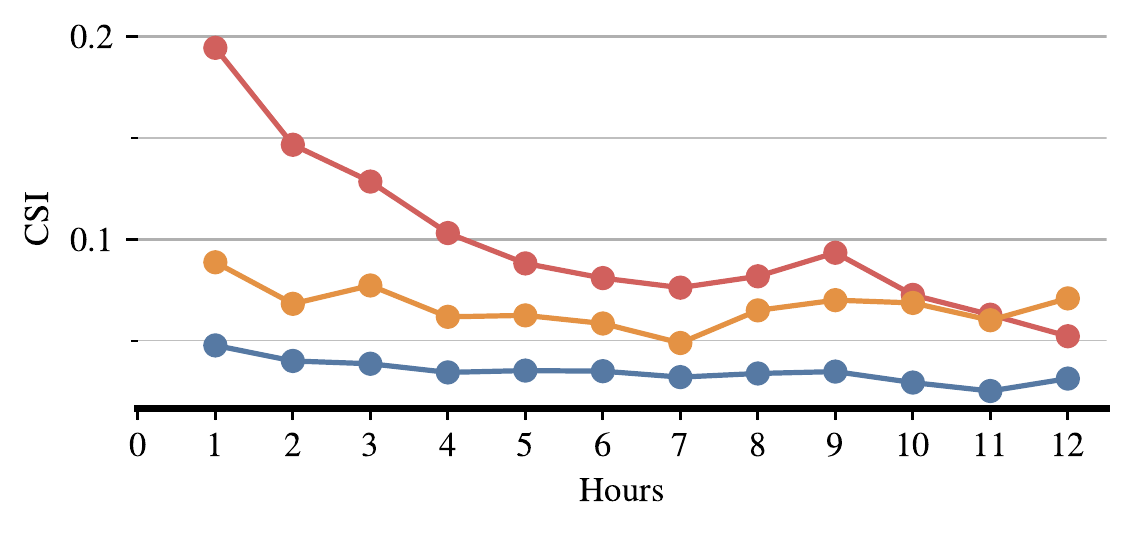}

    \newsubcap{South Plains\\(Oklahoma)}
    \label{fig:southplains}
\end{subfigure}%
\begin{subfigure}{.333\textwidth}
    \centering
    \smallb{.2 mm}
    \\
    \includegraphics[width=1\textwidth]{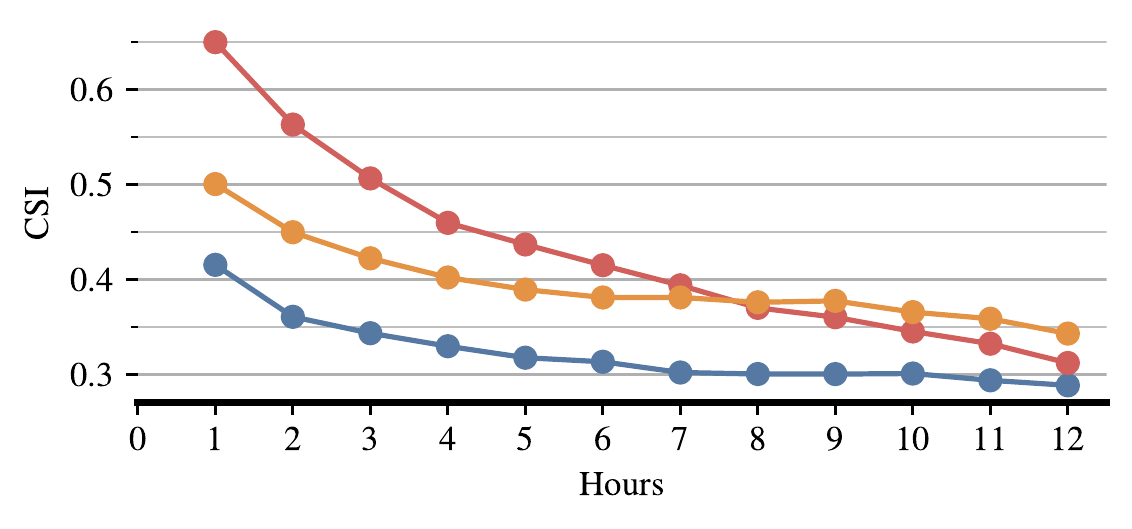}
    \\
    \smallb{1 mm}
    \\
    \includegraphics[width=1\textwidth]{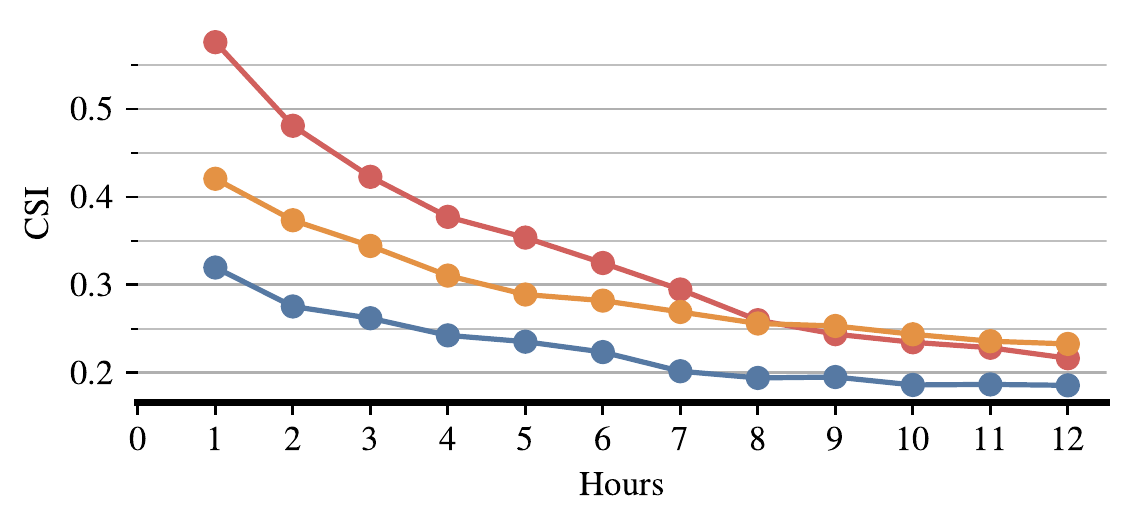}
    \\
    \smallb{2 mm}
    \\
    \includegraphics[width=1\textwidth]{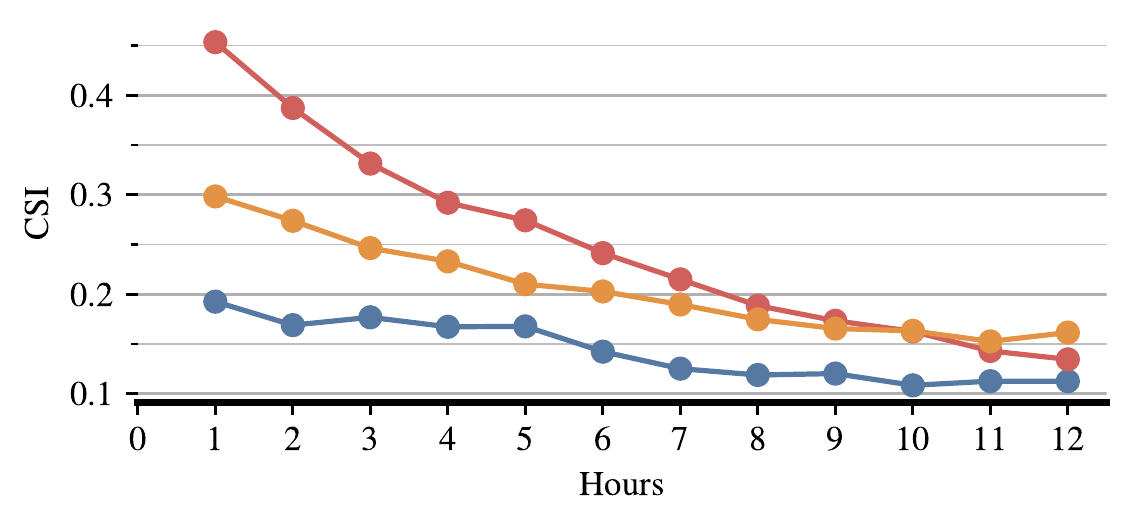}
    \\
    \smallb{4 mm}
    \\
    \includegraphics[width=1\textwidth]{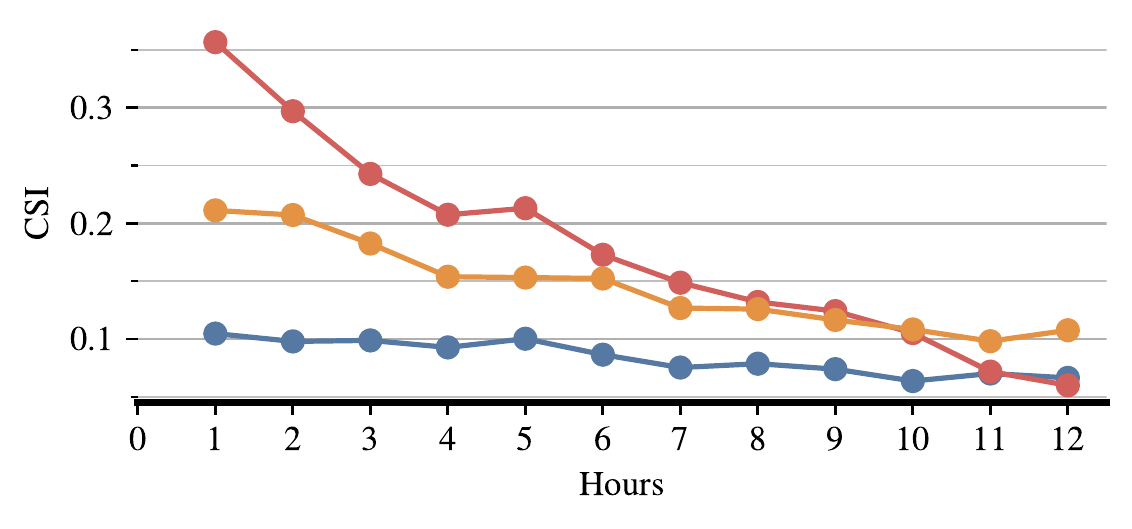}
    \\
    \smallb{8 mm}
    \\
    \includegraphics[width=1\textwidth]{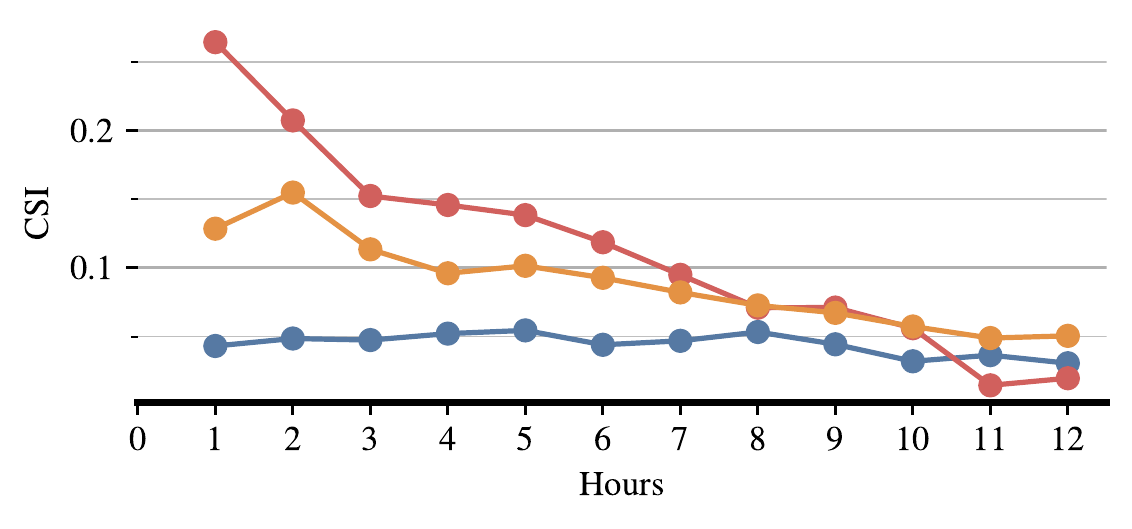}

    \newsubcap{North Plains (Nebraska)\\\,}
    \label{fig:northplains}
\end{subfigure}%
\begin{subfigure}{.333\textwidth}
    \centering
    \smallb{.2 mm}
    \\
    \includegraphics[width=1\textwidth]{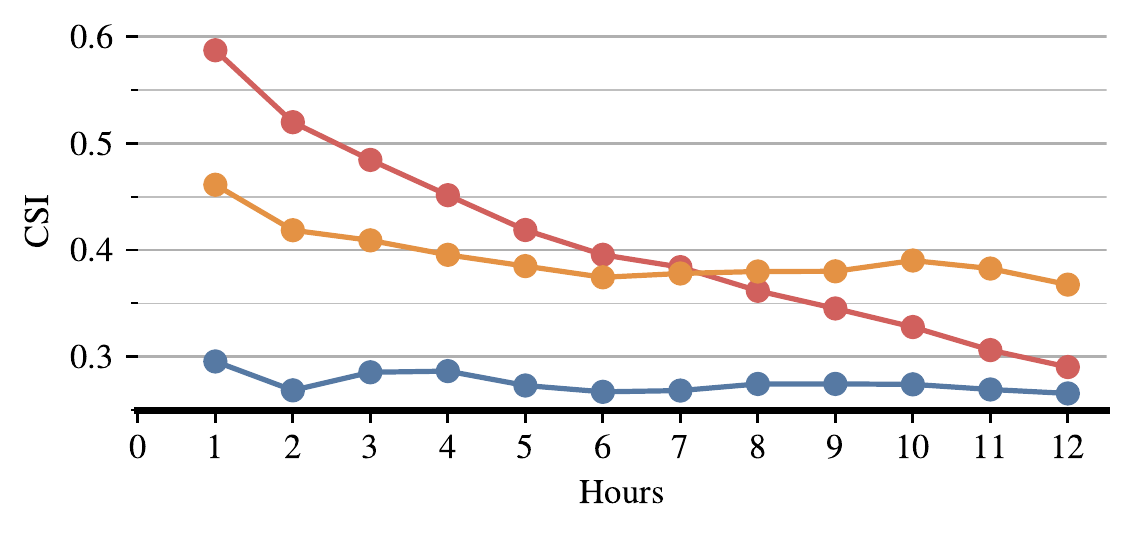}
    \\
    \smallb{1 mm}
    \\
    \includegraphics[width=1\textwidth]{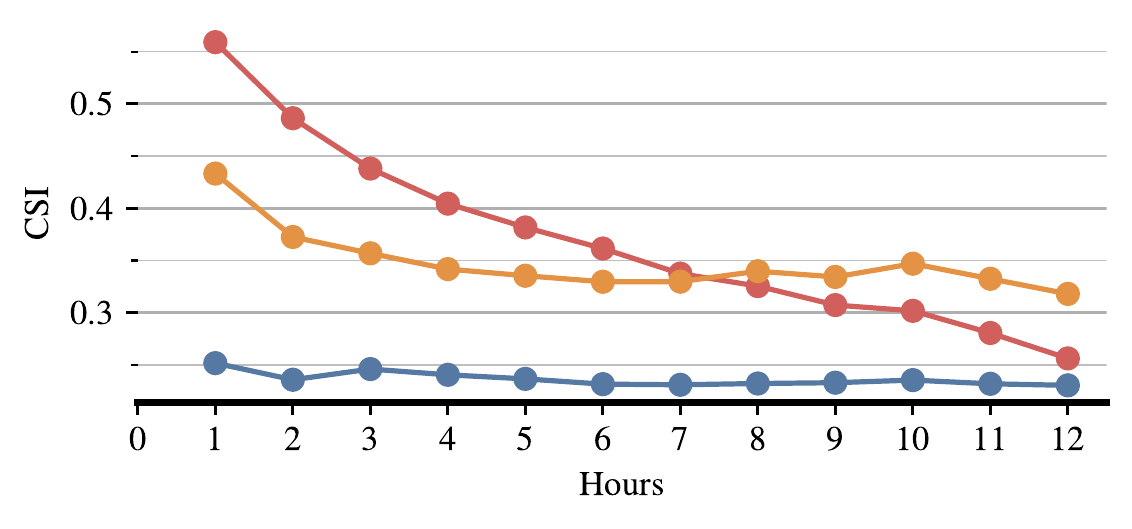}
    \\
    \smallb{2 mm}
    \\
    \includegraphics[width=1\textwidth]{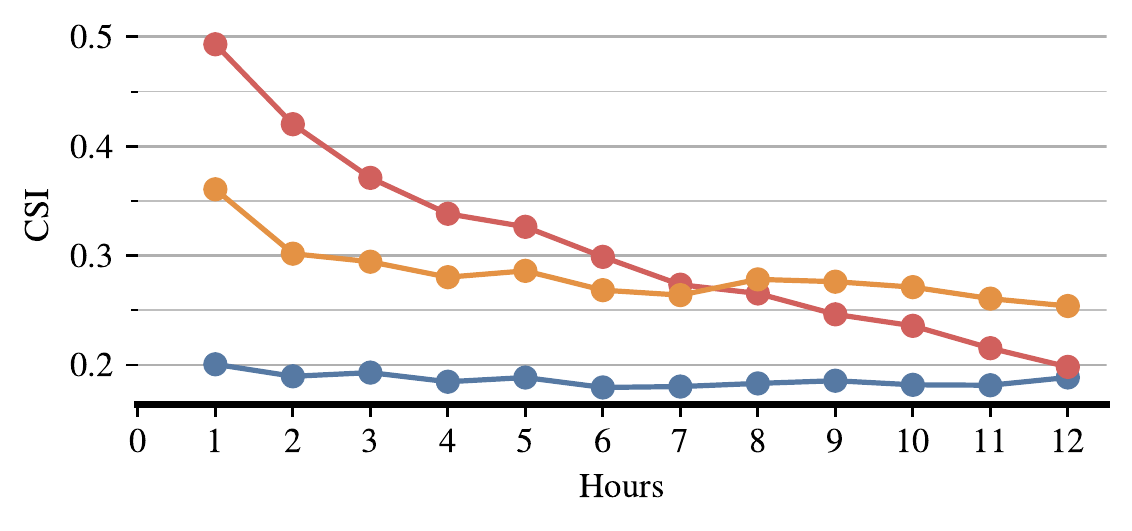}
    \\
    \smallb{4 mm}
    \\
    \includegraphics[width=1\textwidth]{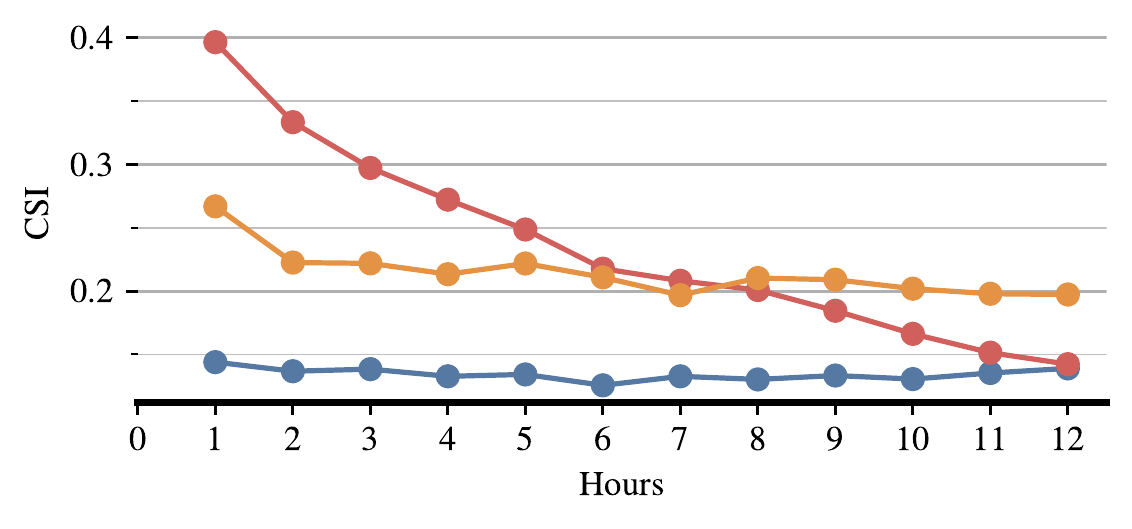}
    \\
    \smallb{8 mm}
    \\
    \includegraphics[width=1\textwidth]{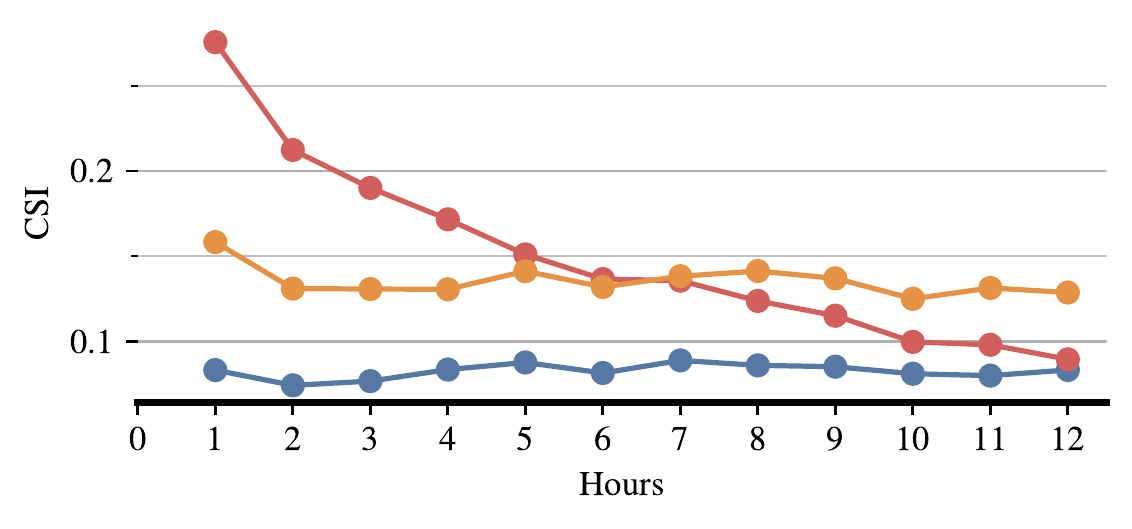}

    \newsubcap{Texas\\\,}
    \label{fig:Texas}
\end{subfigure}
\\
\includegraphics[scale=.6]{plots/rate_nwpacific_large_test_legend_better_CSI.pdf}
\end{figure}

\newpage

\end{document}